\documentclass{article} 
\usepackage{conference,times}

\usepackage{amsmath,amsfonts,bm}

\def\eqref#1{equation~\ref{#1}}

\def\1{\bm{1}}

\DeclareMathAlphabet{\mathsfit}{\encodingdefault}{\sfdefault}{m}{sl}
\SetMathAlphabet{\mathsfit}{bold}{\encodingdefault}{\sfdefault}{bx}{n}

\usepackage[utf8]{inputenc} 
\usepackage[T1]{fontenc}    
\usepackage{hyperref}
\usepackage{url}
\usepackage{booktabs}
\usepackage{graphicx}
\usepackage{multirow} 
\usepackage{xspace}
\usepackage{arydshln}
\usepackage{subfigure}
\usepackage{wrapfig}
\usepackage{pifont}

\newcommand{\revision}[1]{{#1}}
\newcommand{\coder}{Retrieval\xspace}
\newcommand{\Finetune}{Retrieval \& Tuning\xspace}
\newcommand{\benchmark}{\textsc{M$^2$rc-Eval}}
\newcommand{\instruct}{\textsc{M$^2$rc-Instruct}}

\title{\benchmark: Massively Multilingual Repository-level Code Completion Evaluation}

\author{
Jiaheng Liu$^{1*,\dagger}$, ~~Ken Deng$^{1*}$,~~Congnan Liu$^{1}$,
~~\textbf{Jian Yang}$^{1}$,
~~\textbf{Shukai Liu}$^{1}$,\\
~~\textbf{He Zhu}$^{1}$,
\textbf{Peng Zhao}$^{1}$,~~\textbf{Linzheng Chai}$^{1}$,~~\textbf{Yanan Wu}$^1$,
~~\textbf{Ke Jin}$^{1}$,
~\textbf{Ge Zhang}$^2$,\\
~~\textbf{Zekun Wang}$^{2}$,
~~\textbf{Guoan Zhang}$^{1}$,
~~\textbf{Bangyu Xiang}$^{1}$,
~~\textbf{Wenbo Su}$^1$, ~~\textbf{Bo Zheng}$^1$\\
~$^1$Alibaba Group,  ~$^2$University of Waterloo\\ 
~~\texttt{\{ljh411989, dengken.deng\}@alibaba-inc.com
}\\
}

\begin{document}

\maketitle
\let\oldthefootnote\thefootnote

\let\thefootnote\relax\footnotetext{* First two authors contributed equally. ~~$^\dagger$ Corresponding Author: Jiaheng Liu.}
\let\thefootnote\oldthefootnote

\newcommand{\fix}{\marginpar{FIX}}
\newcommand{\new}{\marginpar{NEW}}

\maketitle

\begin{abstract}
Repository-level code completion has drawn great attention in software engineering, and several benchmark datasets have been introduced.
However, existing repository-level code completion benchmarks usually focus on a limited
number of languages ($<$5), which cannot evaluate the general code intelligence abilities across different languages for existing code Large Language Models (LLMs).
Besides,
the existing benchmarks usually report overall average scores of different languages,
where the fine-grained abilities in different completion scenarios are ignored.
Therefore,
to facilitate the research of code LLMs in multilingual scenarios, we propose a massively multilingual repository-level code completion benchmark covering 18 programming languages (called \textbf{\benchmark{}}), 
and two types of fine-grained annotations (i.e., \textbf{bucket-level} and \textbf{semantic-level}) on different completion scenarios are provided,
where we obtain these annotations based on the parsed abstract syntax tree. 
Moreover, we also curate a massively multilingual instruction corpora \textbf{\instruct{}} dataset to improve the repository-level code completion abilities of existing code LLMs. Comprehensive experimental results demonstrate the effectiveness of our \benchmark{} and \instruct{}. Code and data will be released at ~\url{https://github.com/M2RC-Eval-Team/M2RC-Eval}.
\end{abstract}

\section{Introduction}
The emergence of Large Language Models (LLMs) specifically designed for code-related tasks has marked a significant advancement in code generation. The code LLMs \citep{codellama,codegeex,deepseek_coder,qwen25coder} pre-trained on extensive datasets comprising billions of code-related tokens further revolutionize the automation of software development tasks, providing contextually relevant code suggestions and facilitating the translation from natural language to code. The generation capability of code LLMs opens up diverse applications in software development, promising to enhance productivity and streamline coding processes. As the field continues to evolve, it presents exciting opportunities for future developments and innovations in automated programming and code assistance.

The code completion task is crucial in modern software development, enhancing coding efficiency and accuracy by predicting and suggesting code segments based on context. Recent advancements in code LLMs \citep{fim} have introduced sophisticated completion techniques, such as prefix-suffix-middle (PSM) and suffix-prefix-middle (SPM) paradigms, which can complete middle code segments given the surrounding context. However, the current benchmark \citep{cceval,repobench} mainly focuses on several programming languages.
For example,
the CrossCodeEval~\citep{cceval} includes four languages (i.e., Python, Java, TypeScript, C\#).
Besides,
existing benchmarks can only provide the average score among all samples,
which can not provide a language-specific evaluation for different programming languages based on their intrinsic structure. \textit{Inspired by the multilingual in-file code generation benchmark MultiPL-E \cite{multiple} and McEval \citep{mceval}, we create a massively multilingual repository-level code completion Evaluation benchmark called \textbf{\benchmark{}} to facilitate the research of the community.}

\begin{figure*}[t]
    \centering
    \includegraphics[width=1.0\linewidth]{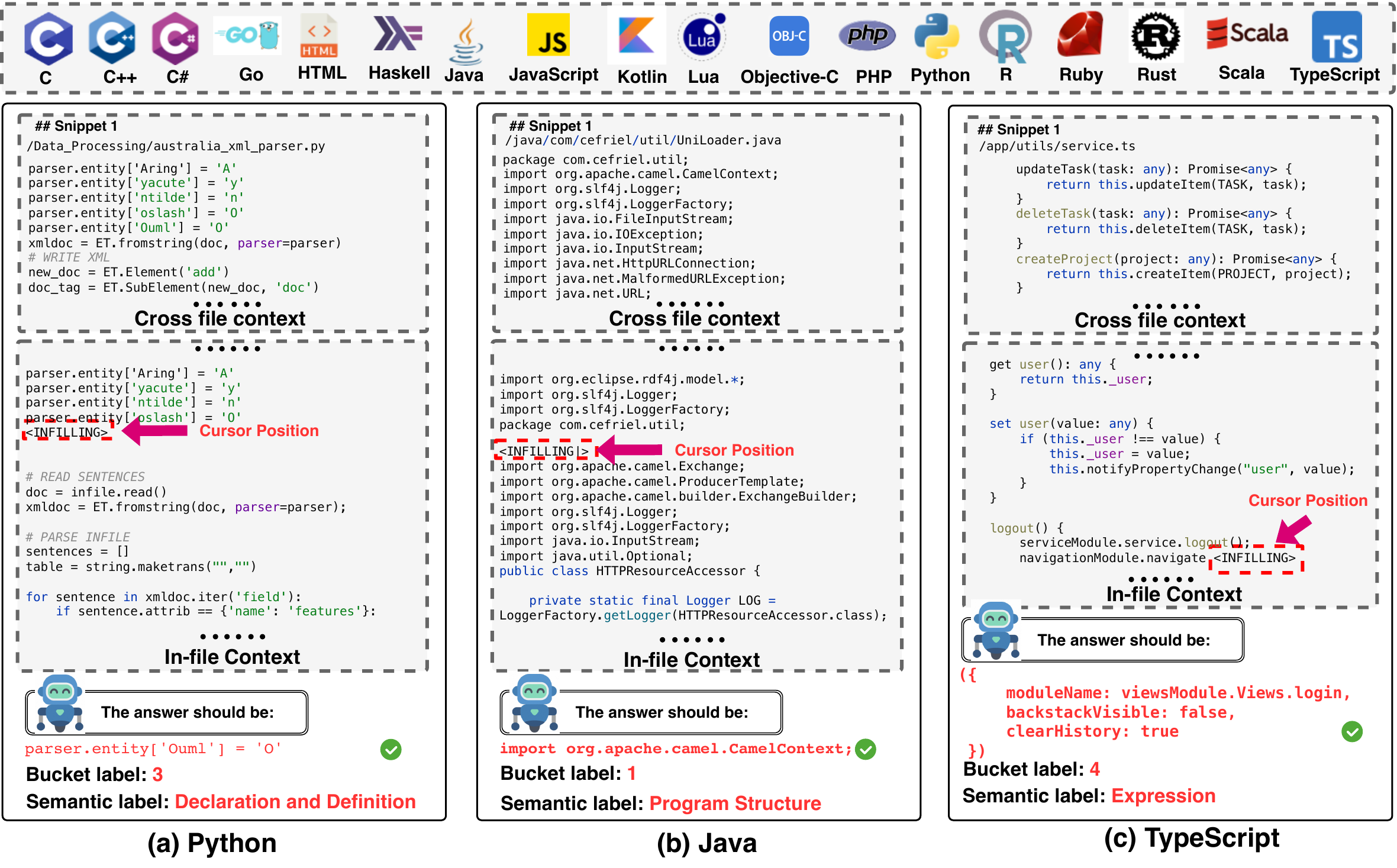}
    \caption{Overview of our proposed \benchmark{} with 18 languages. Specifically, first, we provide three samples from different languages (i.e., Python, Java, TypeScript) for illustration, where the bucket label and semantic label for the corresponding cursor position are provided. 
    Second, the code LLMs need to predict the completion results given the in-file context from the current code file and the cross file context retrieved from other code files in the current repository.
    Note that ``$<\mathrm{INFILLING}>$'' denotes that the current position will be triggered for code completion.}
    \vspace{-5mm}
    \label{fig:intro}
\end{figure*}

In this paper, as shown in Fig.~\ref{fig:intro},
our \benchmark{} includes 18 programming languages with two types of fine-grained annotations (i.e., \textbf{bucket-level} and \textbf{semantic-level}),
where each language contains 100 validation and 500 test samples,
respectively.
Specifically,
for the bucket-level annotations,
we first generate abstract syntax tree with $N$ layers using code parser (i.e., Treesitter~\footnote{\url{https://tree-sitter.github.io/tree-sitter/}}),
and divide these $N$ into fixed $M$ buckets,
Then,
for each completion cursor position,
we annotate the corresponding bucket-level label based on 
the layer to which the location belongs.
In this way,
we can obtain different code completion scenarios with different difficulties.

For the semantic-level annotations,
inspired by~\citep{10.1016/j.infsof.2023.107336},
we first pre-define 11 major semantic labels (e.g., Program Structure, Statement) for each completion cursor position,
which aims to analyze the fine-grained performance across different code semantics.
Note that as different languages usually have specific syntax,
we carefully design the subcategories under each major semantic label for different languages.
Then,
as the code parser usually provides syntax labels (e.g., functions, variables, classes, empty lines)\footnote{Note that the syntax label provided by code parser (e.g., tree-sitter) are highly detailed.} for each completion cursor position,
we carefully define the mappings between the syntax labels to our designed semantic labels and build the semantic-level annotations for our \benchmark{}.
Finally,
to enhance the performance of repository-level code completion for existing code LLMs,
we also create a massively multilingual instruction corpora
\textbf{\instruct{}} of 18 languages. 

The contributions are summarized as follows:
\begin{itemize}
    \item We propose the first massively multilingual repository-level code completion benchmark \benchmark{} covering 18 languages, 
where two types of annotations (bucket-level and semantic-level labels) are provided based on the parsed abstract syntax tree. 
\item We introduce \instruct{}, the
massively multilingual repository-level code instruction corpora covering the multilingual code snippet from 18 languages,
which can greatly enhance the performance of repository-level code completion results.
\item Comprehensive evaluation results and analysis demonstrate the effectiveness of our proposed \benchmark{} and \instruct{}.
\end{itemize}






\section{Related Works}
\paragraph{Code Large Language Models.}

Code large language models (LLMs)~\citep{Chen2021Evaluating, Zhao2024CodeGemmaOC, gpt-neo, black2022gptneox, Le2022CodeRLMC, chowdhery2023palm, nijkamp2022codegen, fried2022incoder, xu2022systematic} are increasingly involved in modern programming, due to excellent capabilities of code generation~\citep{li2022competition,allal2023santacoder}, code repair~\citep{wang-etal-2021-codet5, wang2023codet5+}, code translation~\citep{codegeex,li2023starcoder}, and other coding tasks.
UniCoder~\citep{unicoder} and SPT-Code~\citep{Niu2022SPTCodeSP} introduce the pseudo-code generation and the alignment between Abstract Syntax Tree (AST) and code.
Recent code LLMs such as Code Llama~\citep{codellama}, DeepSeek-Coder~\citep{deepseek_coder}, and Qwen2.5-Coder~\citep{qwen25coder} incorporate the fill-in-the-middle (FIM) task into their training stage for code completion.
Moreover, there is a wide variety of in-file benchmarks to evaluate different capabilities of code LLMs ~\citep{codegeex,mbpp,livecodebench}, which focus on a limited range of programming languages (e.g. Python and Java). 
The recent work~\citep{mceval} extends the number of programming languages to 40 for multilingual evaluation scenarios,
which has not considered the repository-level code completion.

\paragraph{Repository-level Code Completion.} 

The latest repository-level code completion methods~\citep{bairi2023codeplan,Phan2024RepoHyperSO,liao2023a3codgen,shrivastava2023repofusion,agrawal2023guiding, shrivastava2022repository,pei2023better, zhang2023repocoder} are similar to RAG, aim to precisely retrieve all related code snippets across files within a repository.
Further, repository-level benchmarks are proposed to estimate the capability of code LLMs in a more realistic software engineering scenario.
But these datasets~\citep{ding2023cceval, ding2022cocomic, allal2023santacoder} are primarily concentrated on several programming languages.
Regarding difficulty categorization, most methods only consider the number of files involved in the completion content, overlooking the code's structural and semantic context within the entire project.
Repofusion~\citep{shrivastava2023repofusion} and Repocoder~\citep{zhang2023repocoder} predict one line based on the prefix and suffix code, while CoderEval~\citep{yu2024codereval} measures how many third-party libraries are called. 
To comprehensively evaluate the multilingual repository-based code completion of different code LLMs, we push the boundaries of programming languages into 18 languages in \benchmark{} with fine-grained annotations.



\section{M$^2$rc-Eval}
\label{sec:mceval}
\subsection{Data Collection}
\noindent\textbf{The Overall Data Pool.}
\label{sec:data_pool}
We begin by collecting The Stack v2 \citep{lozhkov2024starcoder2}, which consists of permissively licensed repositories from GitHub. Next, we adopt the \texttt{The-stack-v2-dedup}, which includes 784 million source code files spanning 619 programming languages with manual and heuristic pre-processing. Further, we keep only repositories receiving more than 5 stars and containing $[10, 50]$ files. Lastly, preserving files written in 18 common languages, we have 431,353,244 files remaining, constituting the overall data pool.

\begin{figure}[t]
    \centering
    \includegraphics[width=0.8\linewidth]{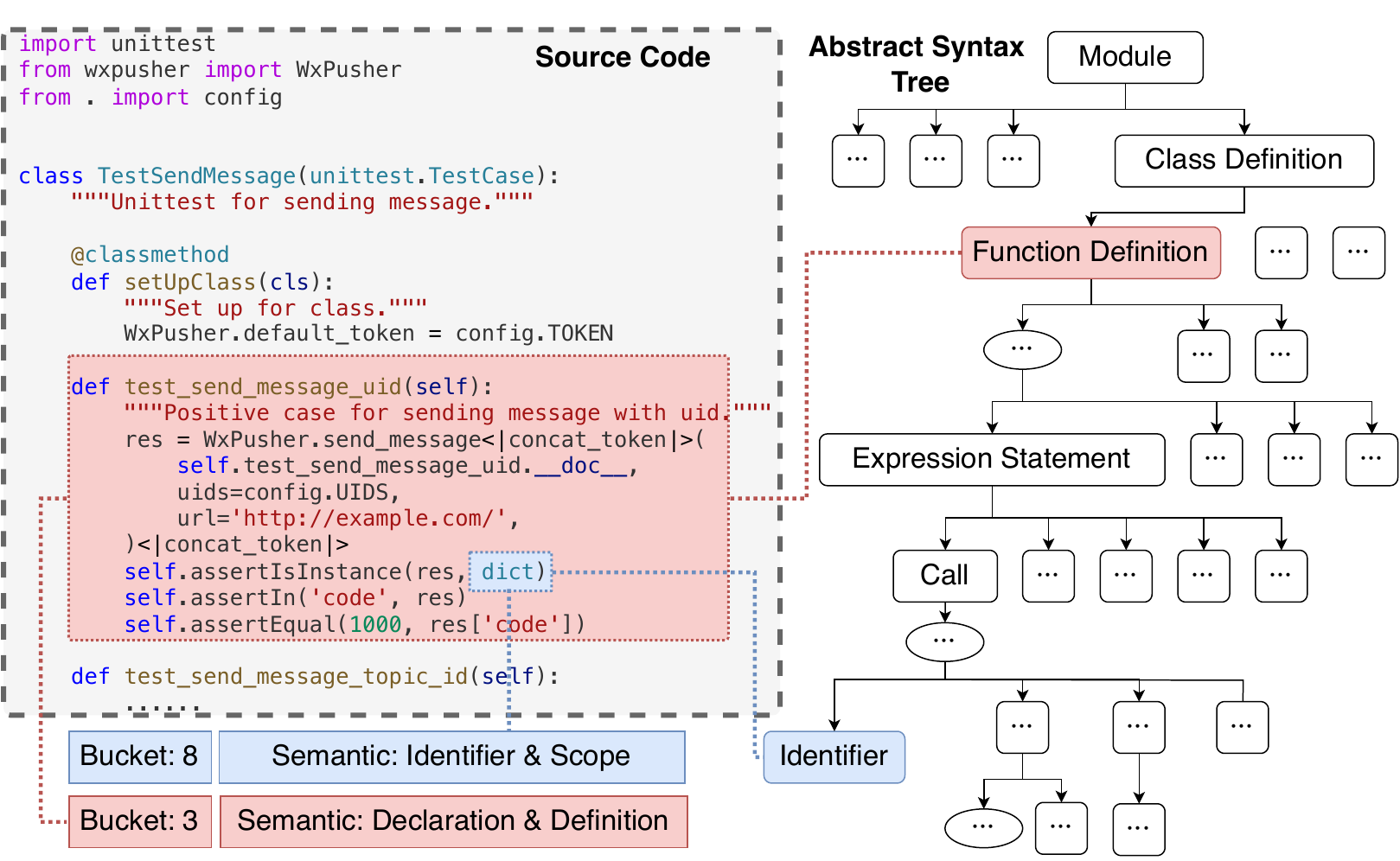}
    \caption{Illustration on generating completion cursor position and fine-grained annotations. Specifically, we first parse the source code into an abstract syntax tree (AST). Then, we choose one node as the completion cursor position and generate the bucket label based on the belonged layer number in AST, and obtain the semantic label based on the node type parsed by the Tree-sitter.}
    \label{fig:parsing}
\end{figure}

\noindent\textbf{Completion Cursor Position Selection.}
\label{sec:data_generation}
Completion cursor position selection significantly impacts the quality of a code completion benchmark. Previous studies \citep{cceval, repobench} randomly select a segment of consecutive characters as the completion span, which does not guarantee the integrity of identifiers and statements.
On the contrary, in \benchmark, we first parse the abstract syntax tree (AST) of each source code file, and then we randomly choose a node (e.g., the node of ``Function Definition'' in Fig.~\ref{fig:parsing}) on the AST as the completion cursor position.
After that, we obtain the corresponding code to obtain the ground-truth for the current completion cursor position. 
Finally,
at inference,
the code LLMs need to predict the current code span given the in-file and cross file contexts.
Similarly,
in training,
we just use the ground-truth to supervise the tuning process of the code LLMs.

\subsection{Quality Control}
\label{sec:quality}
We build a suite of post-processing filters to enhance the quality of \instruct{}. We eliminate examples based on two heuristic rules: (1) The completion cursor position should be no longer than 5 lines. (2) If the completion ground truth is fewer than 20 characters, at least 20\% of them should be alphabetic. To improve data independence and inference difficulty, we apply extra filters to the test cases in \benchmark{}. (a) Repositories in \benchmark{} should be absent from \instruct{}. (b) We ensure that 30\% of the completion ground truth is not shorter than 2 lines. (c) The completion cursor position should not be fully white-spaced. (d) We discard test cases that could be exactly predicted by \texttt{DeepSeekCoder-1.3B} \citep{guo2024deepseek} without cross file contexts. 
\begin{figure}
    \centering
    \includegraphics[width=0.98\linewidth]{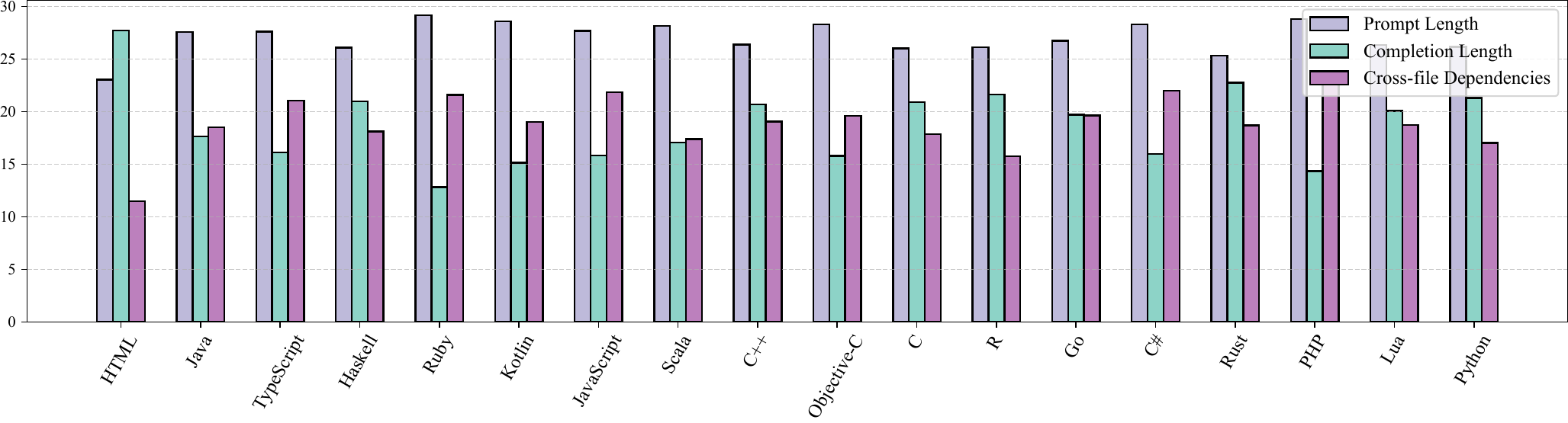}
    \caption{The average prompt length (100x tokens), completion span length (50x tokens), and cross-file dependencies (1x) in the testing set of \benchmark{}. We define the number of other files,
    which are explicitly imported and implicitly referenced by the current file,
    as cross-file dependencies.}
    \label{fig:test_stats}
\end{figure}
\begin{table}[t!]
\centering
\caption{A comparison with existing notable repository-level code completion datasets.}

\begin{tabular}{c|cccc}
\toprule
Benchmark & \# Languages & Fine-grained & Training Set &\# Test Repos \\
\midrule
RepoBench~\citep{repobench} & 2 & \ding{55} & \ding{51} & 1669 \\
CrossCodeEval~\citep{cceval} &4 & \ding{55} & \ding{55} & 1002 \\
R$^2$C$^2$-Bench~\citep{Deng2024R2C2CoderEA} &4 & \ding{55} & \ding{51} & 1353 \\
\midrule
\benchmark \& \instruct &18 & \ding{51} & \ding{51} & 5993 \\ 
\bottomrule
\end{tabular}
\label{tab:dataset_compare}
\vspace{-4mm}
\end{table}
\subsection{Dataset Statistics}
Following the quality filters in \S(\ref{sec:quality}) from the overall data pool \S(\ref{sec:data_pool}).
We sample 50,000 files per language to construct our \instruct{},
and sample 100, and 500 files per language to build the validation and test sets of our \benchmark{},
respectively.
The statistics of the test set are shown in Fig.~\ref{fig:test_stats},
and we also provide a detailed comparison between our \benchmark{} with existing repository-level code completion datasets in Table~\ref{tab:dataset_compare}.

\subsection{Fine-grained Annotations}
\label{sec: anno}
\begin{figure*}[!htb]
	\centering
	\subfigure[Java]{\includegraphics[width=0.32\textwidth]{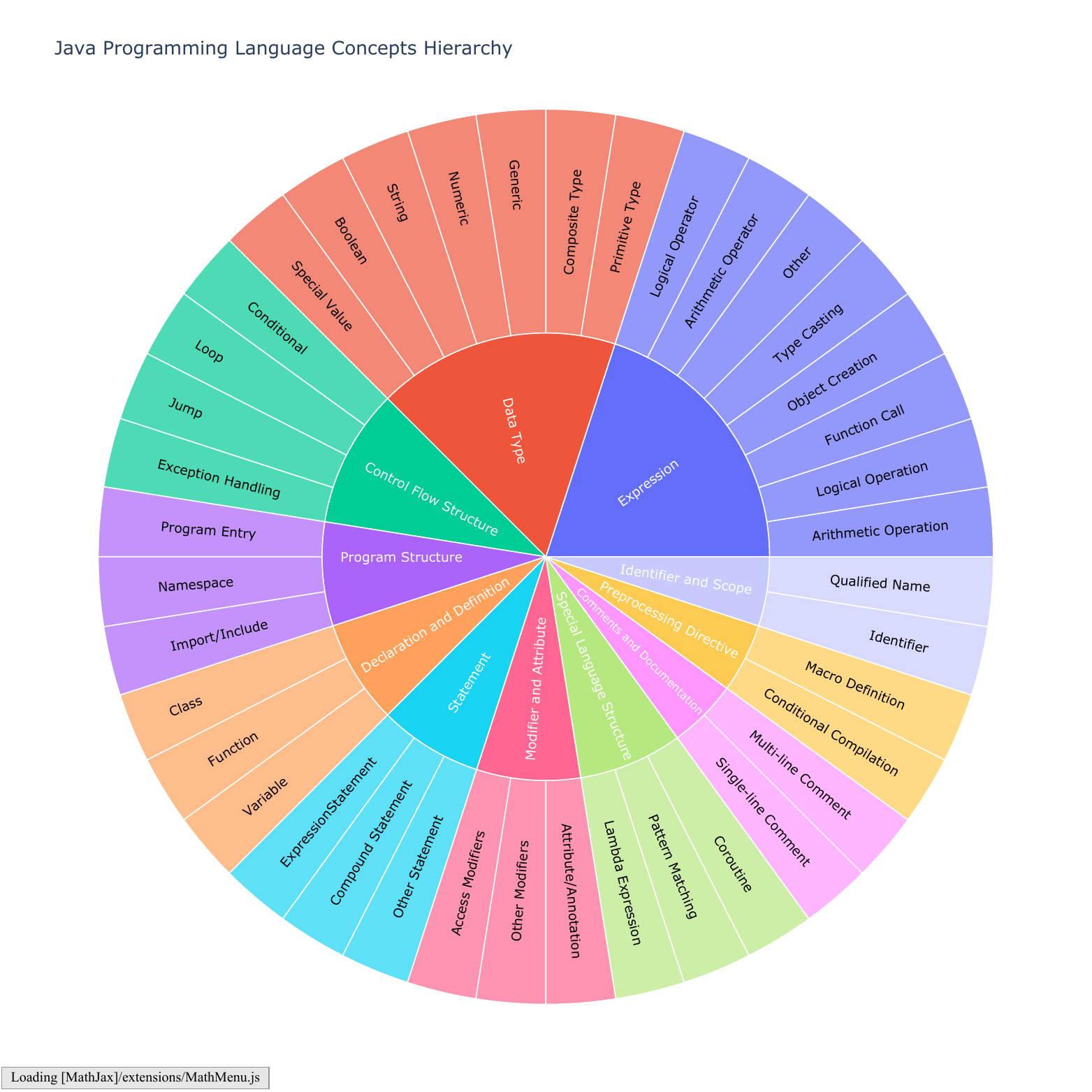}}
	\subfigure[Go]{\includegraphics[width=0.32\textwidth]{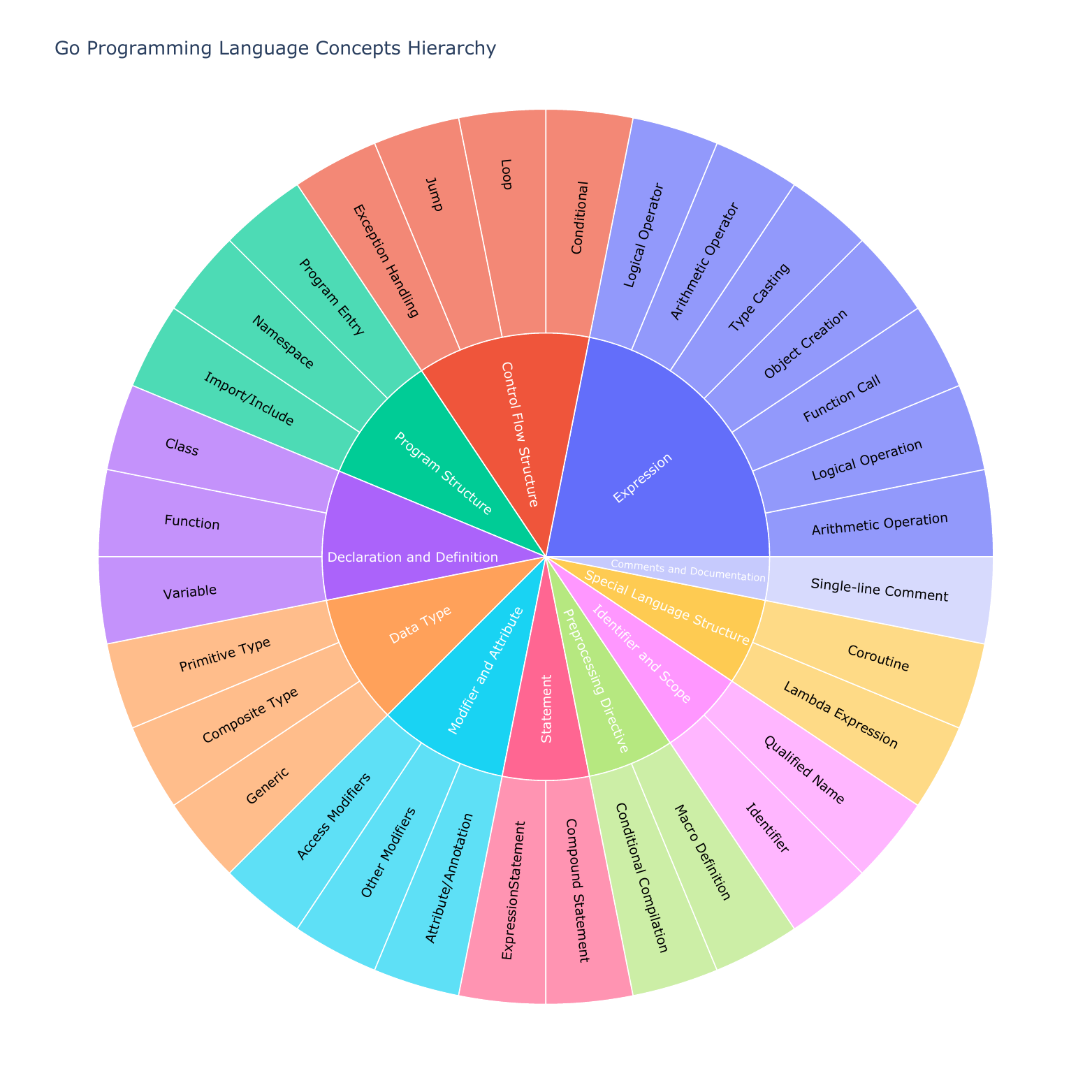}}
	\subfigure[Scala]{\includegraphics[width=0.32\textwidth]{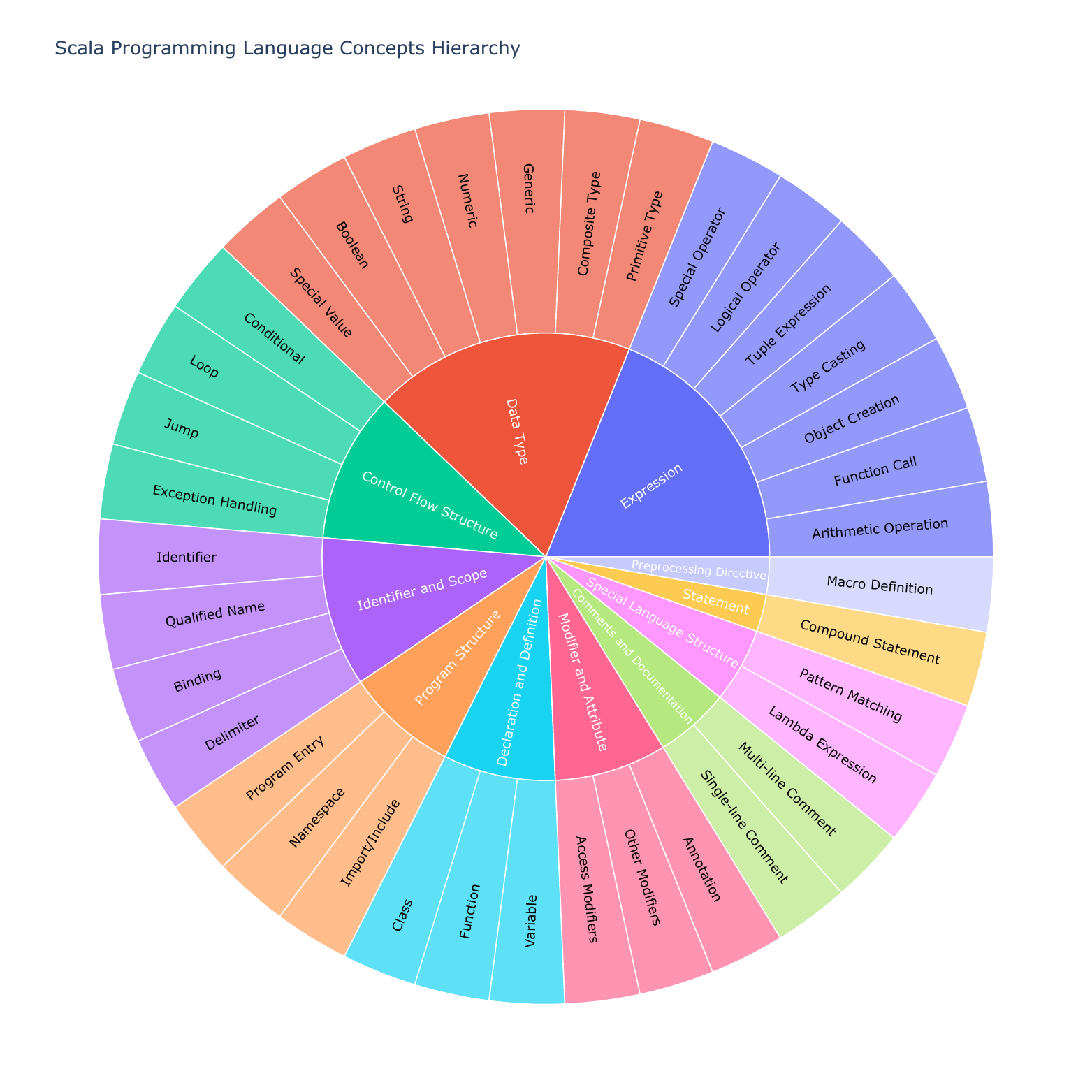}}
	\caption{Semantic-level annotations on different types of programming languages.}
	\label{fig:semantic-label}
\end{figure*}

As shown in Fig.~\ref{fig:parsing},
to analyze the performance in a fine-grained manner,
we further provide two types of fine-grained annotations (i.e., bucket-level and semantic-level) for each completion cursor.
Specifically,
we first generate the abstract syntax tree.
For the bucket-level annotations,
we first simply divide each tree into $M$ buckets based on the depth degree of the abstract syntax tree.
Note that we set $M$ as 10 in our \benchmark{}.
For example,
if the number of layers for the current abstract syntax tree is $N$,
the $i$-th layer of the tree belongs to the $\lceil \frac{i}{N/M} \rceil$ bucket.
Then,
for each completion cursor node, we annotate the bucket label based on the layer number of each node.
Similarly,
for the semantic-level annotations,
we directly annotate the semantic-level label for each completion cursor node.
Specifically,
we pre-define 11 major classes (i.e., ``Program Structure'', ``Declaration and Definition'', ``Control Flow Structure'', ``Expression'', ``Data Type'', ``Statement'', ``Modifier and Attribute'', ``Comments and Documentation'', ``Preprocessing Directive'', ``Identifier and Scope'', ``Special Language Structure'').
Then,
as different languages have many specific designs, the subcategories under each major class are carefully annotated for different languages.
As shown in Fig.~\ref{fig:semantic-label},
we provide the semantic-level annotations on three main-stream programming languages (Java, Go, Scala),
where the annotations on the remained 15 languages are provided in Fig.~\ref{fig:semantic-label-app} of the Appendix.

\section{Experiments}
We perform detailed experiments on  \benchmark{} with three popular Code LLMs (i.e., \textbf{StarCoder-7B}~\citep{li2023starcoder}, \textbf{DeepSeekCoder-6.7B} \citep{guo2024deepseek} and \textbf{Code Llama-7B} \citep{codellama}) (See Appendix~\ref{app:models} for more details).

\subsection{Evaluation Metrics}
Following~\citep{ding2023cceval},
we compare the generated code with the reference and 
compute the exact match (\textbf{EM}) and edit similarity (\textbf{ES}) metrics,
which assesses the textural similarities and ignores semantic structure similarities among predictions and ground-truth.

\subsection{Experimental Setup}
\label{subsec:exp_setup}

\noindent\textbf{Baseline.}
Only the original code file, where the cursor position is located,
is provided for the code LLMs.
As no explicit inter-file context is supplied, the model must utilize its inherent knowledge-based reasoning abilities to generate appropriate code.

\noindent\textbf{+ Retrieval.} 
In line with the approach outlined in CrossCodeEval~\citep{ding2023cceval}, the retrieval process begins by examining files within the same repository. Continuous code segments of $L$ lines are extracted, where $L$ matches the length of the retrieval query and is set as 10 by default. Subsequently, these extracted candidates are prioritized based on their Jaccard similarity scores. The most relevant fragments are then appended to the beginning of the in-file context in descending order of similarity. This concatenation continues until the total length, including both the added candidates and the original in-file context, reaches the predetermined maximum token limit of 4096.



\noindent\textbf{+ Retrieval \& Tuning.}
To further improve the performance of repository-level code completion, we fine-tune code LLMs on the \instruct{} dataset mentioned in \S (\ref{sec:mceval}).
At inference,
we use the same inference strategy as discussed in ``+ Retrieval''.

\subsection{Main Results}
\begin{table*}[t]
    \centering
        \caption{Exact match (\%) and edit similarity (\%) performance on \benchmark{}.}
    \resizebox{0.98\textwidth}{!}{
    \begin{tabular}{lcccccccccccccc}
    \toprule
    \multirow{2}{*}{\bf Model}  &
    \multicolumn{2}{c}{\textbf{C}}&
    \multicolumn{2}{c}{\textbf{C\#}}&
    \multicolumn{2}{c}{\textbf{C++}}&
    \multicolumn{2}{c}{\textbf{Go}}&
    \multicolumn{2}{c}{\textbf{HTML}}&
    \multicolumn{2}{c}{\textbf{Haskell}}&
    \multicolumn{2}{c}{-}\\
    \cmidrule(lr){2-3} \cmidrule(lr){4-5} \cmidrule(lr){6-7} \cmidrule(lr){8-9}\cmidrule(lr){10-11} \cmidrule(lr){12-13} \cmidrule(lr){14-15} &{EM} &{ES}  &{EM} &{ES} &{EM} &{ES} &{EM} &{ES}  &{EM} &{ES}  &{EM} &{ES}  &{EM} &{ES} \\
    \midrule
            Code Llama-7B & 18.6 & 47.2&19.6 & 52.6&21.8 & 51.1&26.0 & 53.6&20.6 & 40.4&22.6 & 48.9 & - & -\\ 
            \noalign{\vskip 0.3ex}\; + \revision{\coder}&
            21.8 & 47.2 & 22.9 & 48.9 & 23.2 & 46.6 & 23.8 & 52.4 & 12.6 & 35.6 & 22.6 & 48.9 & - & -\\
            \noalign{\vskip 0.3ex}\; + \revision{\Finetune}&
            45.4 & 72.0 & 43.5 & 72.3 & 50.8 & 74.9 & 43.4 & 72.9 & 41.8 & 63.6 & 39.8 & 66.3 & - & -\\
            \midrule
            StarCoder-7B  & 20.0 & 50.4 & 20.0 & 53.3 & 22.4 & 51.8 & 25.4 & 58.2 & 17.4 & 40.7 & 25.0 & 51.1 & - & -\\ 
            \noalign{\vskip 0.3ex}\; + \revision{\coder}&
            23.8 & 47.8 & 27.1 & 53.2 & 24.6 & 48.0 & 26.0 & 53.6 & 20.6 & 40.4 & 25.0 & 47.7  & - & -\\
            \noalign{\vskip 0.3ex}\; + \revision{\Finetune}&
            47.0 & 72.7 & 45.1 & 74.8 & 52.4 & 76.3 & 43.2 & 73.7 & 45.8 & 67.1 & 44.8 & 70.2  & - & -\\
            \midrule
            DeepSeekCoder-6.7B & 22.4 & 53.7 & 21.4 & 56.2 & 23.2 & 54.2 & 29.4 & 61.4 & 17.6 & 43.4 & 25.2 & 51.3 & - & -\\
            \noalign{\vskip 0.3ex}\; + \revision{\coder}&
            28.2 & 52.6 & 25.3 & 52.6 & 27.6 & 52.2 & 29.4 & 61.4 & 17.6 & 43.4 & 25.8 & 51.0 & - & -\\
            \noalign{\vskip 0.3ex}\; + \revision{\Finetune}&
            48.6 & 75.2 & 47.9 & 76.9 & 54.4 & 78.2 & 48.8 & 78.4 & 45.0 & 66.3 & 45.8 & 72.0 & - & -\\
    \midrule
            
    \multirow{1}{*}{\bf Model}  &
    \multicolumn{2}{c}{\textbf{Java}}&
    \multicolumn{2}{c}{\textbf{JavaScript}}&
    \multicolumn{2}{c}{\textbf{Kotlin}}&
    \multicolumn{2}{c}{\textbf{Lua}}&
    \multicolumn{2}{c}{\textbf{Objective-C}}&
    \multicolumn{2}{c}{\textbf{PHP}}&
    \multicolumn{2}{c}{-}\\ \midrule
            Code Llama-7B & 23.4 & 58.5&17.2 & 52.0&23.6 & 57.0&20.0 & 45.7&17.8 & 49.5&19.2 & 54.9 & - & -\\
            \noalign{\vskip 0.3ex}\; + \revision{\coder}&
            23.4 & 57.5 & 19.6 & 48.0 & 20.8 & 50.0 & 19.6 & 42.2 & 21.4 & 46.6 & 21.2 & 49.0 & - & -\\
            \noalign{\vskip 0.3ex}\; + \revision{\Finetune}&
            41.8 & 74.1 & 38.8 & 70.1 & 45.0 & 75.6 & 43.8 & 70.5 & 49.8 & 75.9 & 45.6 & 76.7 & - & -\\
            \midrule
            StarCoder-7B  &24.0 & 59.2 & 16.6 & 52.0 & 24.4 & 59.3 & 21.4 & 48.6 & 17.6 & 49.6 & 18.6 & 54.4 & - & -\\ 
            \noalign{\vskip 0.3ex}\; + \revision{\coder}&
            25.0 & 53.1 & 22.0 & 50.8 & 22.8 & 52.6 & 26.4 & 48.5 & 23.6 & 48.0 & 18.6 & 54.4& - & -\\
            \noalign{\vskip 0.3ex}\; + \revision{\Finetune}&
            47.4 & 76.9 & 38.8 & 70.1 & 45.0 & 75.6 & 43.8 & 70.5 & 50.8 & 75.9 & 45.6 & 76.7& - & -\\
            \midrule
            DeepSeekCoder-6.7B & 22.2 & 61.0 & 20.4 & 56.5 & 26.0 & 61.0 & 22.0 & 48.8 & 21.0 & 55.6 & 24.2 & 58.6 & - & -\\
            \noalign{\vskip 0.3ex}\; + \revision{\coder}&
            21.6 & 51.4 & 24.4 & 53.6 & 26.0 & 61.0 & 22.0 & 49.9 & 27.6 & 53.5 & 28.6 & 56.9 & - & -\\
            \noalign{\vskip 0.3ex}\; + \revision{\Finetune}&
            48.2 & 79.1 & 43.6 & 73.5 & 46.0 & 75.7 & 44.6 & 70.6 & 52.2 & 77.6 & 49.8 & 78.8& - & -\\   
    \midrule

    \multirow{1}{*}{\bf Model}  &
    \multicolumn{2}{c}{\textbf{Python}}&
    \multicolumn{2}{c}{\textbf{R}}&
    \multicolumn{2}{c}{\textbf{Ruby}}&
    \multicolumn{2}{c}{\textbf{Rust}}&
    \multicolumn{2}{c}{\textbf{Scala}}&
    \multicolumn{2}{c}{\textbf{TypeScript}}&
    \multicolumn{2}{c}{\textbf{Avg.}}\\\midrule
            Code Llama-7B & 24.6 & 54.2&15.2 & 41.2&17.2 & 45.8&26.2 & 56.0&22.8 & 48.5&23.4 & 52.3& 19.4&  50.3\\ 
            \noalign{\vskip 0.3ex}\; + \revision{\coder}&
            17.4 & 46.4 & 15.2 & 39.8 & 17.2 & 42.3 & 26.0 & 51.3 & 22.8 & 48.5 & 19.4 & 48.6 & 20.2 & 46.1\\
            \noalign{\vskip 0.3ex}\; + \revision{\Finetune}&
            39.2 & 69.9 & 38.6 & 65.5 & 43.0 & 68.5 & 42.0 & 69.2 & 41.0 & 70.1 & 37.0 & 68.2 & 41.9 & 70.0\\
            \midrule
            StarCoder-7B & 19.4 & 52.9 & 16.4 & 43.7 & 19.4 & 47.4 & 26.2 & 56.0 & 23.6 & 53.4 & 19.8 & 53.3 & 21.0 & 52.0\\
            \noalign{\vskip 0.3ex}\; + \revision{\coder}&
            24.6 & 54.2 & 22.6 & 47.2 & 23.6 & 47.4 & 26.4 & 53.5 & 22.8 & 48.5 & 23.4 & 52.3 & 24.1 & 50.0\\
            \noalign{\vskip 0.3ex}\; + \revision{\Finetune}&
            39.2 & 69.9 & 41.0 & 66.6 & 43.0 & 68.5 & 45.8 & 72.6 & 43.6 & 71.5 & 39.2 & 69.7 & 44.5 & 72.2\\
            \midrule
            DeepSeekCoder-6.7B & 21.8 & 55.1 & 19.4 & 48.5 & 23.6 & 52.2 & 23.8 & 54.3 & 24.6 & 56.7 & 19.4 & 55.4 & 22.6 & 54.7\\
            \noalign{\vskip 0.3ex}\; + \revision{\coder}&
            21.8 & 55.1 & 19.4 & 48.5 & 23.6 & 52.2 & 23.8 & 54.3 & 22.4 & 50.4 & 26.0 & 54.5 & 25.1 & 51.7\\
            \noalign{\vskip 0.3ex}\; + \revision{\Finetune}&
            41.6 & 71.3 & 45.4 & 69.4 & 45.6 & 70.3 & 47.6 & 73.4 & 44.8 & 73.7 & 43.2 & 73.4 & \textbf{46.8} & \textbf{74.1}\\  
    \bottomrule
    \end{tabular}}

    \label{tab:main_benchmark}
\end{table*}

We present the results on \benchmark{} in Table~\ref{tab:main_benchmark}. 
We observe that different code LLMs have different repository-level code completion abilities for different programming languages. For instance, DeepSeekCoder-6.7B demonstrates strong completion ability for Go, while its performance is weaker with HTML, a markup language,
which demonstrates the necessity of evaluating code LLMs for multilingual capabilities.
Besides, the results indicate that cross file context is highly effective, resulting in a significant improvement compared to using only in-file context. In particular, the multilingual SFT on our created instruction corpora \instruct{} also significantly enhances performance on \benchmark{}. Notably, after SFT on \instruct{}, Code Llama-7B, which originally ranked lowest with in-file context, outperformed the non-finetuned StarCoder-7B, demonstrating the effectiveness of \instruct{}.

\subsection{Analysis}
\noindent\textbf{Analysis on different model sizes.}
In Table~\ref{tab:abs_modelscale}, we provide the performance of StarCoder with respect to different model sizes in the validation set of \benchmark{}. Notably, StarCoder-7B consistently outperforms StarCoder-3B under comparable conditions. However, following the application of SFT on \instruct{}, the results of StarCoder-3B exceed those of the inference-only StarCoder-7B. This finding underscores the effectiveness of our \instruct{} in augmenting the capabilities of smaller models in repository-level code completion.
\begin{wraptable}{r}{0.5\textwidth}
    \centering
    \caption{Performance on  \benchmark.}
    \label{tab:abs_modelscale}
    \begin{tabular}{l c c}
        \toprule
        \multirow{2}{*}{\bf Model}  &
        \multicolumn{2}{c}{\textbf{Average}}\\
        \cmidrule(lr){2-3} 
        &{EM} &{ES}  \\\midrule
        
        \revision{StarCoder-3B} &
        14.9& 43.5   \\
        \hdashline
        \noalign{\vskip 0.4ex}\; + \revision{\coder}&
        14.6& 38.4 \\
        \hdashline
        \noalign{\vskip 0.4ex}\; + \revision{\Finetune}&
        41.7 & 69.1 \\
        \midrule
        \revision{StarCoder-7B} &
        20.6 & 49.9   \\
        \hdashline
        \noalign{\vskip 0.4ex}\; + \revision{\coder}&
        23.6& 49.3\\
        \hdashline
        \noalign{\vskip 0.4ex}\; + \revision{\Finetune}&
        44.4 & 71.4   \\
        \bottomrule
    \end{tabular}
    \vspace{-3mm}
\end{wraptable}
            
\begin{figure}[t]
    \centering
    \includegraphics[width=0.9\linewidth]{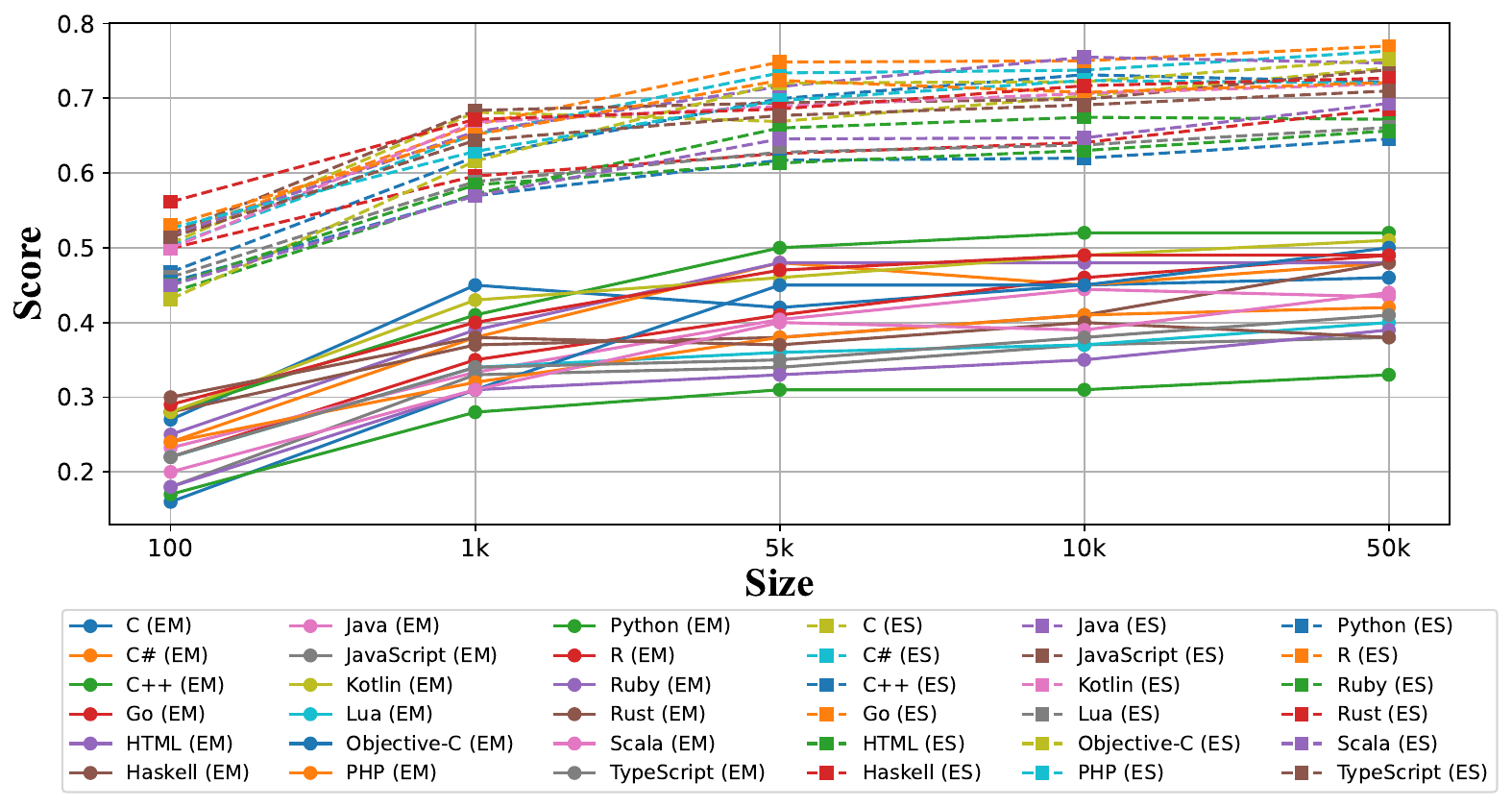}
    \caption{Effectiveness of using different training data sizes.}
    \vspace{-3mm}
    \label{fig:overview}
\end{figure}

    
\noindent\textbf{Analysis on different training data sizes.}
In Fig.~\ref{fig:overview},  we conduct a evaluation of the fine-tuned StarCoder-7B by employing varying sizes of \instruct{} and report the results on validation set \benchmark{}.
Our observations indicate that increasing the dataset from 0.1k to 50k samples per language yields improved results. This suggests that more training data can help boost the model's performance. Therefore, we select 50k samples per language as the default training set size for our \benchmark{}.

\begin{figure}[t]
    \centering
    \includegraphics[width=0.7\linewidth]{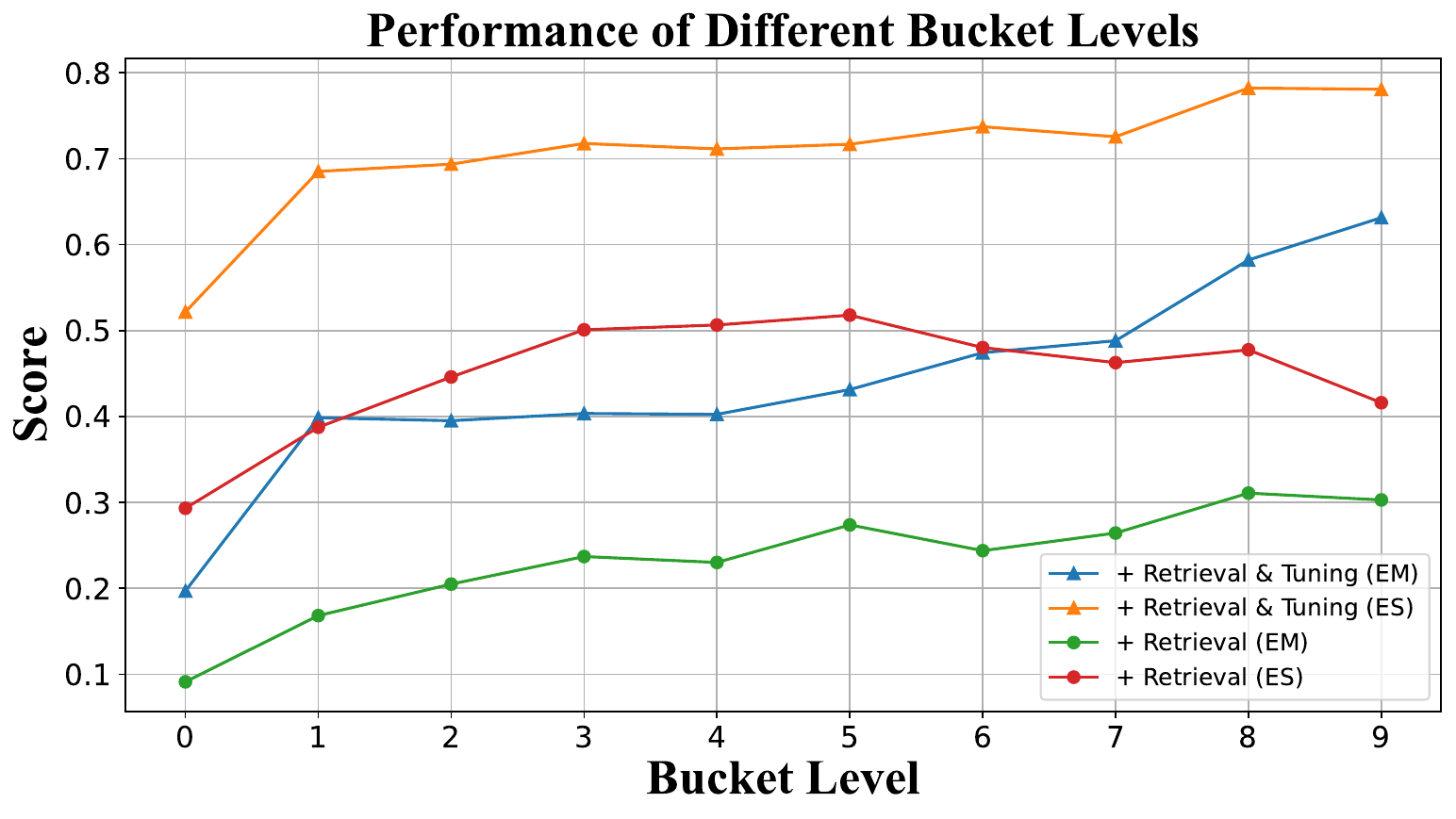}
    \vspace{-4mm}
    \caption{Effectiveness of different bucket levels based on StarCoder-7B.}
    \vspace{-5mm}
    \label{fig:depth levels}
\end{figure} 

\noindent\textbf{Analysis on the granularity of different bucket levels.}
As mentioned in \S (~\ref{sec: anno}),
we categorize \benchmark{} into ten bucket levels based on the positions of the code requiring completion within the abstract syntax tree. As shown in Fig.~\ref{fig:depth levels},
we presents the performance of StarCoder-7B on the test set of \benchmark{} across these different bucket levels,
and we observe that as the bucket level decreases, the performance of StarCoder-7B correspondingly declines,
which means that the code completion on the shadow layer is usually more challenging than on the deep layer.
For more experimental data on single-language completion performance and its relation to bucket levels, please refer to  Fig.\ref{fig:level-python-12} and Fig.\ref{fig:level-python-6} in the Appendix.
These findings suggest that the code LLMs encounter challenges when addressing shallow nodes within the syntax tree during the code completion process.
\begin{figure}[t]
    \centering
    \includegraphics[width=1.0\linewidth]{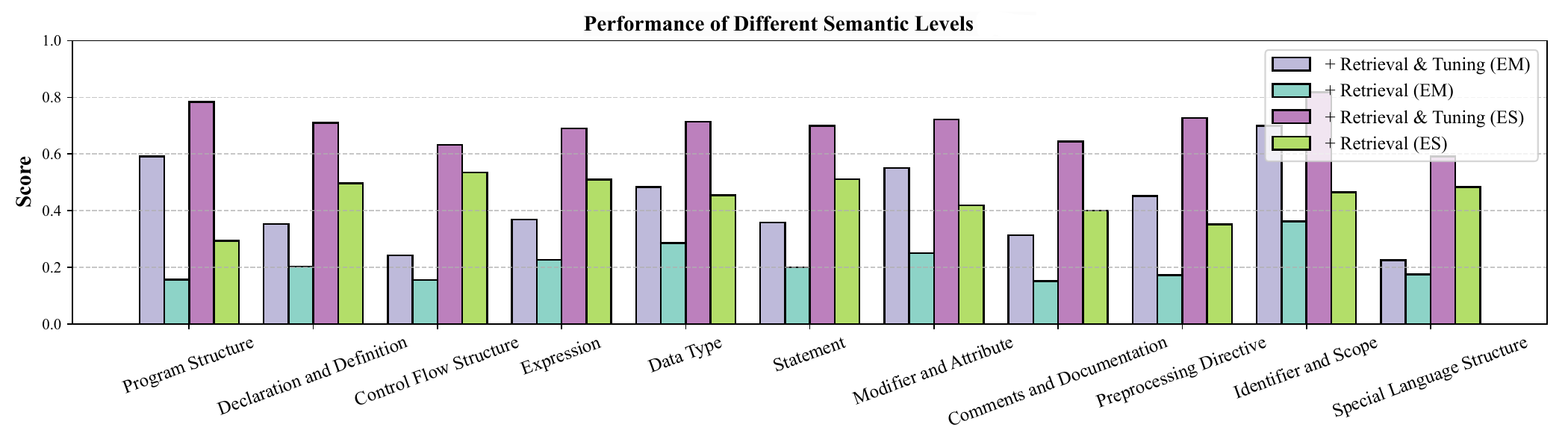}
    \vspace{-4mm}
    \caption{Effectiveness of different semantic levels based on StarCoder-7B.}
    \vspace{-5mm}
    \label{overview-all}
\end{figure} 

\begin{table}[t]
    \centering
        \caption{CodeBLEU results on ten representative programming languages.}
    \resizebox{1.0\textwidth}{!}{
    \begin{tabular}{lccccccccccc}
    \toprule
    {\bf Model}  &\textbf{C}&\textbf{C\#}&\textbf{C++}&\textbf{Go}&\textbf{Java}&\textbf{JavaScript}&\textbf{PHP}&\textbf{Python}&\textbf{Ruby}&\textbf{Rust}&\textbf{Avg.}\\
    \midrule

            StarCoder-7B  & 48.3 & 48.9& 50.4 & 51.5 & 50.6& 46.4 & 48.2 & 46.4 & 46.1 & 50.4 & 48.7\\ 
            \noalign{\vskip 0.3ex}\; + \revision{\coder}&
            50.1 & 52.3 & 51.1 & 52.5& 51.4 & 49.3 & 52.2& 49.3& 49.1 & 51.4 & 50.9\\
            \noalign{\vskip 0.3ex}\; + \revision{\Finetune}&
            56.0 & 57.4 & 57.6& 57.0 & 57.6 & 54.8 & 57.8 & 52.0& 52.9 & 55.5 & \textbf{55.9} \\
    \bottomrule
    \end{tabular}}
    \label{tab:codebleu}
\end{table}

\noindent\textbf{Analysis on the granularity of different semantic levels.}
Similarly,
in \S(~\ref{sec: anno}),
we also categorize the nodes within the abstract syntax tree into eleven primary semantic levels based on their semantic characteristics,
and we provide the performance of StarCoder-7B on repository-level code completion for these various semantic levels  across multilingual languages on the test set of the \benchmark{}.
Notably, we observe significant performance disparities across different semantic levels.
Specifically,  StarCoder-7B shows superior performance on ``Identifier and Scope'', while it exhibits lower efficacy on ``Special Language Structure'',
This suggests that current code LLMs are proficient at completing tasks related to variable definitions and references, yet their capacity to handle characteristics of different languages requires further enhancement. 
For single-language completion performance across various node types, please refer to Fig.~\ref{fig:semantic-levels} in the Appendix.

\begin{wrapfigure}{r}{0.35\linewidth} 
    \centering
    \includegraphics[width=\linewidth]{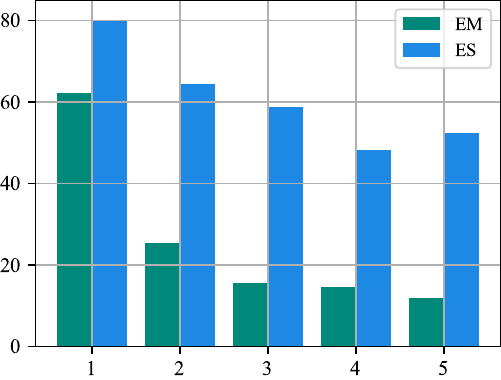}
    \caption{Effectiveness of code completion on different lines based on StarCoder-7B.}
    \label{fig:different lines}
\end{wrapfigure}

\begin{figure}[t]
    \centering
    \includegraphics[width=1.0\linewidth]{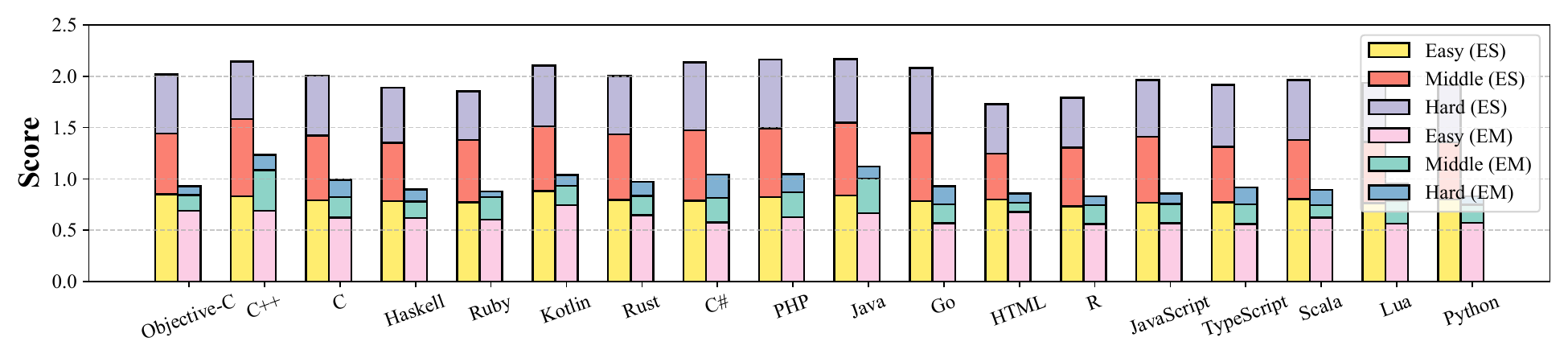}
    \caption{Performance on \benchmark{} for problems of different difficulty levels.}
    \label{fig:multi-line-lang}
\end{figure}

\noindent\textbf{Analysis on completion on different lines.}
To explore the differences in the code completion capabilities of code LLMs for varying line counts, we conduct a statistical analysis of the performance of StarCoder-7B (``\Finetune'') across various completion lengths on the test set of \benchmark{}. As shown in Fig.\ref{fig:different lines}, StarCoder-7B can effectively complete tasks involving a small number of lines. However, as the number of lines to be completed increases, the scores of the generated code gradually declines. This indicates that completing multi-line code remains a challenge for code LLMs.
For better illustration,
we also provide the results of different difficulties for different languages in Fig.~\ref{fig:multi-line-lang},
where we define completion on  1 line, completion on  2-3 lines and completion on  4-5 lines as easy, middle, and hard settings,
respectively.


			
\begin{wraptable}{r}{0.6\textwidth}
    \centering
    \caption{Performance on \benchmark.}
    \label{tab:abs_crosslingual}
    \begin{tabular}{l c c}
        \toprule
        \multirow{2}{*}{\bf Model}  &
        \multicolumn{2}{c}{\textbf{Average}}\\
        \cmidrule(lr){2-3} 
        &{EM} &{ES}  \\\midrule
        \noalign{\vskip 0.4ex}\; + \revision{\coder}&
        23.6& 49.3\\
        \hdashline
        \noalign{\vskip 0.4ex}\; + \revision{\Finetune}&
        44.4 & 71.4   \\
                \noalign{\vskip 0.4ex}\; + \revision{\Finetune} (Python Only)&
        39.2 & 67.9   \\
        \bottomrule
    \end{tabular}
\end{wraptable}

\noindent\textbf{Analysis on cross-lingual transfer.}
We fine-tune the StarCoder-7B model using Python-only data (50k) in
\instruct{} and compare it with the results of using our whole training data.
In Table~\ref{tab:abs_crosslingual},
we report the results on the validation set of \benchmark{},
and observe that fine-tuning the model exclusively with Python data resulted in a significant improvement in its \benchmark{} score, coming close to the ES performance achieved through fine-tuning with data from 18 languages. This outcome suggests that the StarCoder-7B base model already demonstrates strong coding proficiency, but lacks robust instruction-following capabilities. Thus, even when fine-tuned solely on Python data, the model is able to effectively transfer instruction-following skills to other languages, thereby achieving exceptional multilingual performance.



\noindent\textbf{Analysis on CodeBLEU metric.}
In Table~\ref{tab:main_benchmark},
we mainly report the EM and ES metrics based on the textural similarity,
which neglects important syntactic and semantic features of codes and underestimates different outputs with the same semantic logic.
Thus, the CodeBLEU~\citep{ren2020codebleu}~\footnote{We test the CodeBLEU metric based on \url{https://github.com/k4black/codebleu}.} is proposed,
which considers information from not only the shallow match, but also the syntactic match and the semantic match.
In Table~\ref{tab:codebleu},
we report the results of 10 popular programming languages using the test split of \benchmark{} based on the StarCoder-7B model and observe that we can still achieve better performance by fine-tuning on our constructed \instruct{},
which further demonstrates the effectiveness of our \instruct{} on repository-level code completion.

\begin{wrapfigure}[15]{r}{0.45\linewidth} 
    \centering
    \includegraphics[width=\linewidth]{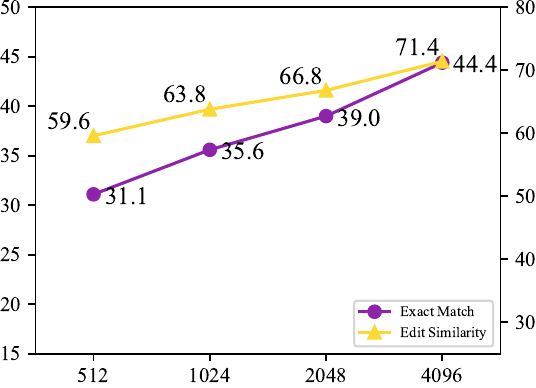}
    \caption{Performance on \benchmark{} with various input lengths based on StarCoder-7B.}
	\label{fig:input-2}
\end{wrapfigure}

\noindent\textbf{Analysis on various input lengths.} As shown in Fig.~\ref{fig:input-2}, we report the results produced by StarCoder-7B (``\Finetune'') on our \benchmark{} when the input lengths of range in \{512, 1024, 2048, 4096\} tokens. 
In Fig.~\ref{fig:input-2}, we observe that a scaling law exists,
where better performance is achieved when the input length is larger.
Thus,
we set the default input length as 4096 tokens.



\section{Conclusion}
In this paper, we propose the first massively multilingual repository-level code completion benchmark (\benchmark{}) with 18 popular programming languages,
where two types of fine-grained annotations (bucket-level and semantic-level) are provided to comprehensively analyze the effectiveness of different code LLMs.
Besides,
we also curate a high-quality instruction corpus \instruct{} to enhance the performance of existing models on repository-level code completion.
Extensive experimental results and detailed discussions demonstrate the effectiveness of our proposed \benchmark{} and \instruct{}.
Finally, we hope \benchmark{} could guide the developers and researchers to understand the repository-level code completion capabilities of LLMs and facilitate the growth of code intelligence and software engineering.

\bibliography{M2RC2_eval}

\begin{thebibliography}{49}
\providecommand{\natexlab}[1]{#1}
\providecommand{\url}[1]{\texttt{#1}}
\expandafter\ifx\csname urlstyle\endcsname\relax
  \providecommand{\doi}[1]{doi: #1}\else
  \providecommand{\doi}{doi: \begingroup \urlstyle{rm}\Url}\fi

\bibitem[Agrawal et~al.(2023)Agrawal, Kanade, Goyal, Lahiri, and Rajamani]{agrawal2023guiding}
Lakshya~A Agrawal, Aditya Kanade, Navin Goyal, Shuvendu~K Lahiri, and Sriram~K Rajamani.
\newblock Guiding language models of code with global context using monitors.
\newblock 2023.

\bibitem[Allal et~al.(2023)Allal, Li, Kocetkov, Mou, Akiki, Ferrandis, Muennighoff, Mishra, Gu, Dey, et~al.]{allal2023santacoder}
Loubna~Ben Allal, Raymond Li, Denis Kocetkov, Chenghao Mou, Christopher Akiki, Carlos~Munoz Ferrandis, Niklas Muennighoff, Mayank Mishra, Alex Gu, Manan Dey, et~al.
\newblock Santacoder: don't reach for the stars!
\newblock \emph{arXiv preprint arXiv:2301.03988}, 2023.
\newblock URL \url{https://arxiv.org/abs/2301.03988}.

\bibitem[Austin et~al.(2021)Austin, Odena, Nye, Bosma, Michalewski, Dohan, Jiang, Cai, Terry, Le, et~al.]{mbpp}
Jacob Austin, Augustus Odena, Maxwell Nye, Maarten Bosma, Henryk Michalewski, David Dohan, Ellen Jiang, Carrie Cai, Michael Terry, Quoc Le, et~al.
\newblock Program synthesis with large language models.
\newblock \emph{arXiv preprint arXiv:2108.07732}, 2021.
\newblock URL \url{https://arxiv.org/abs/2108.07732}.

\bibitem[Bairi et~al.(2023)Bairi, Sonwane, Kanade, Iyer, Parthasarathy, Rajamani, Ashok, Shet, et~al.]{bairi2023codeplan}
Ramakrishna Bairi, Atharv Sonwane, Aditya Kanade, Arun Iyer, Suresh Parthasarathy, Sriram Rajamani, B~Ashok, Shashank Shet, et~al.
\newblock Codeplan: Repository-level coding using llms and planning.
\newblock \emph{arXiv preprint arXiv:2309.12499}, 2023.

\bibitem[Bavarian et~al.(2022{\natexlab{a}})Bavarian, Jun, Tezak, Schulman, McLeavey, Tworek, and Chen]{bavarian2022efficient}
Mohammad Bavarian, Heewoo Jun, Nikolas Tezak, John Schulman, Christine McLeavey, Jerry Tworek, and Mark Chen.
\newblock Efficient training of language models to fill in the middle.
\newblock \emph{arXiv preprint arXiv:2207.14255}, 2022{\natexlab{a}}.
\newblock URL \url{https://arxiv.org/abs/2207.14255}.

\bibitem[Bavarian et~al.(2022{\natexlab{b}})Bavarian, Jun, Tezak, Schulman, McLeavey, Tworek, and Chen]{fim}
Mohammad Bavarian, Heewoo Jun, Nikolas Tezak, John Schulman, Christine McLeavey, Jerry Tworek, and Mark Chen.
\newblock Efficient training of language models to fill in the middle.
\newblock \emph{arXiv preprint arXiv:2207.14255}, 2022{\natexlab{b}}.

\bibitem[Black et~al.(2021)Black, Gao, Wang, Leahy, and Biderman]{gpt-neo}
Sid Black, Leo Gao, Phil Wang, Connor Leahy, and Stella Biderman.
\newblock {GPT-Neo: Large Scale Autoregressive Language Modeling with Mesh-Tensorflow}, 2021.
\newblock URL \url{https://doi.org/10.5281/zenodo.5297715}.

\bibitem[Black et~al.(2022)Black, Biderman, Hallahan, Anthony, Gao, Golding, He, Leahy, McDonell, Phang, Pieler, Prashanth, Purohit, Reynolds, Tow, Wang, and Weinbach]{black2022gptneox}
Sidney Black, Stella Biderman, Eric Hallahan, Quentin Anthony, Leo Gao, Laurence Golding, Horace He, Connor Leahy, Kyle McDonell, Jason Phang, Michael Pieler, Usvsn~Sai Prashanth, Shivanshu Purohit, Laria Reynolds, Jonathan Tow, Ben Wang, and Samuel Weinbach.
\newblock {GPT}-{N}eo{X}-20{B}: An open-source autoregressive language model.
\newblock In \emph{Proceedings of BigScience Episode {\#}5 -- Workshop on Challenges {\&} Perspectives in Creating Large Language Models}, pp.\  95--136, virtual+Dublin, 2022. Association for Computational Linguistics.
\newblock \doi{10.18653/v1/2022.bigscience-1.9}.
\newblock URL \url{https://aclanthology.org/2022.bigscience-1.9}.

\bibitem[Cassano et~al.(2022)Cassano, Gouwar, Nguyen, Nguyen, Phipps-Costin, Pinckney, Yee, Zi, Anderson, Feldman, et~al.]{multiple}
Federico Cassano, John Gouwar, Daniel Nguyen, Sydney Nguyen, Luna Phipps-Costin, Donald Pinckney, Ming-Ho Yee, Yangtian Zi, Carolyn~Jane Anderson, Molly~Q Feldman, et~al.
\newblock Multipl-e: A scalable and extensible approach to benchmarking neural code generation.
\newblock \emph{arXiv preprint arXiv:2208.08227}, 2022.

\bibitem[Chai et~al.(2024)Chai, Liu, Yang, Yin, Jin, Liu, Sun, Zhang, Ren, Guo, et~al.]{mceval}
Linzheng Chai, Shukai Liu, Jian Yang, Yuwei Yin, Ke~Jin, Jiaheng Liu, Tao Sun, Ge~Zhang, Changyu Ren, Hongcheng Guo, et~al.
\newblock Mceval: Massively multilingual code evaluation.
\newblock \emph{arXiv preprint arXiv:2406.07436}, 2024.

\bibitem[Chen et~al.(2021)Chen, Tworek, Jun, Yuan, Pinto, Kaplan, Edwards, Burda, Joseph, Brockman, et~al.]{Chen2021Evaluating}
Mark Chen, Jerry Tworek, Heewoo Jun, Qiming Yuan, Henrique Ponde de~Oliveira Pinto, Jared Kaplan, Harri Edwards, Yuri Burda, Nicholas Joseph, Greg Brockman, et~al.
\newblock Evaluating large language models trained on code.
\newblock \emph{ArXiv preprint}, abs/2107.03374, 2021.
\newblock URL \url{https://arxiv.org/abs/2107.03374}.

\bibitem[Chowdhery et~al.(2023)Chowdhery, Narang, Devlin, Bosma, Mishra, Roberts, Barham, Chung, Sutton, Gehrmann, et~al.]{chowdhery2023palm}
Aakanksha Chowdhery, Sharan Narang, Jacob Devlin, Maarten Bosma, Gaurav Mishra, Adam Roberts, Paul Barham, Hyung~Won Chung, Charles Sutton, Sebastian Gehrmann, et~al.
\newblock Palm: Scaling language modeling with pathways.
\newblock 24\penalty0 (240):\penalty0 1--113, 2023.

\bibitem[CodeGeeX(2022)]{humaneval_x}
CodeGeeX, 2022.
\newblock \url{https://github.com/THUDM/CodeGeeX}.

\bibitem[Deng et~al.(2024)Deng, Liu, Zhu, Liu, Li, Wang, Zhao, Zhang, Wu, Yin, Zhang, Su, Xiang, Ge, and Zheng]{Deng2024R2C2CoderEA}
Ken Deng, Jiaheng Liu, He~Zhu, Congnan Liu, Jingxin Li, Jiakai Wang, Peng Zhao, Chenchen Zhang, Yanan Wu, Xueqiao Yin, Yuanxing Zhang, Wenbo Su, Bangyu Xiang, Tiezheng Ge, and Bo~Zheng.
\newblock R2c2-coder: Enhancing and benchmarking real-world repository-level code completion abilities of code large language models.
\newblock \emph{ArXiv}, abs/2406.01359, 2024.

\bibitem[Ding et~al.(2022)Ding, Wang, Ahmad, Ramanathan, Nallapati, Bhatia, Roth, and Xiang]{ding2022cocomic}
Yangruibo Ding, Zijian Wang, Wasi~Uddin Ahmad, Murali~Krishna Ramanathan, Ramesh Nallapati, Parminder Bhatia, Dan Roth, and Bing Xiang.
\newblock Cocomic: Code completion by jointly modeling in-file and cross-file context.
\newblock \emph{arXiv preprint arXiv:2212.10007}, 2022.
\newblock URL \url{https://arxiv.org/abs/2212.10007}.

\bibitem[Ding et~al.(2023)Ding, Wang, Ahmad, Ding, Tan, Jain, Ramanathan, Nallapati, Bhatia, Roth, and Xiang]{ding2023cceval}
Yangruibo Ding, Zijian Wang, Wasi~Uddin Ahmad, Hantian Ding, Ming Tan, Nihal Jain, Murali~Krishna Ramanathan, Ramesh Nallapati, Parminder Bhatia, Dan Roth, and Bing Xiang.
\newblock Crosscodeeval: {A} diverse and multilingual benchmark for cross-file code completion.
\newblock In Alice Oh, Tristan Naumann, Amir Globerson, Kate Saenko, Moritz Hardt, and Sergey Levine (eds.), \emph{Advances in Neural Information Processing Systems 36: Annual Conference on Neural Information Processing Systems 2023, NeurIPS 2023, New Orleans, LA, USA, December 10 - 16, 2023}, 2023.
\newblock URL \url{http://papers.nips.cc/paper\_files/paper/2023/hash/920f2dced7d32ab2ba2f1970bc306af6-Abstract-Datasets\_and\_Benchmarks.html}.

\bibitem[Ding et~al.(2024)Ding, Wang, Ahmad, Ding, Tan, Jain, Ramanathan, Nallapati, Bhatia, Roth, et~al.]{cceval}
Yangruibo Ding, Zijian Wang, Wasi Ahmad, Hantian Ding, Ming Tan, Nihal Jain, Murali~Krishna Ramanathan, Ramesh Nallapati, Parminder Bhatia, Dan Roth, et~al.
\newblock Crosscodeeval: A diverse and multilingual benchmark for cross-file code completion.
\newblock \emph{Advances in Neural Information Processing Systems}, 36, 2024.

\bibitem[Fried et~al.(2023)Fried, Aghajanyan, Lin, Wang, Wallace, Shi, Zhong, Yih, Zettlemoyer, and Lewis]{fried2022incoder}
Daniel Fried, Armen Aghajanyan, Jessy Lin, Sida Wang, Eric Wallace, Freda Shi, Ruiqi Zhong, Scott Yih, Luke Zettlemoyer, and Mike Lewis.
\newblock Incoder: A generative model for code infilling and synthesis.
\newblock In \emph{The Eleventh International Conference on Learning Representations}, 2023.
\newblock URL \url{https://openreview.net/forum?id=hQwb-lbM6EL}.

\bibitem[Guo et~al.(2024{\natexlab{a}})Guo, Zhu, Yang, Xie, Dong, Zhang, Chen, Bi, Wu, Li, et~al.]{deepseek_coder}
Daya Guo, Qihao Zhu, Dejian Yang, Zhenda Xie, Kai Dong, Wentao Zhang, Guanting Chen, Xiao Bi, Y~Wu, YK~Li, et~al.
\newblock Deepseek-coder: When the large language model meets programming--the rise of code intelligence.
\newblock \emph{arXiv preprint arXiv:2401.14196}, 2024{\natexlab{a}}.
\newblock URL \url{https://arxiv.org/abs/2401.14196}.

\bibitem[Guo et~al.(2024{\natexlab{b}})Guo, Zhu, Yang, Xie, Dong, Zhang, Chen, Bi, Wu, Li, et~al.]{guo2024deepseek}
Daya Guo, Qihao Zhu, Dejian Yang, Zhenda Xie, Kai Dong, Wentao Zhang, Guanting Chen, Xiao Bi, Y~Wu, YK~Li, et~al.
\newblock Deepseek-coder: When the large language model meets programming--the rise of code intelligence.
\newblock \emph{arXiv preprint arXiv:2401.14196}, 2024{\natexlab{b}}.

\bibitem[Hui et~al.(2024)Hui, Yang, Cui, Yang, Liu, Zhang, Liu, Zhang, Yu, Dang, et~al.]{qwen25coder}
Binyuan Hui, Jian Yang, Zeyu Cui, Jiaxi Yang, Dayiheng Liu, Lei Zhang, Tianyu Liu, Jiajun Zhang, Bowen Yu, Kai Dang, et~al.
\newblock Qwen2. 5-coder technical report.
\newblock \emph{arXiv preprint arXiv:2409.12186}, 2024.

\bibitem[Jain et~al.(2024)Jain, Han, Gu, Li, Yan, Zhang, Wang, Solar-Lezama, Sen, and Stoica]{livecodebench}
Naman Jain, King Han, Alex Gu, Wen-Ding Li, Fanjia Yan, Tianjun Zhang, Sida Wang, Armando Solar-Lezama, Koushik Sen, and Ion Stoica.
\newblock Livecodebench: Holistic and contamination free evaluation of large language models for code.
\newblock \emph{arXiv preprint arXiv:2403.07974}, 2024.

\bibitem[Kocetkov et~al.(2022)Kocetkov, Li, Allal, Li, Mou, Ferrandis, Jernite, Mitchell, Hughes, Wolf, et~al.]{kocetkov2022stack}
Denis Kocetkov, Raymond Li, Loubna~Ben Allal, Jia Li, Chenghao Mou, Carlos~Mu{\~n}oz Ferrandis, Yacine Jernite, Margaret Mitchell, Sean Hughes, Thomas Wolf, et~al.
\newblock The stack: 3 tb of permissively licensed source code.
\newblock \emph{arXiv preprint arXiv:2211.15533}, 2022.
\newblock URL \url{https://arxiv.org/abs/2211.15533}.

\bibitem[Le et~al.(2022)Le, Wang, Gotmare, Savarese, and Hoi]{Le2022CodeRLMC}
Hung Le, Yue Wang, Akhilesh~Deepak Gotmare, Silvio Savarese, and Steven C.~H. Hoi.
\newblock Coderl: Mastering code generation through pretrained models and deep reinforcement learning.
\newblock \emph{ArXiv}, abs/2207.01780, 2022.
\newblock URL \url{https://api.semanticscholar.org/CorpusID:250280117}.

\bibitem[Li et~al.(2023)Li, Allal, Zi, Muennighoff, Kocetkov, Mou, Marone, Akiki, Li, Chim, et~al.]{li2023starcoder}
Raymond Li, Loubna~Ben Allal, Yangtian Zi, Niklas Muennighoff, Denis Kocetkov, Chenghao Mou, Marc Marone, Christopher Akiki, Jia Li, Jenny Chim, et~al.
\newblock Starcoder: may the source be with you!
\newblock \emph{arXiv preprint arXiv:2305.06161}, 2023.
\newblock URL \url{https://arxiv.org/abs/2305.06161}.

\bibitem[Li et~al.(2022)Li, Choi, Chung, Kushman, Schrittwieser, Leblond, Eccles, Keeling, Gimeno, Lago, et~al.]{li2022competition}
Yujia Li, David Choi, Junyoung Chung, Nate Kushman, Julian Schrittwieser, R{\'e}mi Leblond, Tom Eccles, James Keeling, Felix Gimeno, Agustin~Dal Lago, et~al.
\newblock Competition-level code generation with alphacode.
\newblock \emph{ArXiv preprint}, abs/2203.07814, 2022.
\newblock URL \url{https://arxiv.org/abs/2203.07814}.

\bibitem[Liao et~al.(2023)Liao, Pan, Huang, Ren, Xing, Jin, and Li]{liao2023a3codgen}
Dianshu Liao, Shidong Pan, Qing Huang, Xiaoxue Ren, Zhenchang Xing, Huan Jin, and Qinying Li.
\newblock Context-aware code generation framework for code repositories: Local, global, and third-party library awareness.
\newblock 2023.

\bibitem[Liu et~al.(2023{\natexlab{a}})Liu, Xu, and McAuley]{repobench}
Tianyang Liu, Canwen Xu, and Julian McAuley.
\newblock Repobench: Benchmarking repository-level code auto-completion systems.
\newblock \emph{arXiv preprint arXiv:2306.03091}, 2023{\natexlab{a}}.

\bibitem[Liu et~al.(2023{\natexlab{b}})Liu, Xu, and McAuley]{liu2023repobench}
Tianyang Liu, Canwen Xu, and Julian~J. McAuley.
\newblock Repobench: Benchmarking repository-level code auto-completion systems.
\newblock abs/2306.03091, 2023{\natexlab{b}}.
\newblock \doi{10.48550/ARXIV.2306.03091}.
\newblock URL \url{https://doi.org/10.48550/arXiv.2306.03091}.

\bibitem[Lozhkov et~al.(2024)Lozhkov, Li, Allal, Cassano, Lamy-Poirier, Tazi, Tang, Pykhtar, Liu, Wei, et~al.]{lozhkov2024starcoder2}
Anton Lozhkov, Raymond Li, Loubna~Ben Allal, Federico Cassano, Joel Lamy-Poirier, Nouamane Tazi, Ao~Tang, Dmytro Pykhtar, Jiawei Liu, Yuxiang Wei, et~al.
\newblock Starcoder 2 and the stack v2: The next generation.
\newblock 2024.

\bibitem[Nijkamp et~al.(2023)Nijkamp, Pang, Hayashi, Tu, Wang, Zhou, Savarese, and Xiong]{nijkamp2022codegen}
Erik Nijkamp, Bo~Pang, Hiroaki Hayashi, Lifu Tu, Huan Wang, Yingbo Zhou, Silvio Savarese, and Caiming Xiong.
\newblock Codegen: An open large language model for code with multi-turn program synthesis.
\newblock In \emph{International Conference on Learning Representations}, 2023.
\newblock URL \url{https://openreview.net/forum?id=iaYcJKpY2B_}.

\bibitem[Niu et~al.(2022)Niu, Li, Ng, Ge, Huang, and Luo]{Niu2022SPTCodeSP}
Changan Niu, Chuanyi Li, Vincent Ng, Jidong Ge, LiGuo Huang, and Bin Luo.
\newblock Spt-code: Sequence-to-sequence pre-training for learning source code representations.
\newblock \emph{2022 IEEE/ACM 44th International Conference on Software Engineering (ICSE)}, pp.\  01--13, 2022.
\newblock URL \url{https://api.semanticscholar.org/CorpusID:246077487}.

\bibitem[Pei et~al.(2023)Pei, Zhao, Lausen, Zha, and Karypis]{pei2023better}
Hengzhi Pei, Jinman Zhao, Leonard Lausen, Sheng Zha, and George Karypis.
\newblock Better context makes better code language models: A case study on function call argument completion.
\newblock In \emph{Proceedings of the Thirty-Seventh AAAI Conference on Artificial Intelligence and Thirty-Fifth Conference on Innovative Applications of Artificial Intelligence and Thirteenth Symposium on Educational Advances in Artificial Intelligence}, AAAI'23/IAAI'23/EAAI'23. AAAI Press, 2023.
\newblock ISBN 978-1-57735-880-0.
\newblock \doi{10.1609/aaai.v37i4.25653}.
\newblock URL \url{https://doi.org/10.1609/aaai.v37i4.25653}.

\bibitem[Phan et~al.(2024)Phan, Phan, Nguyen, and Bui]{Phan2024RepoHyperSO}
Huy~Nhat Phan, Hoang~N. Phan, Tien~N. Nguyen, and Nghi D.~Q. Bui.
\newblock Repohyper: Search-expand-refine on semantic graphs for repository-level code completion.
\newblock 2024.
\newblock URL \url{https://api.semanticscholar.org/CorpusID:271860296}.

\bibitem[Ren et~al.(2020)Ren, Guo, Lu, Zhou, Liu, Tang, Sundaresan, Zhou, Blanco, and Ma]{ren2020codebleu}
Shuo Ren, Daya Guo, Shuai Lu, Long Zhou, Shujie Liu, Duyu Tang, Neel Sundaresan, Ming Zhou, Ambrosio Blanco, and Shuai Ma.
\newblock Codebleu: a method for automatic evaluation of code synthesis.
\newblock \emph{arXiv preprint arXiv:2009.10297}, 2020.

\bibitem[Roziere et~al.(2023)Roziere, Gehring, Gloeckle, Sootla, Gat, Tan, Adi, Liu, Remez, Rapin, et~al.]{codellama}
Baptiste Roziere, Jonas Gehring, Fabian Gloeckle, Sten Sootla, Itai Gat, Xiaoqing~Ellen Tan, Yossi Adi, Jingyu Liu, Tal Remez, J{\'e}r{\'e}my Rapin, et~al.
\newblock Code llama: Open foundation models for code.
\newblock 2023.

\bibitem[Shrivastava et~al.(2023{\natexlab{a}})Shrivastava, Kocetkov, de~Vries, Bahdanau, and Scholak]{shrivastava2023repofusion}
Disha Shrivastava, Denis Kocetkov, Harm de~Vries, Dzmitry Bahdanau, and Torsten Scholak.
\newblock Repofusion: Training code models to understand your repository.
\newblock \emph{arXiv preprint arXiv:2306.10998}, 2023{\natexlab{a}}.

\bibitem[Shrivastava et~al.(2023{\natexlab{b}})Shrivastava, Larochelle, and Tarlow]{shrivastava2022repository}
Disha Shrivastava, Hugo Larochelle, and Daniel Tarlow.
\newblock Repository-level prompt generation for large language models of code.
\newblock In Andreas Krause, Emma Brunskill, Kyunghyun Cho, Barbara Engelhardt, Sivan Sabato, and Jonathan Scarlett (eds.), \emph{Proceedings of the 40th International Conference on Machine Learning}, volume 202 of \emph{Proceedings of Machine Learning Research}, pp.\  31693--31715. PMLR, 23--29 Jul 2023{\natexlab{b}}.
\newblock URL \url{https://proceedings.mlr.press/v202/shrivastava23a.html}.

\bibitem[Su et~al.(2024)Su, Ahmed, Lu, Pan, Bo, and Liu]{su2024roformer}
Jianlin Su, Murtadha Ahmed, Yu~Lu, Shengfeng Pan, Wen Bo, and Yunfeng Liu.
\newblock Roformer: Enhanced transformer with rotary position embedding.
\newblock 568:\penalty0 127063, 2024.

\bibitem[Sun et~al.(2024)Sun, Chai, Jian~Yang, Guo, Liu, Wang, Yang, and Li]{unicoder}
Tao Sun, Linzheng Chai, Yuwei~Yin Jian~Yang, Hongcheng Guo, Jiaheng Liu, Bing Wang, Liqun Yang, and Zhoujun Li.
\newblock Unicoder: Scaling code large language model via universal code.
\newblock \emph{ACL}, 2024.

\bibitem[Takerngsaksiri et~al.(2024)Takerngsaksiri, Tantithamthavorn, and Li]{10.1016/j.infsof.2023.107336}
Wannita Takerngsaksiri, Chakkrit Tantithamthavorn, and Yuan-Fang Li.
\newblock Syntax-aware on-the-fly code completion.
\newblock \emph{Inf. Softw. Technol.}, 165\penalty0 (C), January 2024.
\newblock ISSN 0950-5849.

\bibitem[Touvron et~al.(2023)Touvron, Lavril, Izacard, Martinet, Lachaux, Lacroix, Rozi{\`e}re, Goyal, Hambro, Azhar, et~al.]{touvron2023llama}
Hugo Touvron, Thibaut Lavril, Gautier Izacard, Xavier Martinet, Marie-Anne Lachaux, Timoth{\'e}e Lacroix, Baptiste Rozi{\`e}re, Naman Goyal, Eric Hambro, Faisal Azhar, et~al.
\newblock Llama: Open and efficient foundation language models.
\newblock 2023.

\bibitem[Wang et~al.(2021)Wang, Wang, Joty, and Hoi]{wang-etal-2021-codet5}
Yue Wang, Weishi Wang, Shafiq Joty, and Steven~C.H. Hoi.
\newblock {C}ode{T}5: Identifier-aware unified pre-trained encoder-decoder models for code understanding and generation.
\newblock In \emph{Proceedings of the 2021 Conference on Empirical Methods in Natural Language Processing}, pp.\  8696--8708, Online and Punta Cana, Dominican Republic, November 2021. Association for Computational Linguistics.
\newblock \doi{10.18653/v1/2021.emnlp-main.685}.
\newblock URL \url{https://aclanthology.org/2021.emnlp-main.685}.

\bibitem[Wang et~al.(2023)Wang, Le, Gotmare, Bui, Li, and Hoi]{wang2023codet5+}
Yue Wang, Hung Le, Akhilesh~Deepak Gotmare, Nghi~DQ Bui, Junnan Li, and Steven~CH Hoi.
\newblock Codet5+: Open code large language models for code understanding and generation.
\newblock 2023.

\bibitem[Xu et~al.(2022)Xu, Alon, Neubig, and Hellendoorn]{xu2022systematic}
Frank~F Xu, Uri Alon, Graham Neubig, and Vincent~Josua Hellendoorn.
\newblock A systematic evaluation of large language models of code.
\newblock In \emph{Proceedings of the 6th ACM SIGPLAN International Symposium on Machine Programming}, pp.\  1--10, 2022.

\bibitem[Yu et~al.(2024)Yu, Shen, Ran, Zhang, Zhang, Ma, Liang, Li, Wang, and Xie]{yu2024codereval}
Hao Yu, Bo~Shen, Dezhi Ran, Jiaxin Zhang, Qi~Zhang, Yuchi Ma, Guangtai Liang, Ying Li, Qianxiang Wang, and Tao Xie.
\newblock Codereval: A benchmark of pragmatic code generation with generative pre-trained models.
\newblock In \emph{Proceedings of the 46th IEEE/ACM International Conference on Software Engineering}, pp.\  1--12, 2024.

\bibitem[Zhang et~al.(2023)Zhang, Chen, Zhang, Liu, Zan, Mao, Lou, and Chen]{zhang2023repocoder}
Fengji Zhang, Bei Chen, Yue Zhang, Jin Liu, Daoguang Zan, Yi~Mao, Jian-Guang Lou, and Weizhu Chen.
\newblock Repocoder: Repository-level code completion through iterative retrieval and generation.
\newblock \emph{arXiv preprint arXiv:2303.12570}, 2023.
\newblock URL \url{https://arxiv.org/abs/2303.12570}.

\bibitem[Zhao et~al.(2024)Zhao, Hui, Howland, Nguyen, Zuo, Hu, Choquette-Choo, Shen, Kelley, tij Bansal, Vilnis, Wirth, Michel, Choy, Joshi, Kumar, Hashmi, Agrawal, Gong, Fine, Warkentin, Hartman, Ni, Korevec, Schaefer, and Huffman]{Zhao2024CodeGemmaOC}
CodeGemma Team~Heri Zhao, Jeffrey Hui, Joshua Howland, Nam Nguyen, Siqi Zuo, Andrea Hu, Christopher~A. Choquette-Choo, Jingyue Shen, Joe Kelley, Kshi tij Bansal, Luke Vilnis, Mateo Wirth, Paul Michel, Peter Choy, Pratik Joshi, Ravin Kumar, Sarmad Hashmi, Shubham Agrawal, Zhitao Gong, Jane Fine, Tris~Brian Warkentin, Ale~Jakse Hartman, Bin Ni, Kathy Korevec, Kelly Schaefer, and Scott Huffman.
\newblock Codegemma: Open code models based on gemma.
\newblock \emph{ArXiv}, abs/2406.11409, 2024.
\newblock URL \url{https://api.semanticscholar.org/CorpusID:270560319}.

\bibitem[Zheng et~al.(2023)Zheng, Xia, Zou, Dong, Wang, Xue, Wang, Shen, Wang, Li, Su, Yang, and Tang]{codegeex}
Qinkai Zheng, Xiao Xia, Xu~Zou, Yuxiao Dong, Shan Wang, Yufei Xue, Zihan Wang, Lei Shen, Andi Wang, Yang Li, Teng Su, Zhilin Yang, and Jie Tang.
\newblock Codegeex: {A} pre-trained model for code generation with multilingual evaluations on humaneval-x.
\newblock \emph{arXiv preprint arXiv:2303.17568}, abs/2303.17568, 2023.
\newblock \doi{10.48550/ARXIV.2303.17568}.
\newblock URL \url{https://doi.org/10.48550/arXiv.2303.17568}.

\end{thebibliography}
\bibliographystyle{conference}

\newpage
\appendix
\section{Appendix}




\begin{figure}[t]
    \centering
    \includegraphics[width=0.45\linewidth]{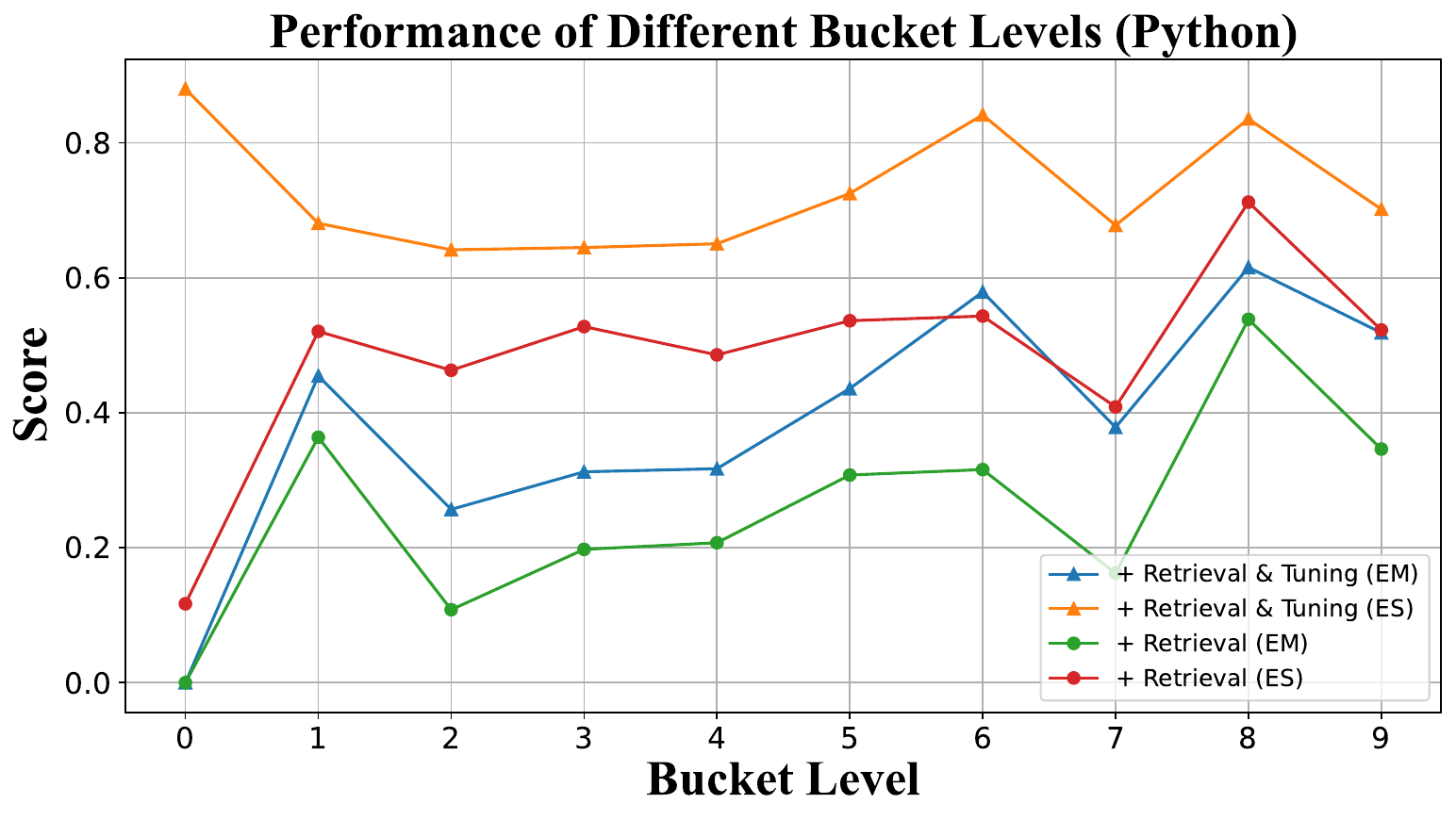}
            \includegraphics[width=0.45\linewidth]{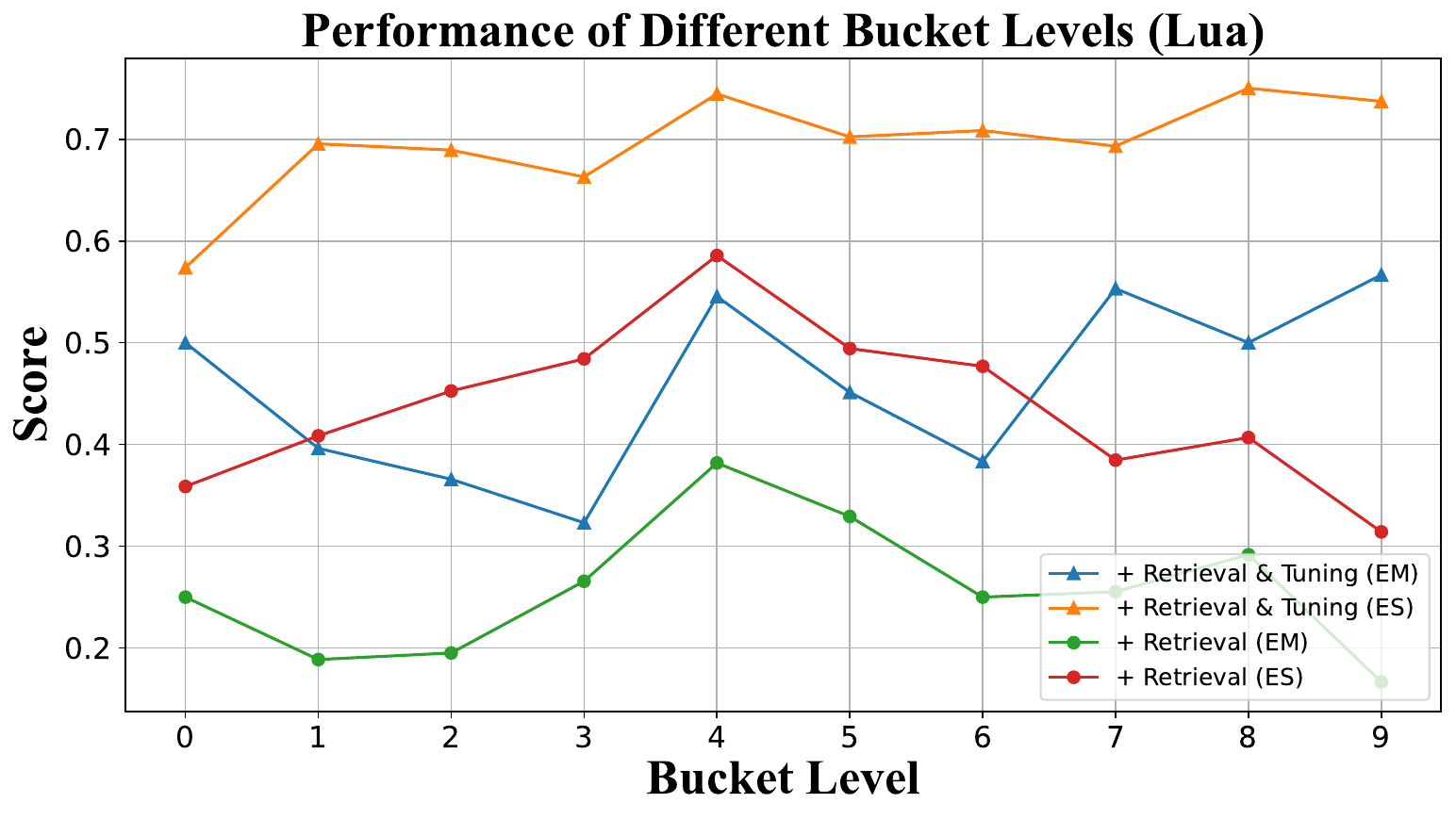}        \includegraphics[width=0.45\linewidth]{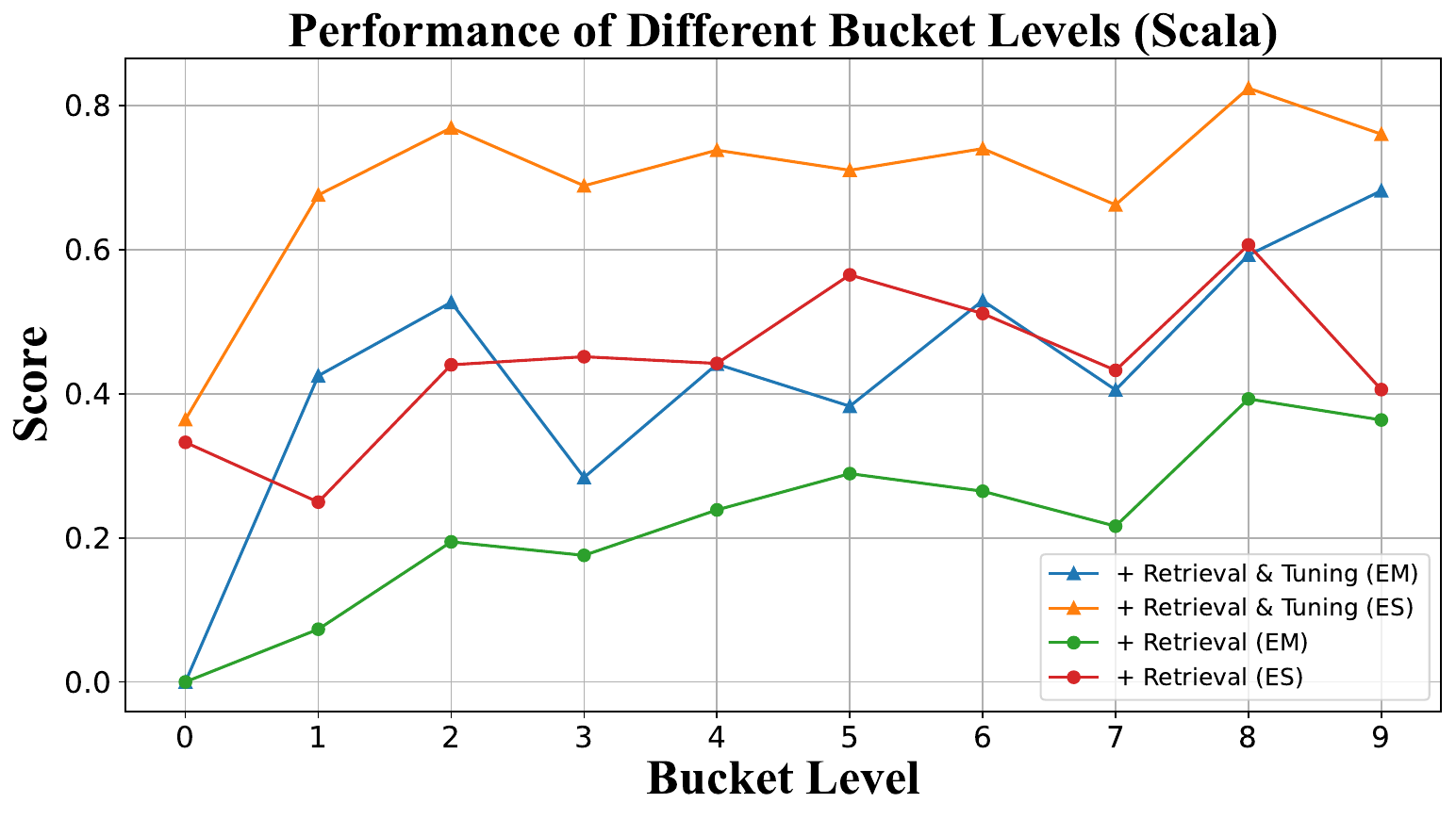}        \includegraphics[width=0.45\linewidth]{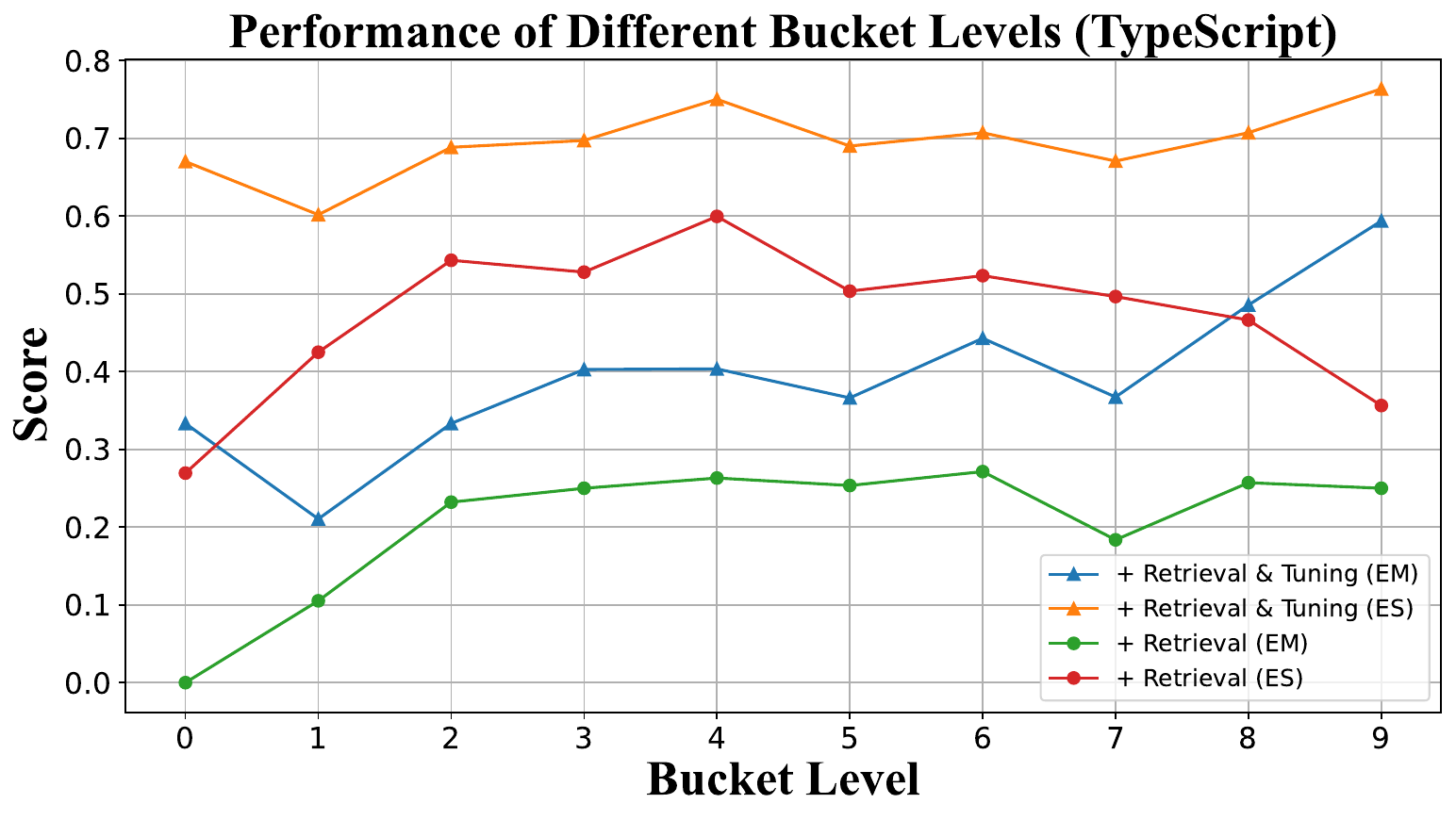}        \includegraphics[width=0.45\linewidth]{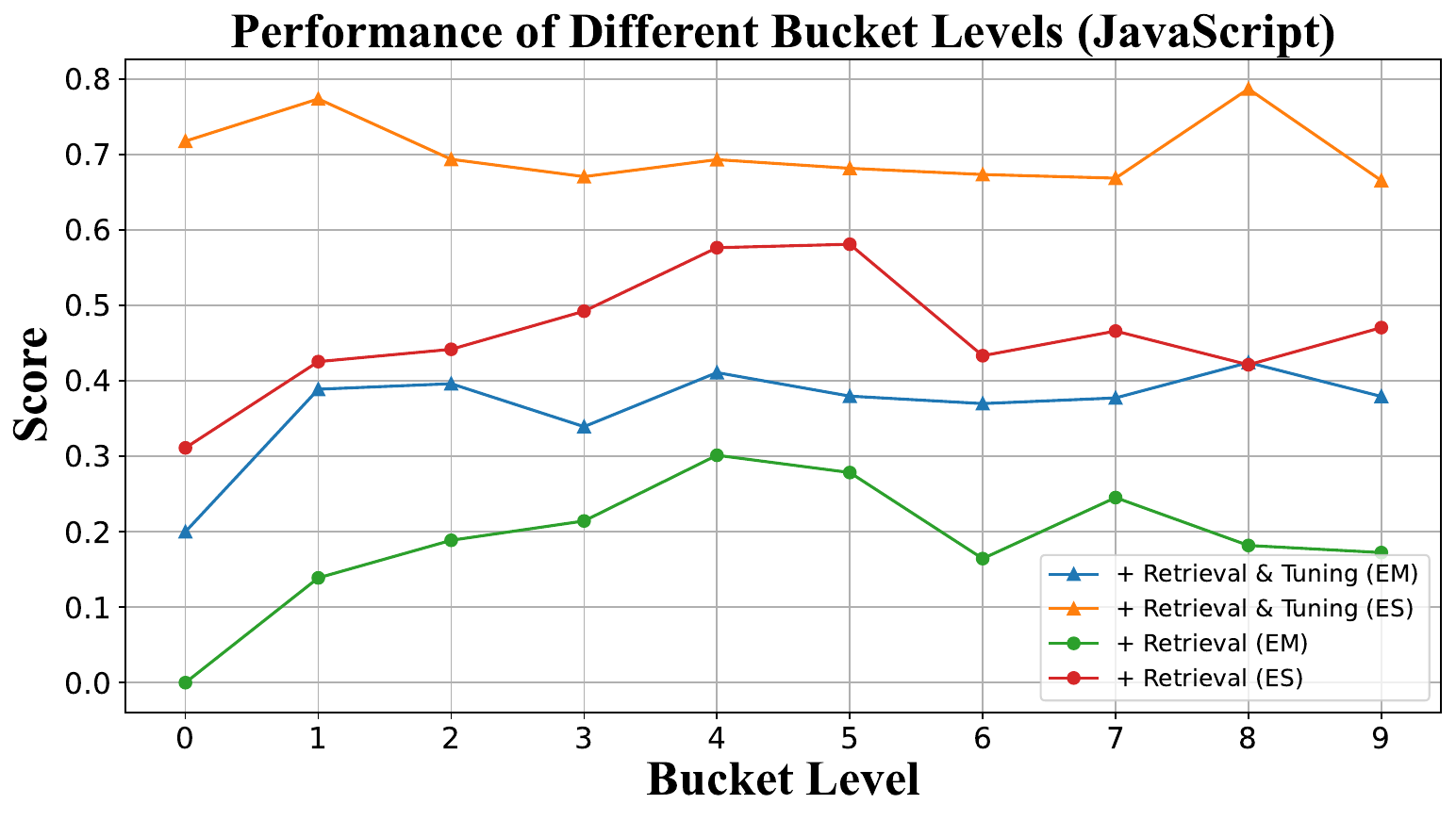}        \includegraphics[width=0.45\linewidth]{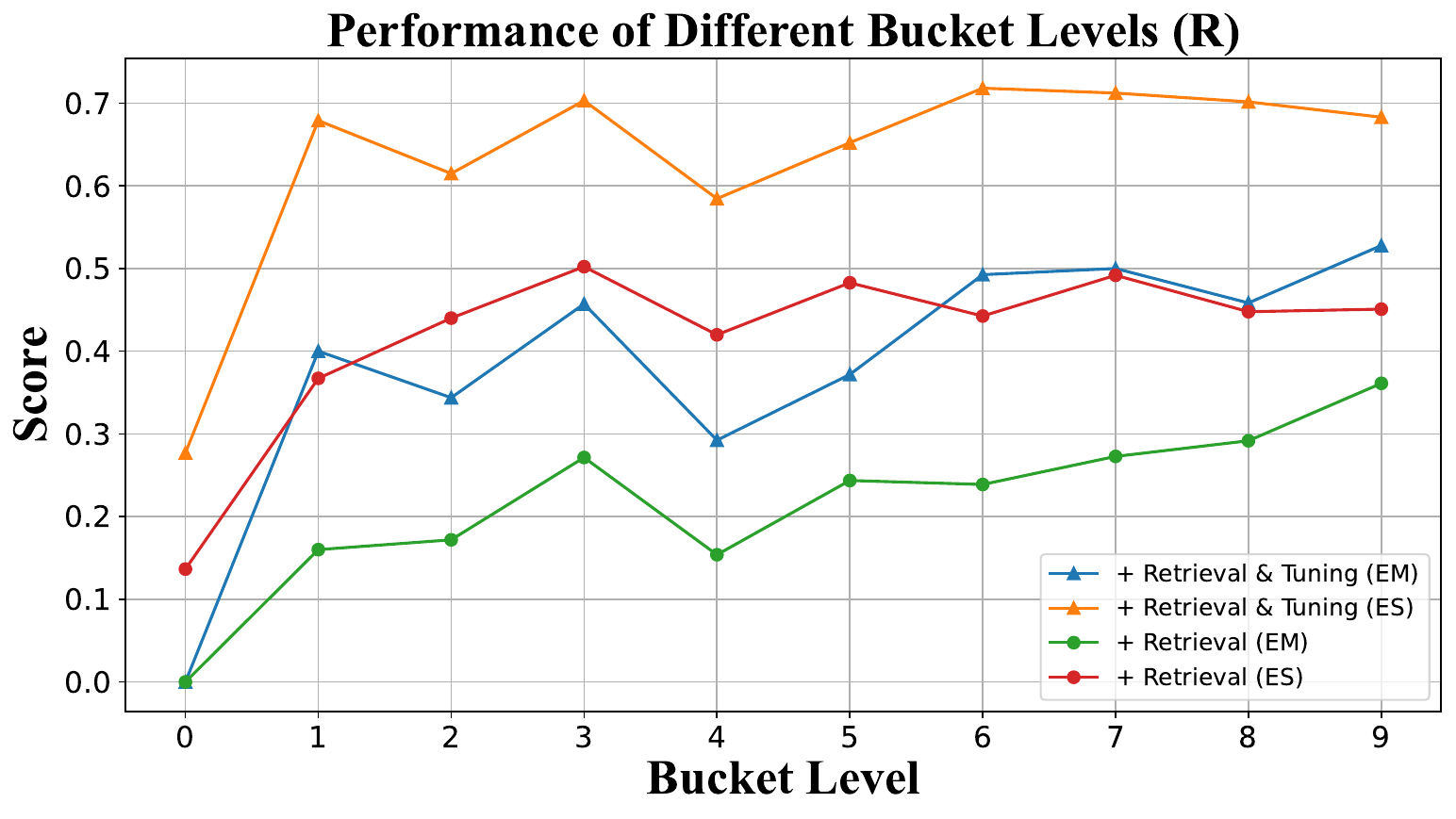}        \includegraphics[width=0.45\linewidth]{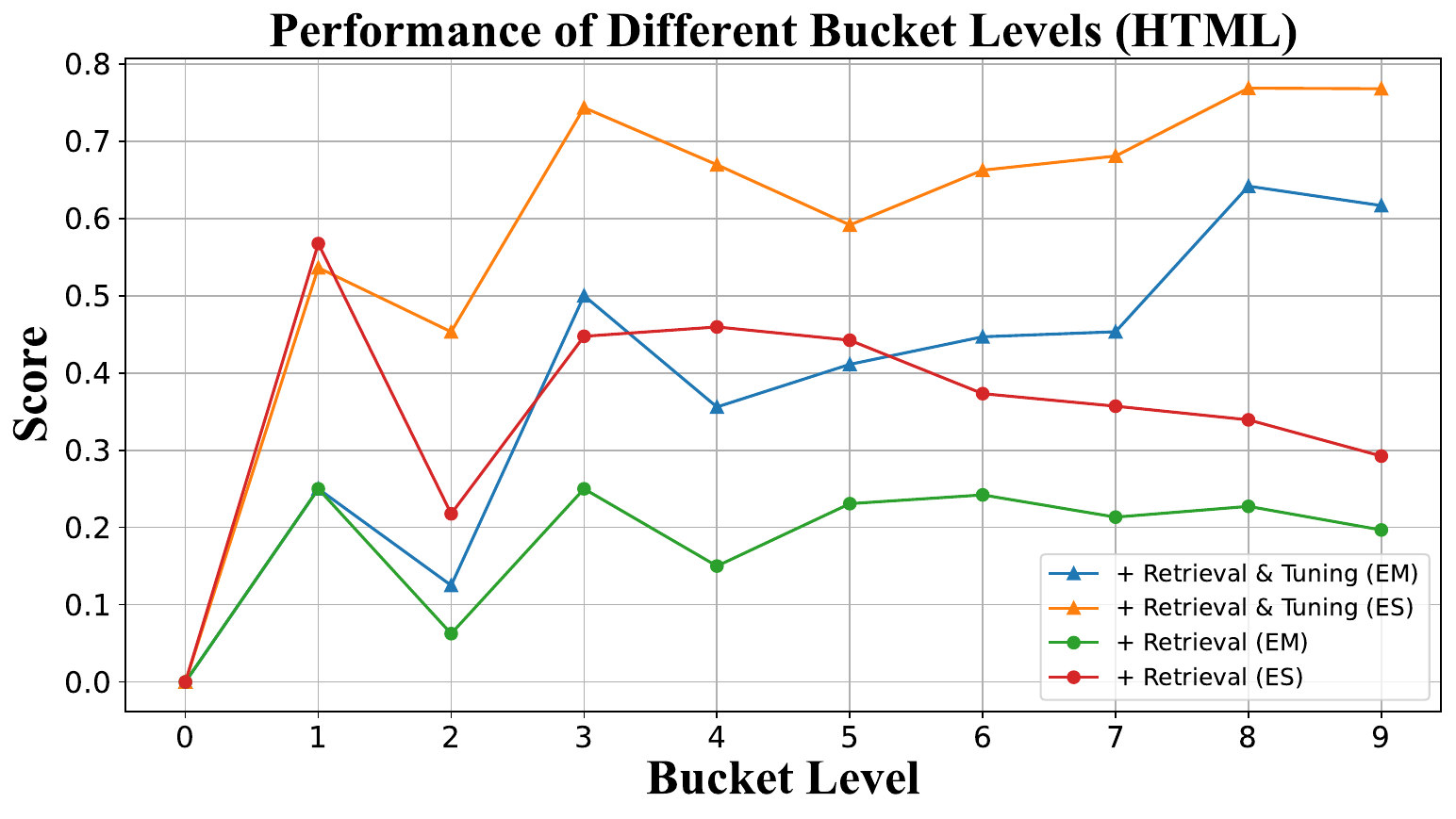}        \includegraphics[width=0.45\linewidth]{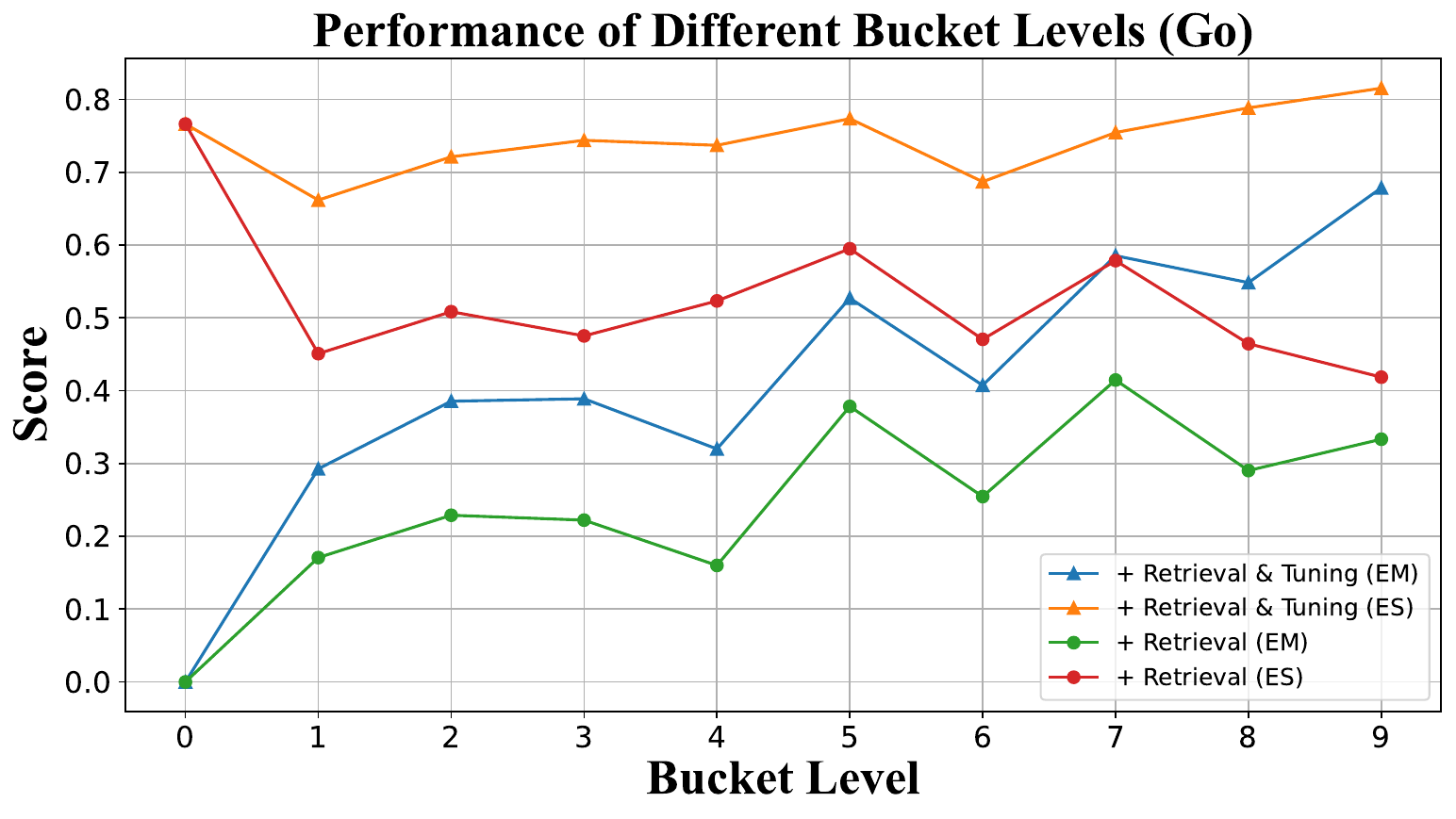}        \includegraphics[width=0.45\linewidth]{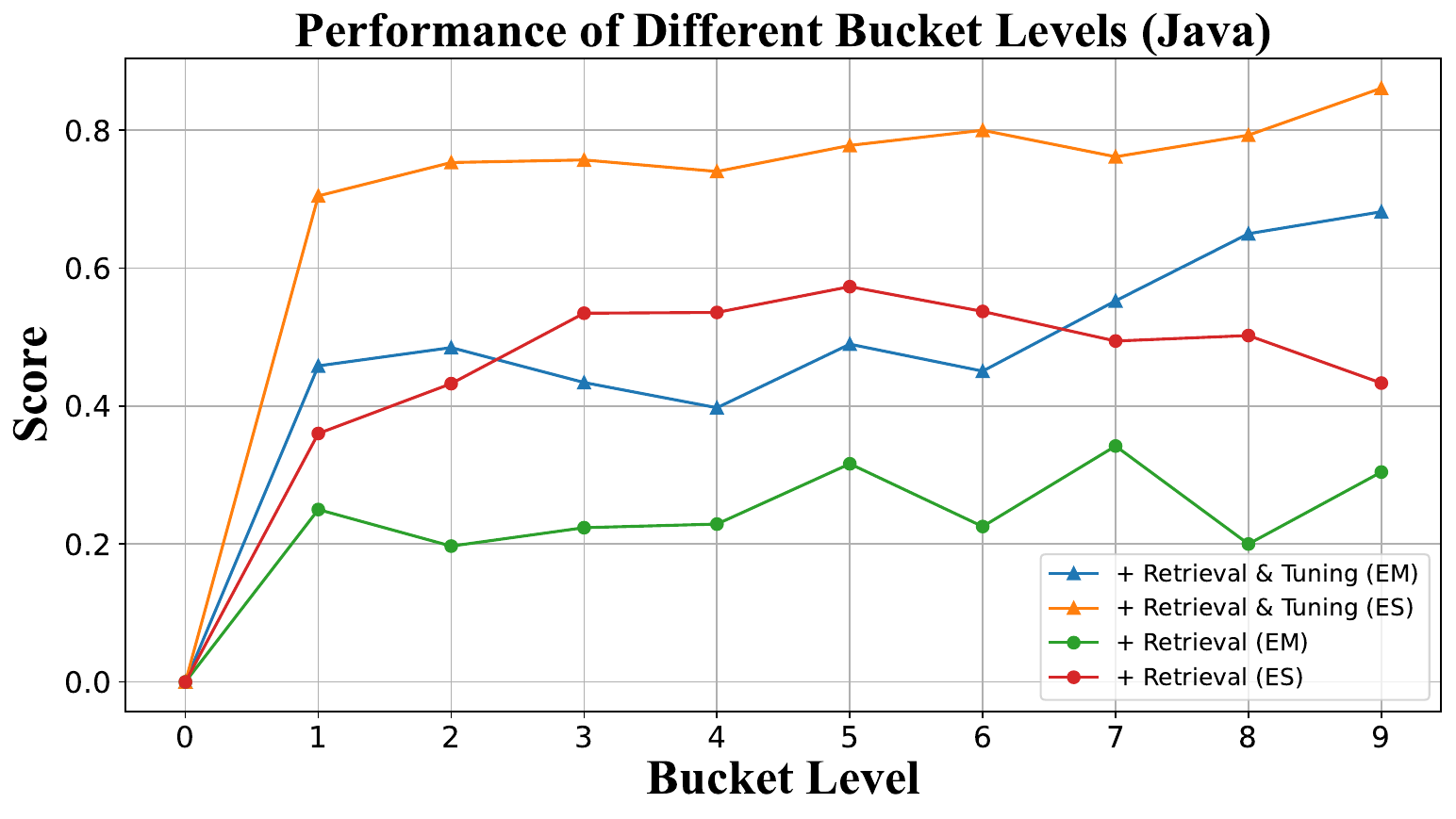}        \includegraphics[width=0.45\linewidth]{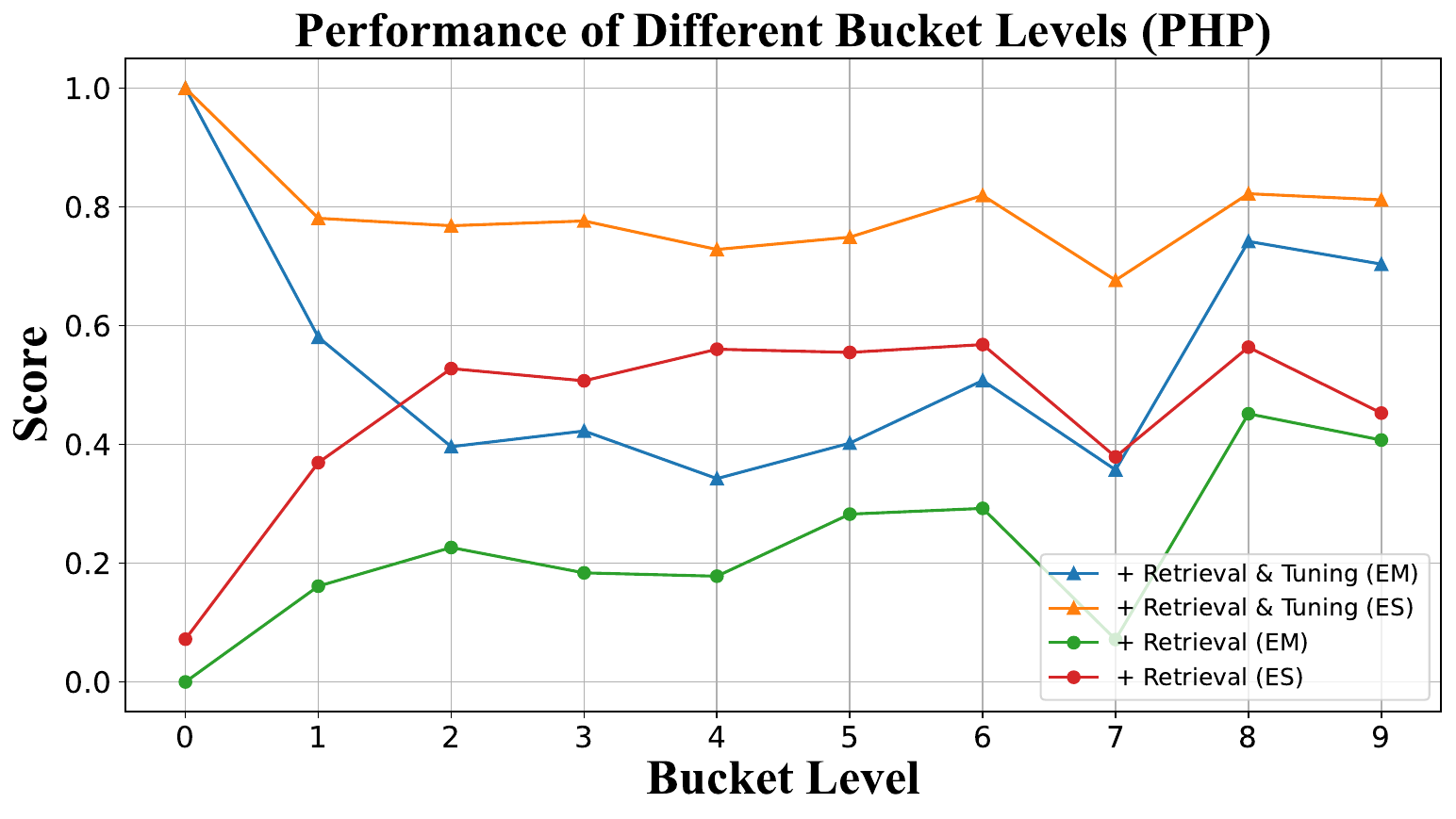}        \includegraphics[width=0.45\linewidth]{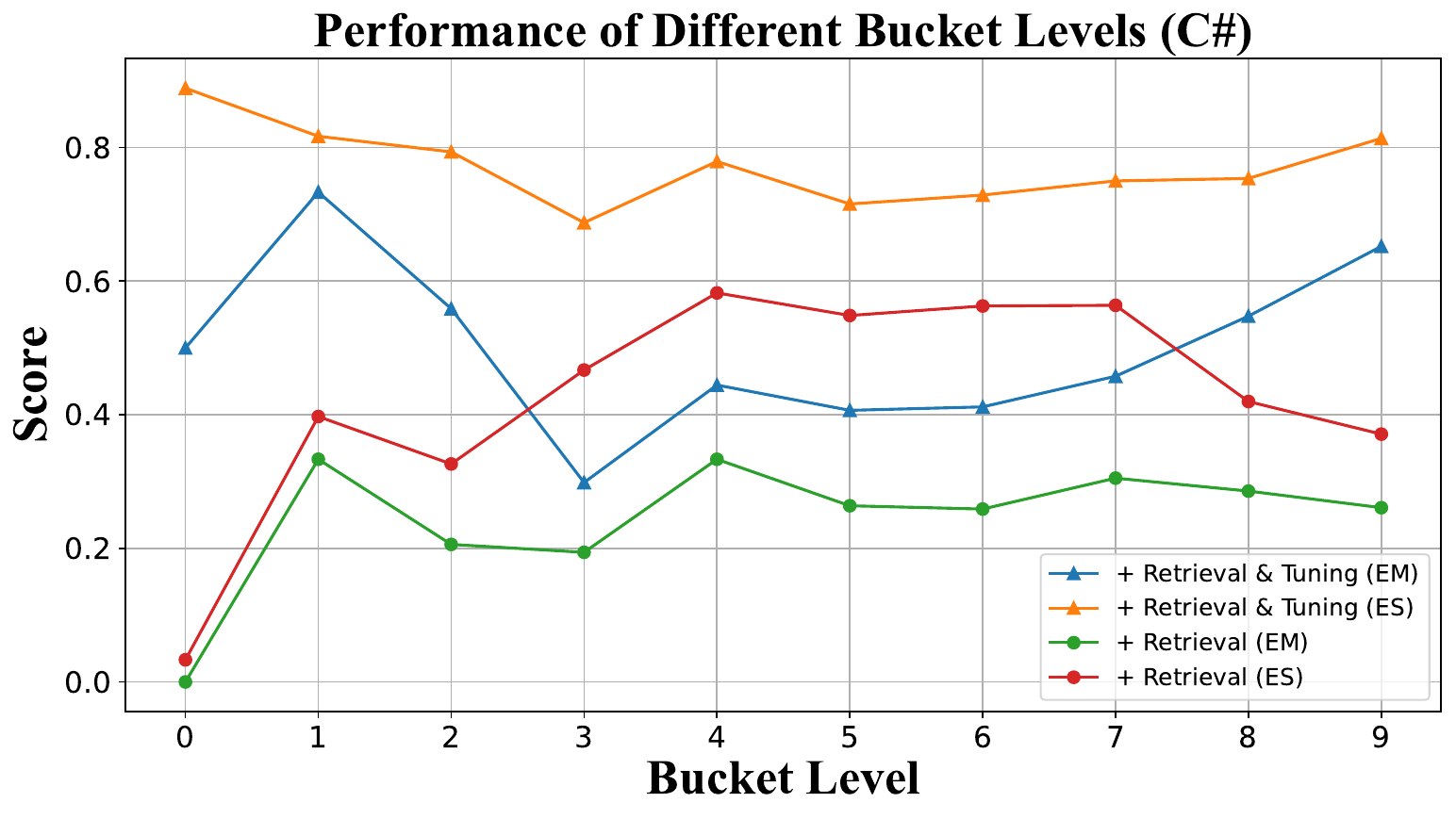}
        \includegraphics[width=0.45\linewidth]{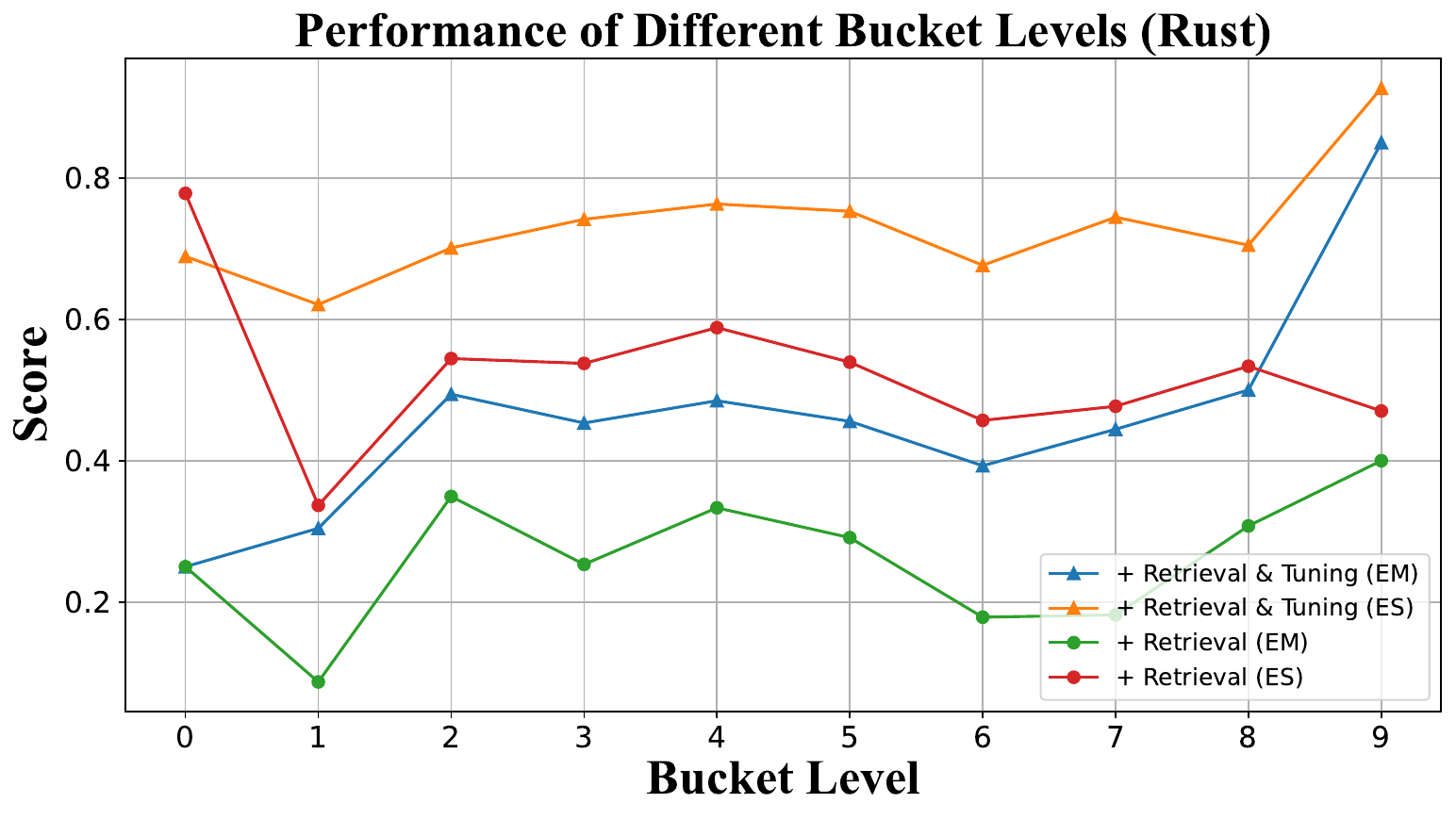}

    \caption{Effectiveness of different bucket levels based on StarCoder-7B for different languages.}
    \label{fig:level-python-12}
\end{figure} 
\begin{figure}[t]
    \centering
          \includegraphics[width=0.45\linewidth]{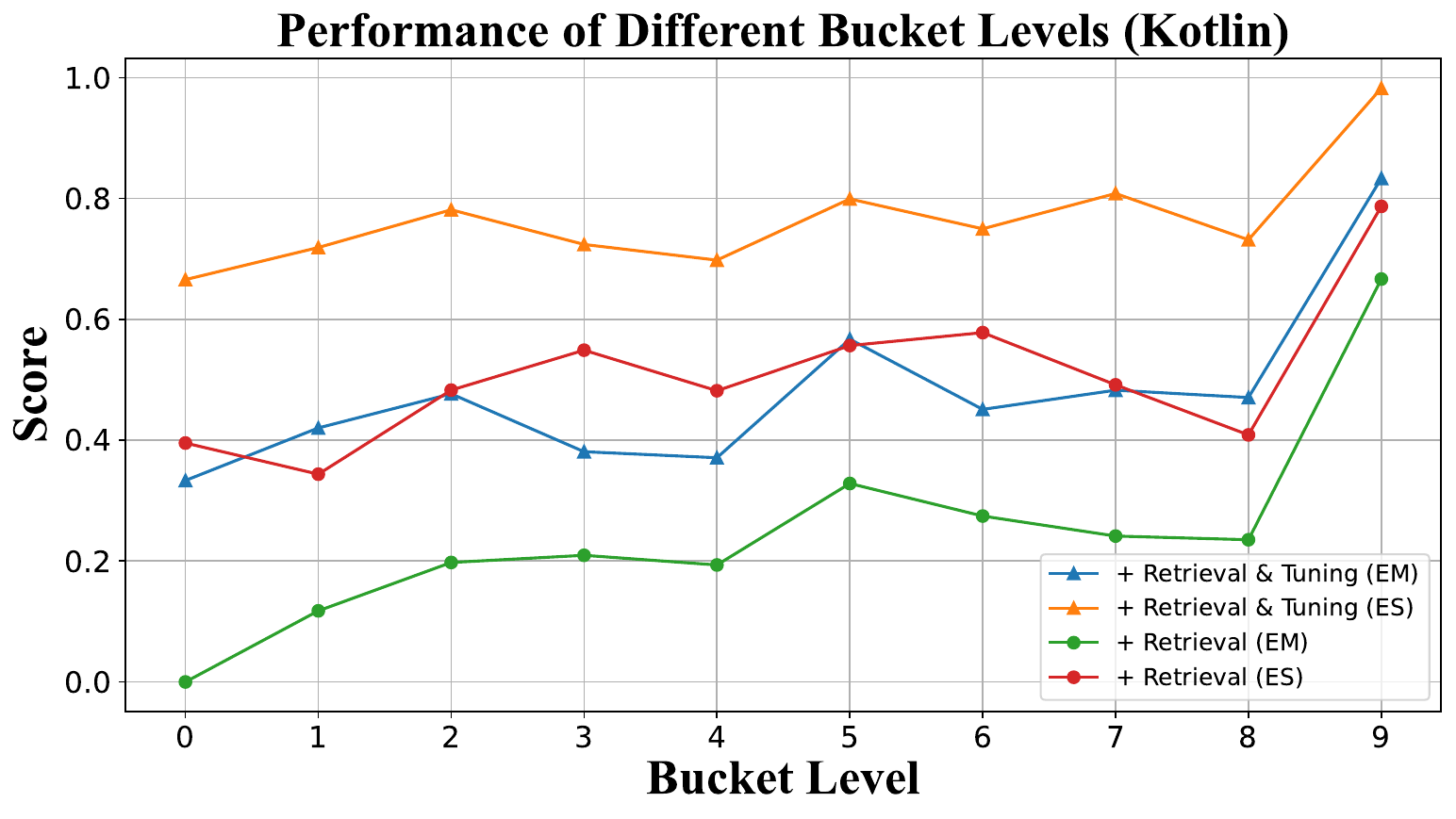}  \includegraphics[width=0.45\linewidth]{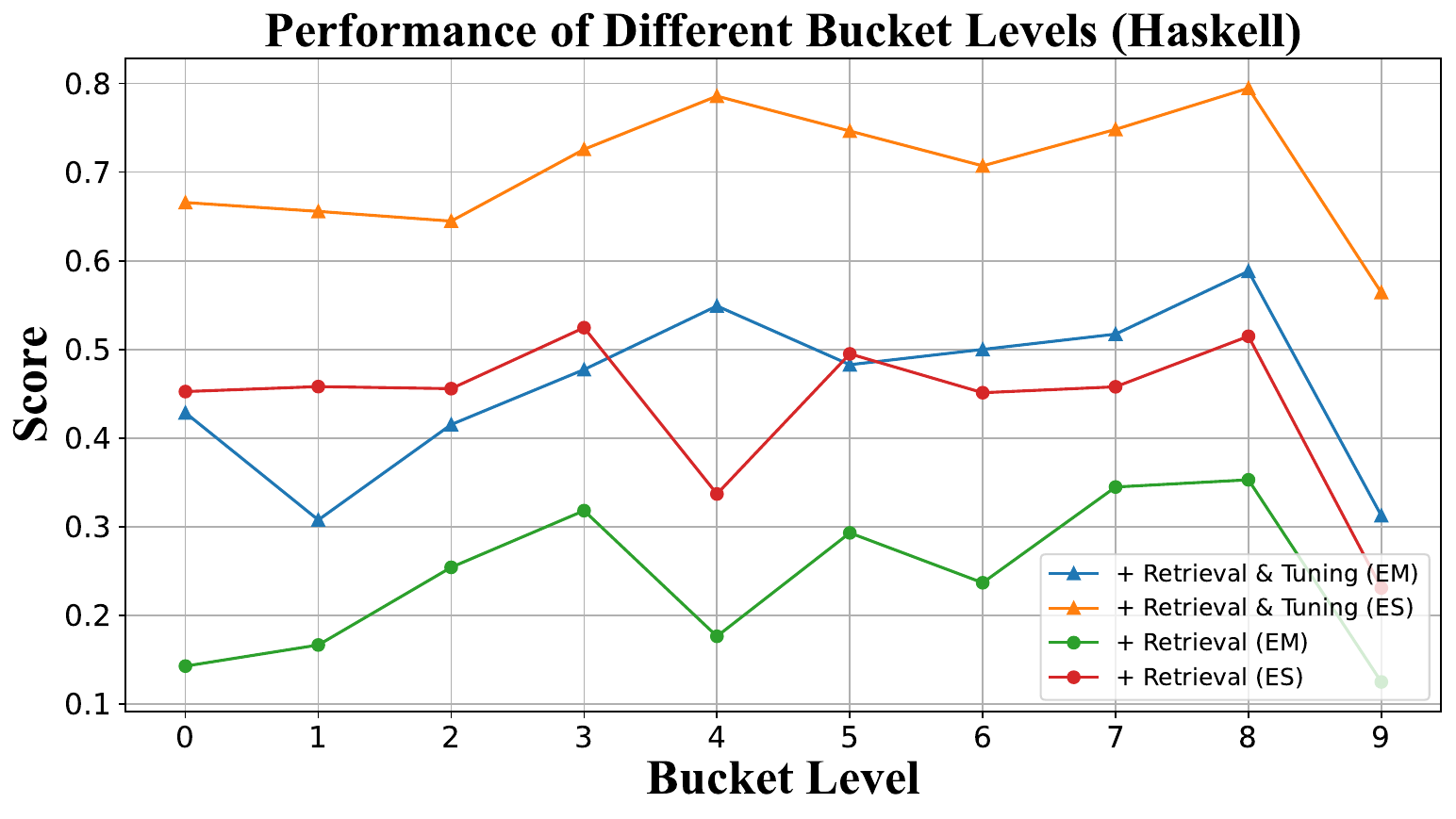}  \includegraphics[width=0.45\linewidth]{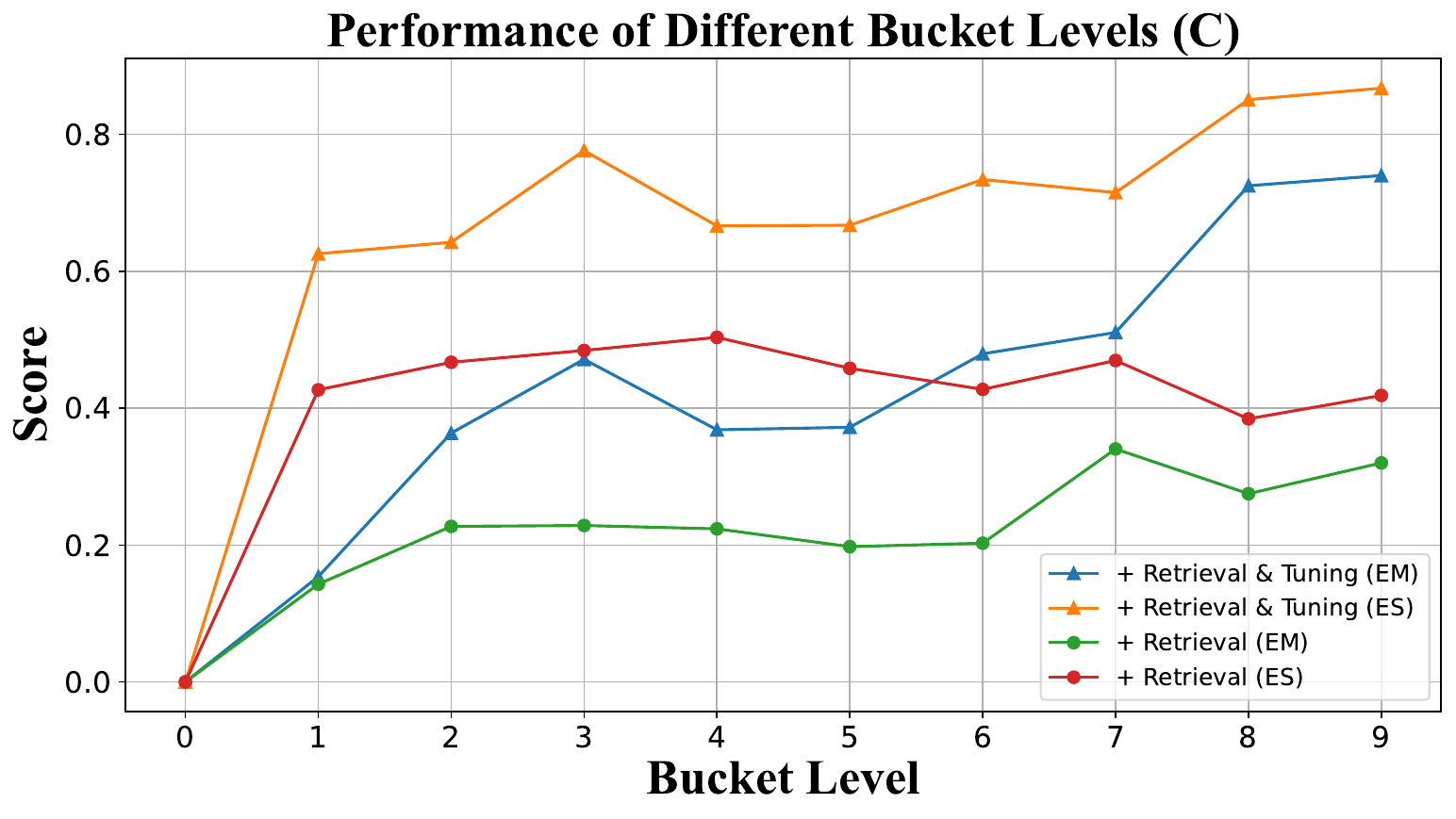}  \includegraphics[width=0.45\linewidth]{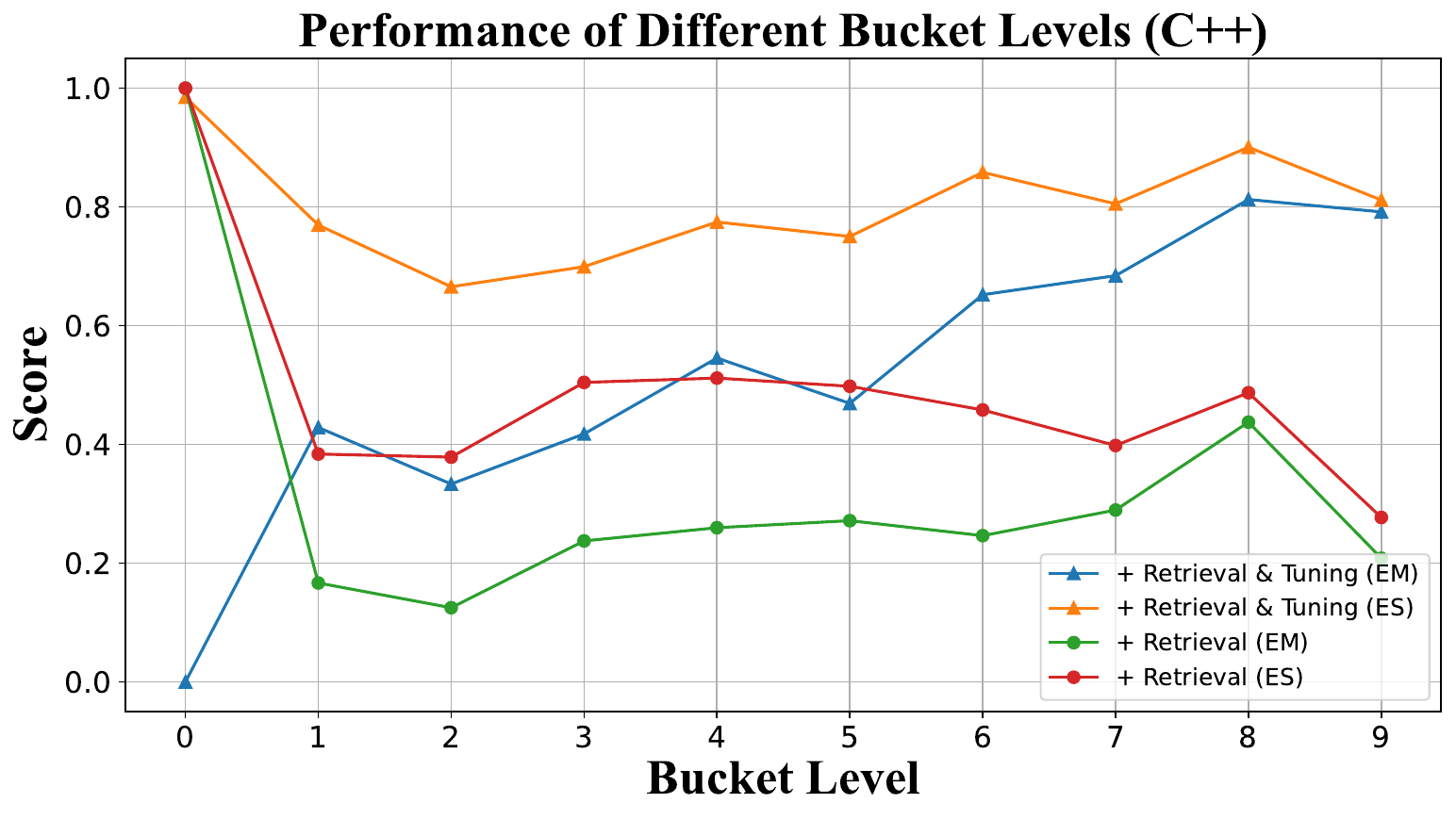}  \includegraphics[width=0.45\linewidth]{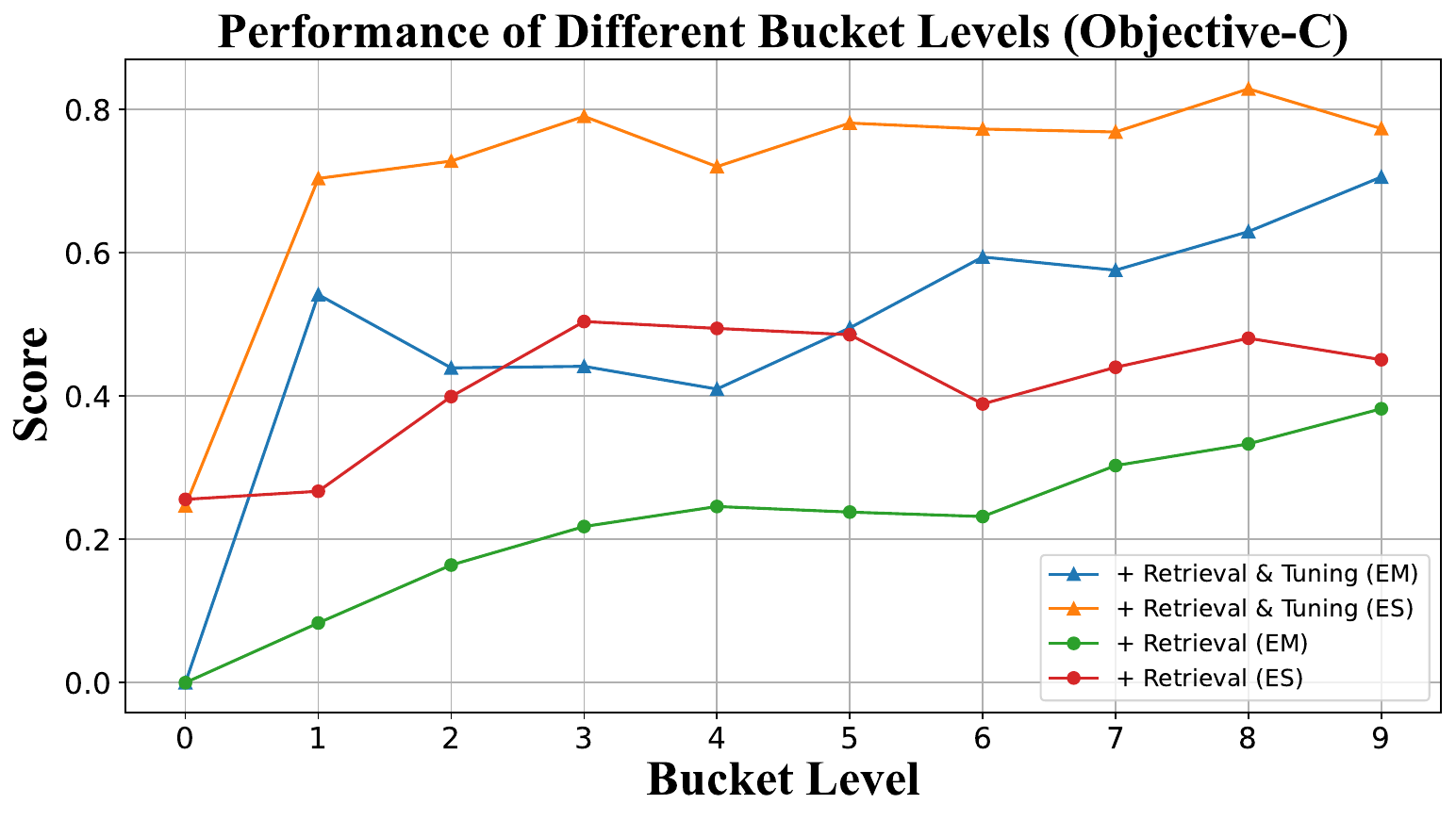}  \includegraphics[width=0.45\linewidth]{level_coder_Rust.pdf}
              \caption{Effectiveness of different bucket levels based on StarCoder-7B for different languages.}
    \label{fig:level-python-6}
\end{figure} 


\begin{figure}[t]
    \centering
    \includegraphics[width=1.0\linewidth]{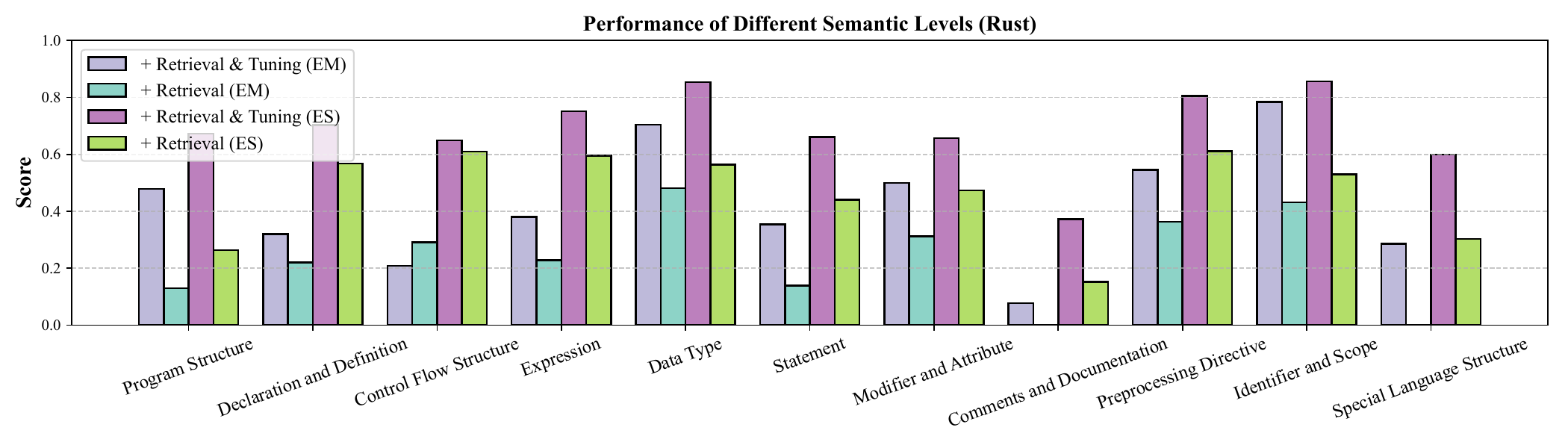}
        \includegraphics[width=1.0\linewidth]{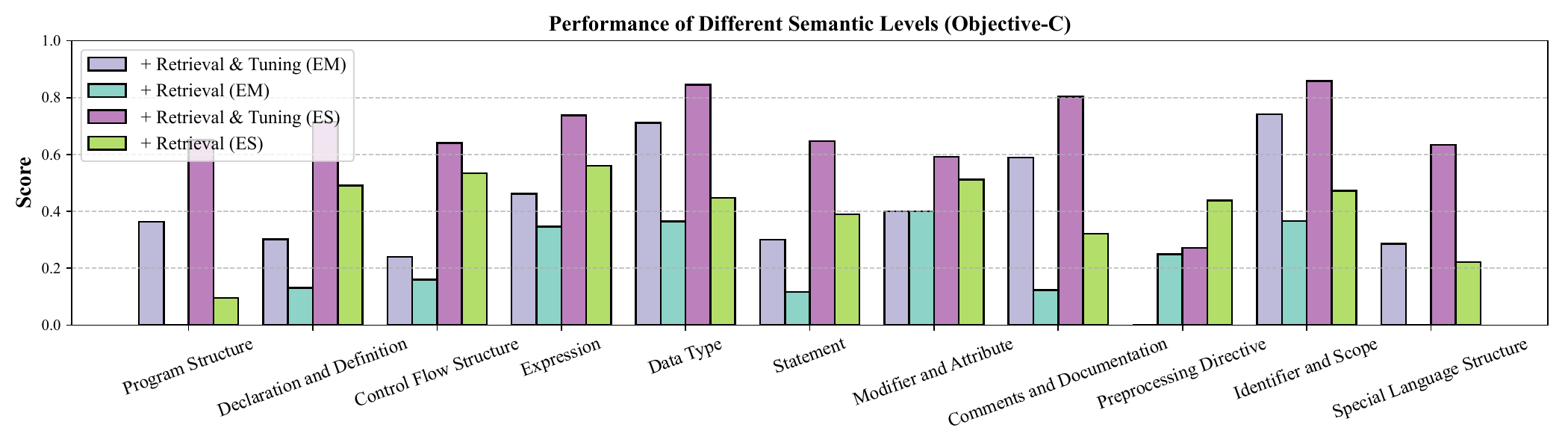}
                \includegraphics[width=1.0\linewidth]{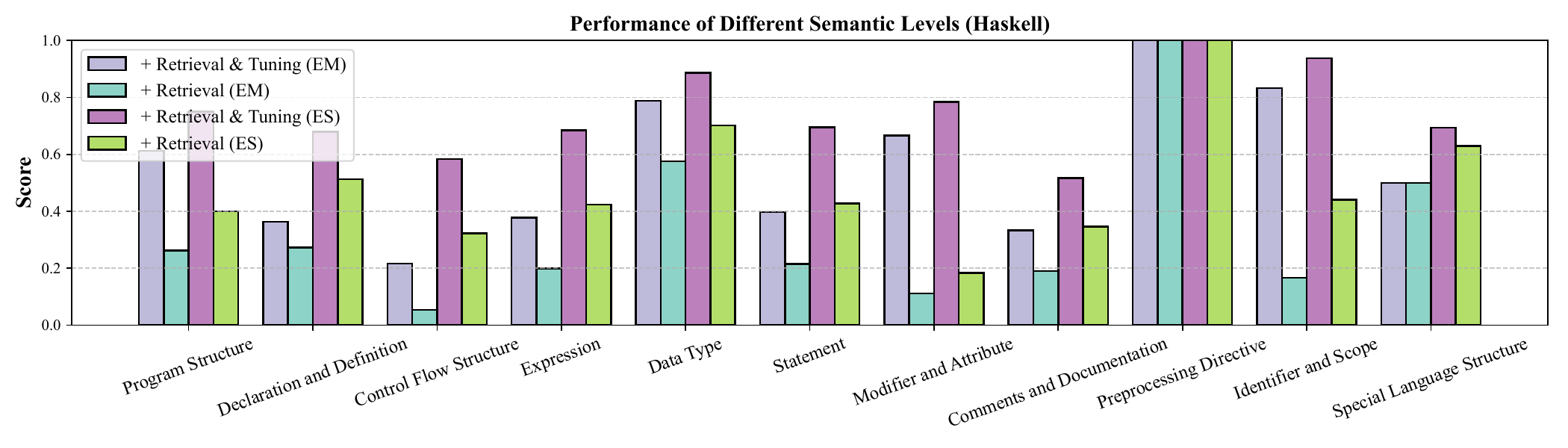}
    \caption{Effectiveness of different semantic levels based on StarCoder-7B.}
    \label{fig:semantic-levels}
\end{figure}


\begin{figure*}[!htb]
	\centering
	\subfigure[C]{\includegraphics[width=0.25\textwidth]{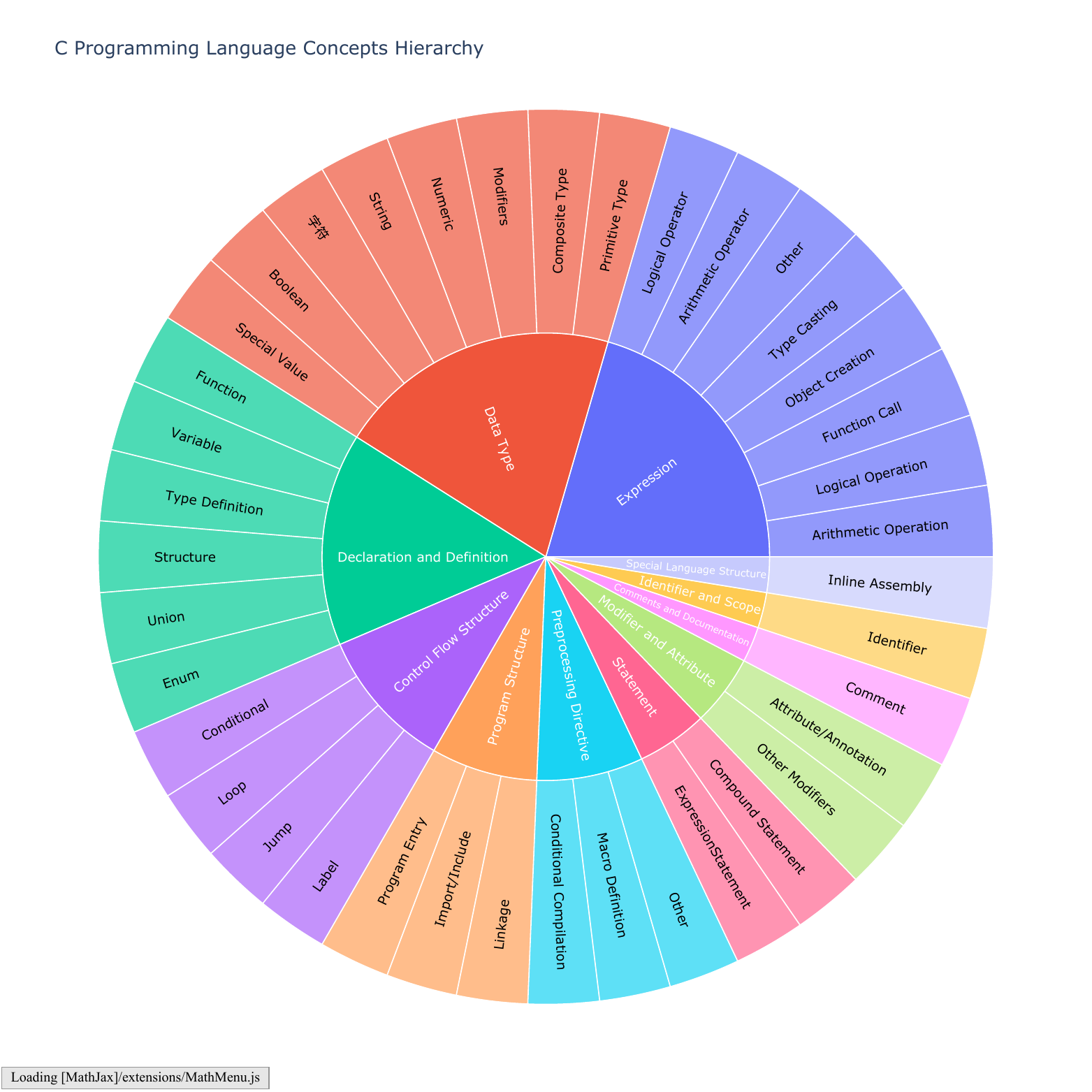}}
	\subfigure[Go]{\includegraphics[width=0.25\textwidth]{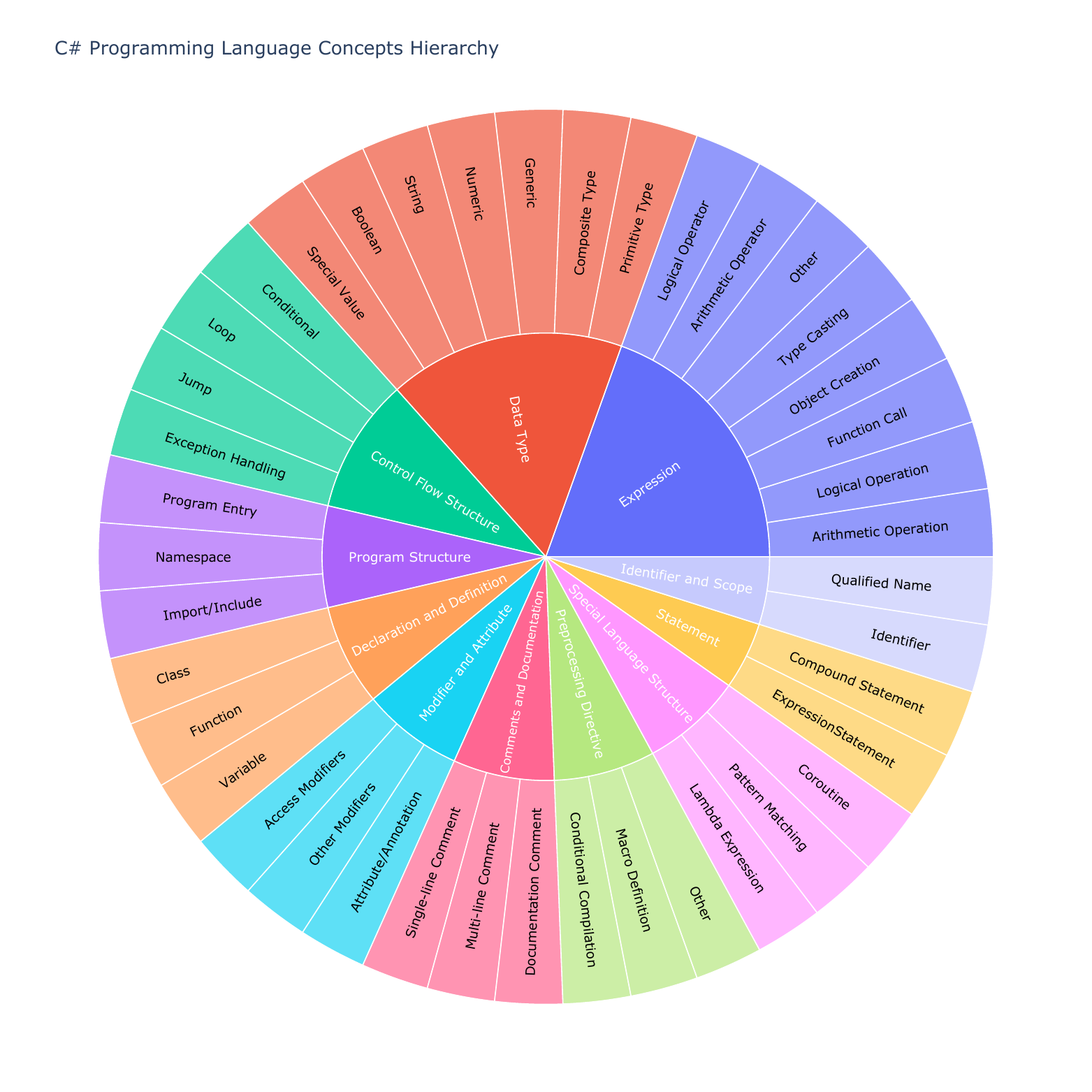}}
	\subfigure[Scala]{\includegraphics[width=0.25\textwidth]{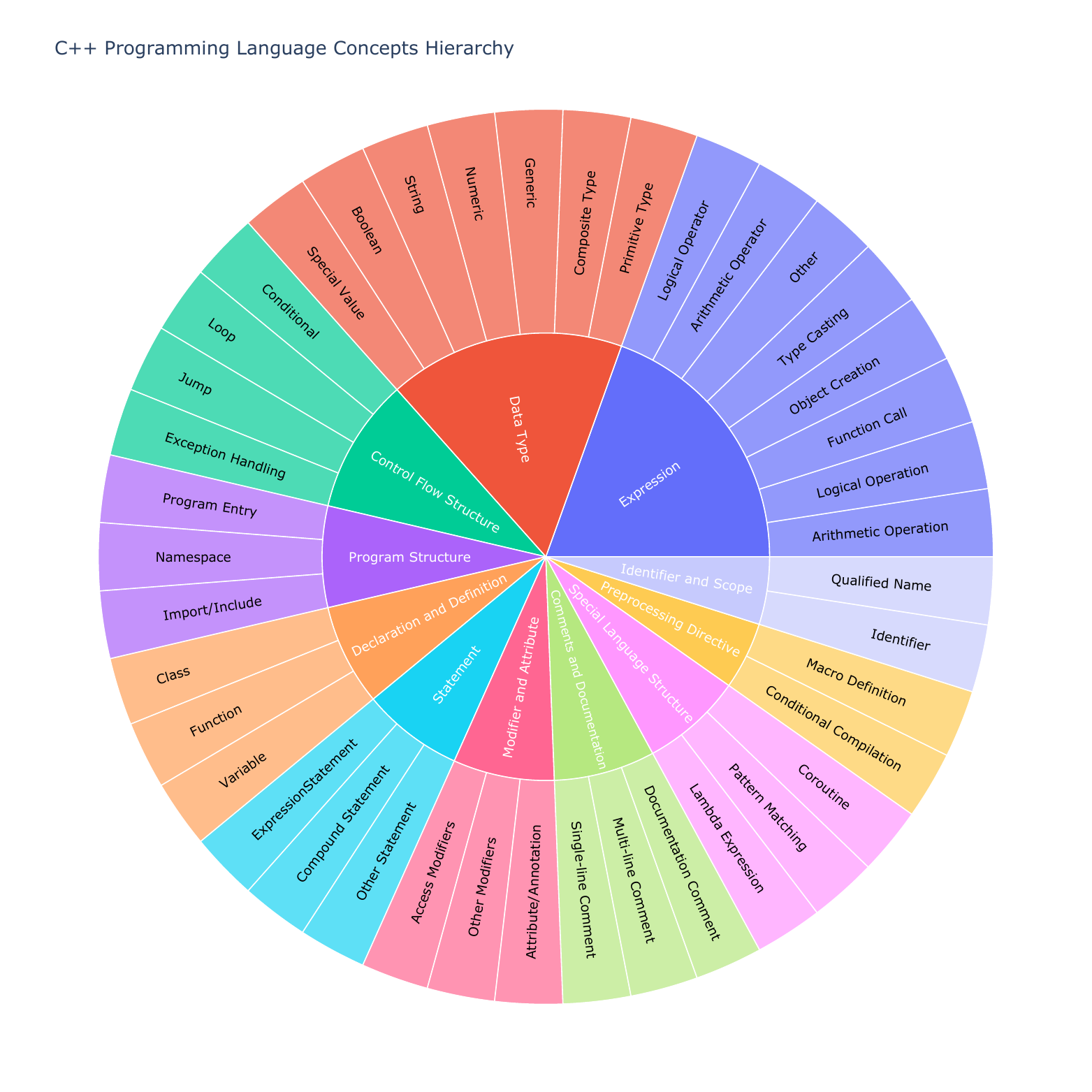}}
 	\subfigure[Java]{\includegraphics[width=0.25\textwidth]{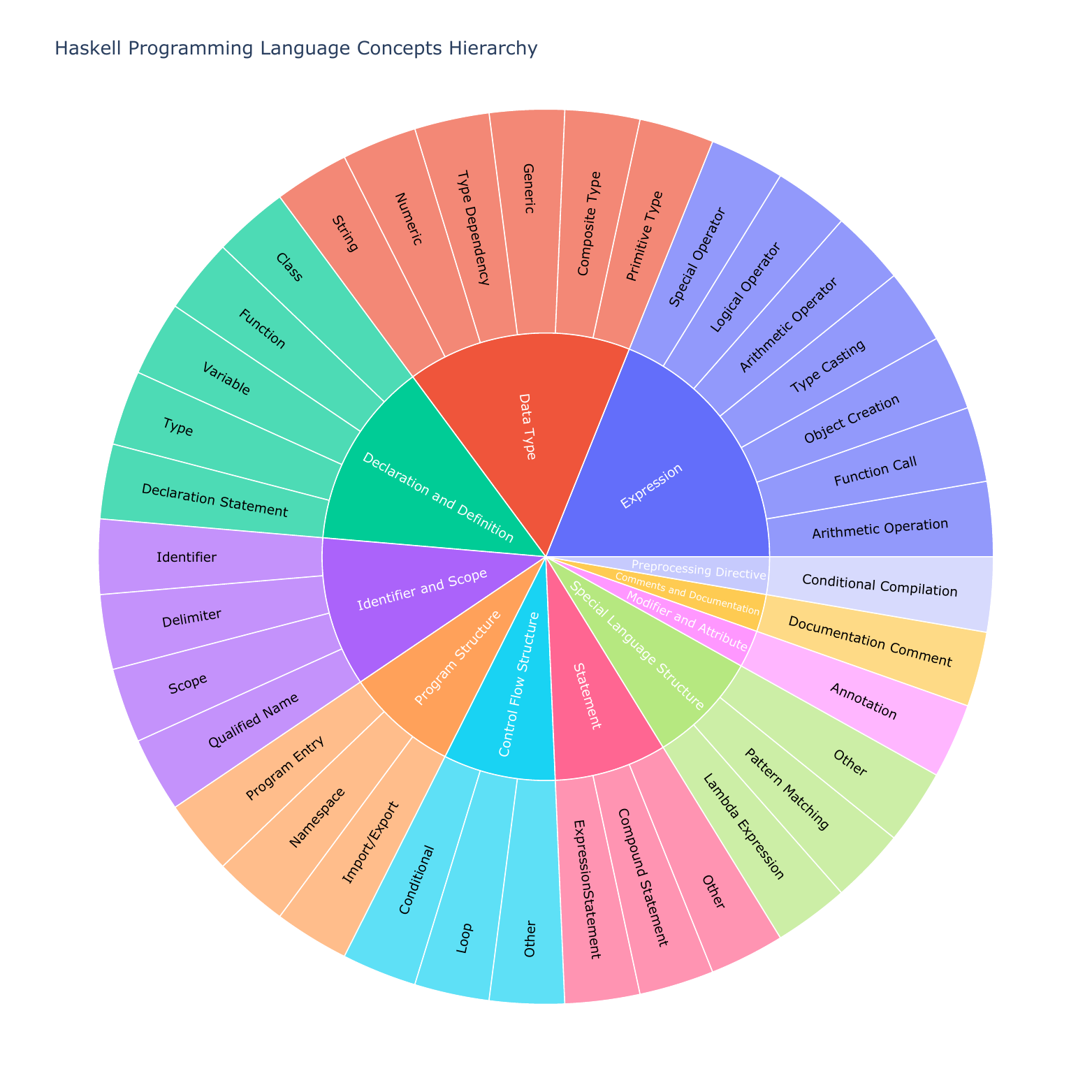}}
	\subfigure[Go]{\includegraphics[width=0.25\textwidth]{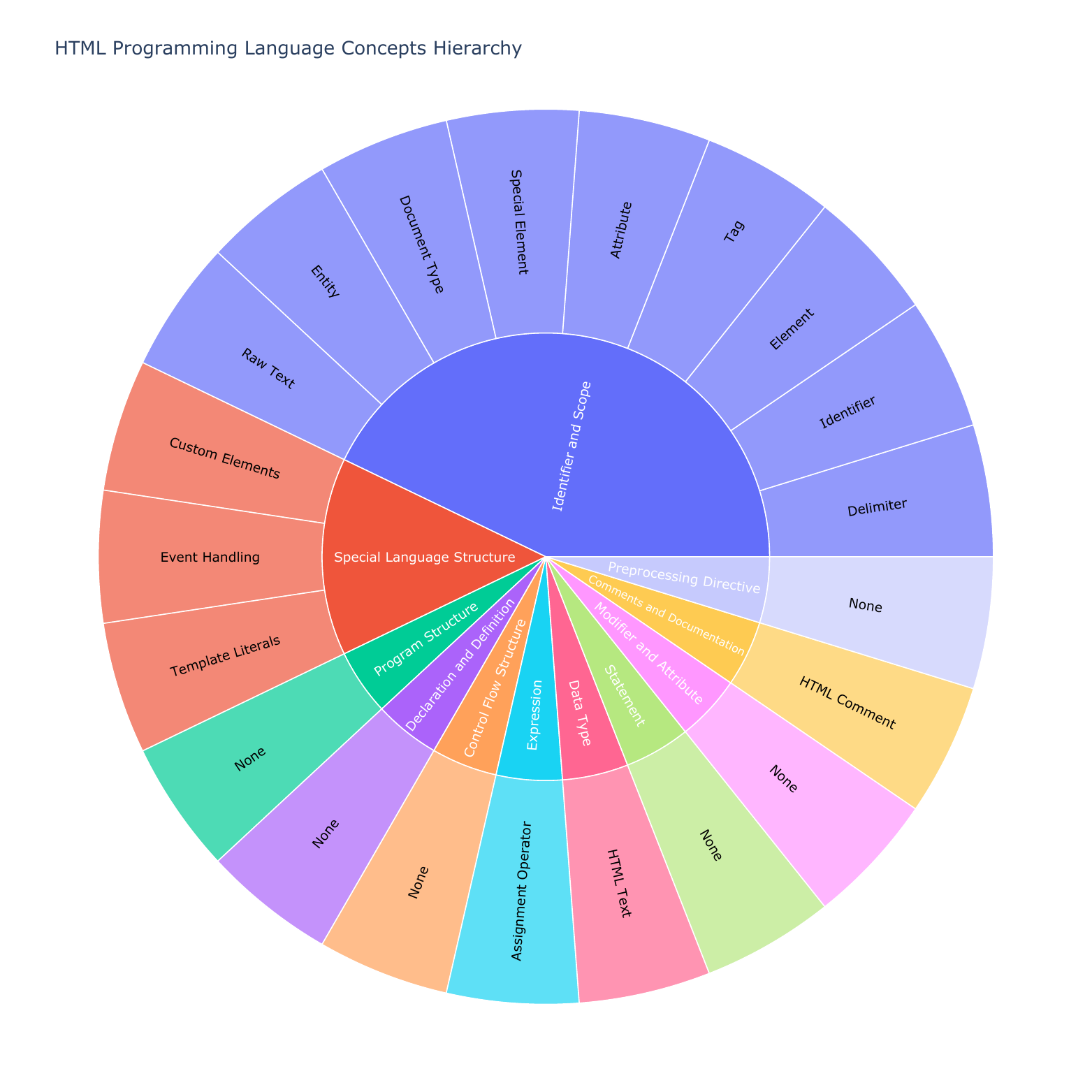}}
	\subfigure[Scala]{\includegraphics[width=0.25\textwidth]{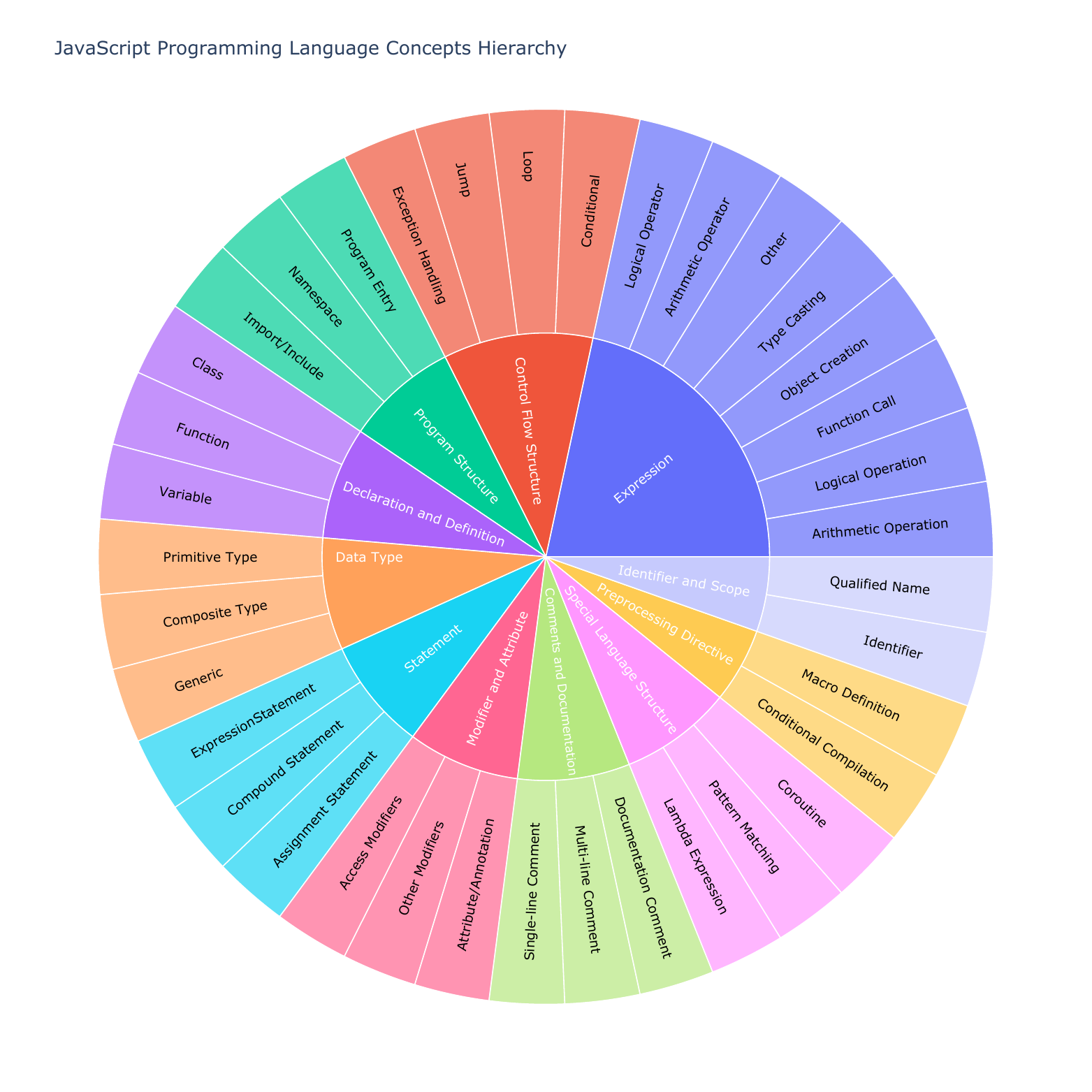}}
 	\subfigure[Java]{\includegraphics[width=0.25\textwidth]{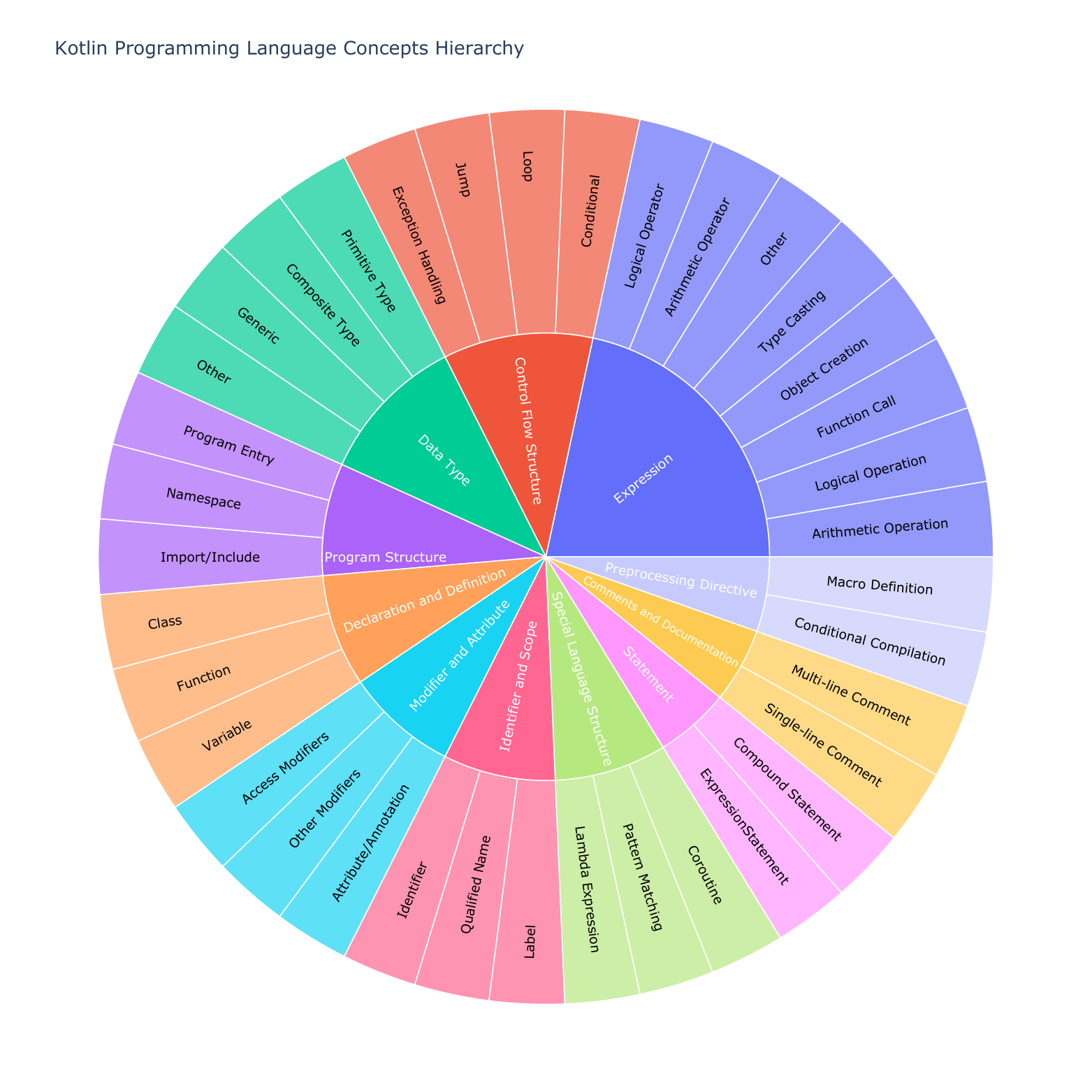}}
	\subfigure[Go]{\includegraphics[width=0.25\textwidth]{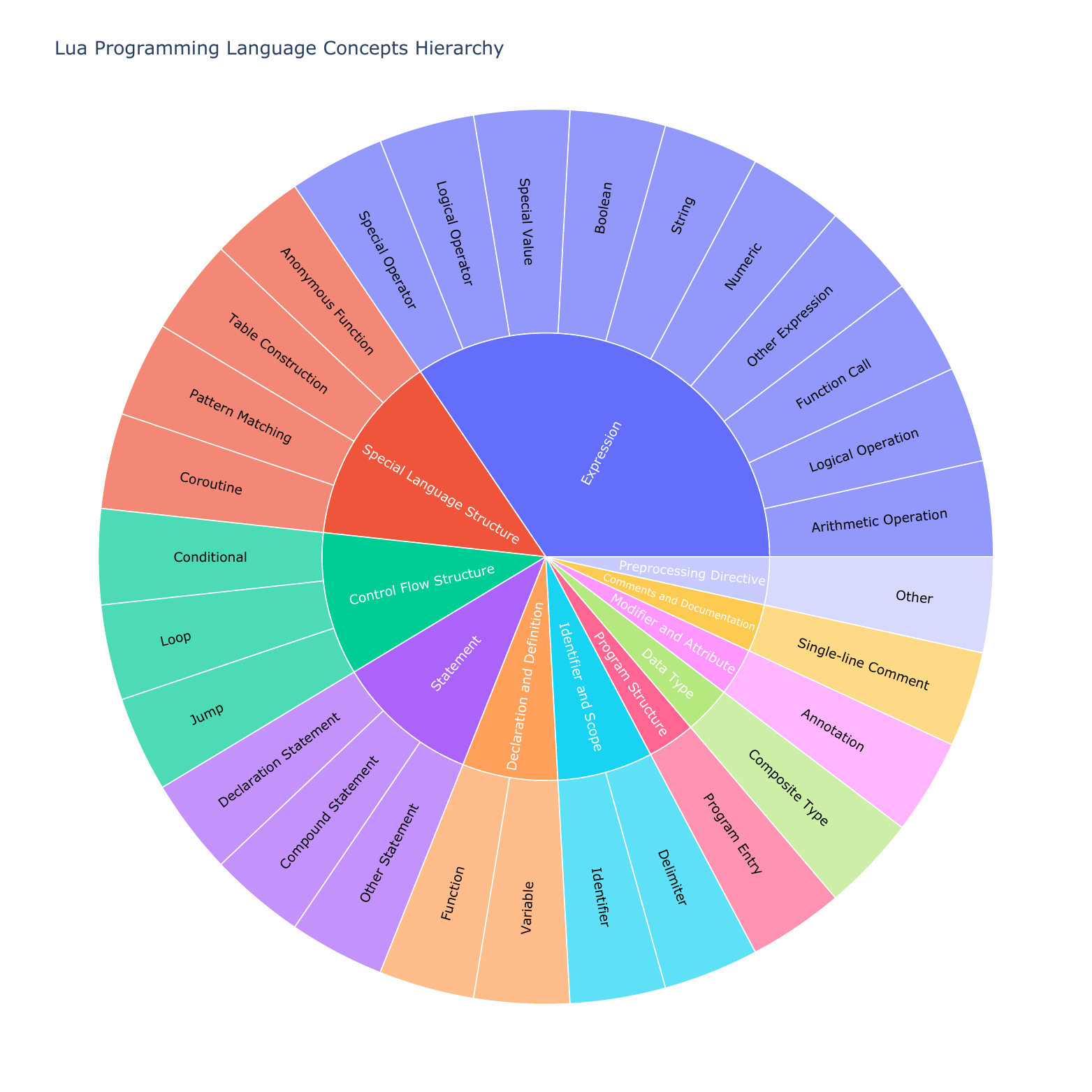}}
	\subfigure[Scala]{\includegraphics[width=0.25\textwidth]{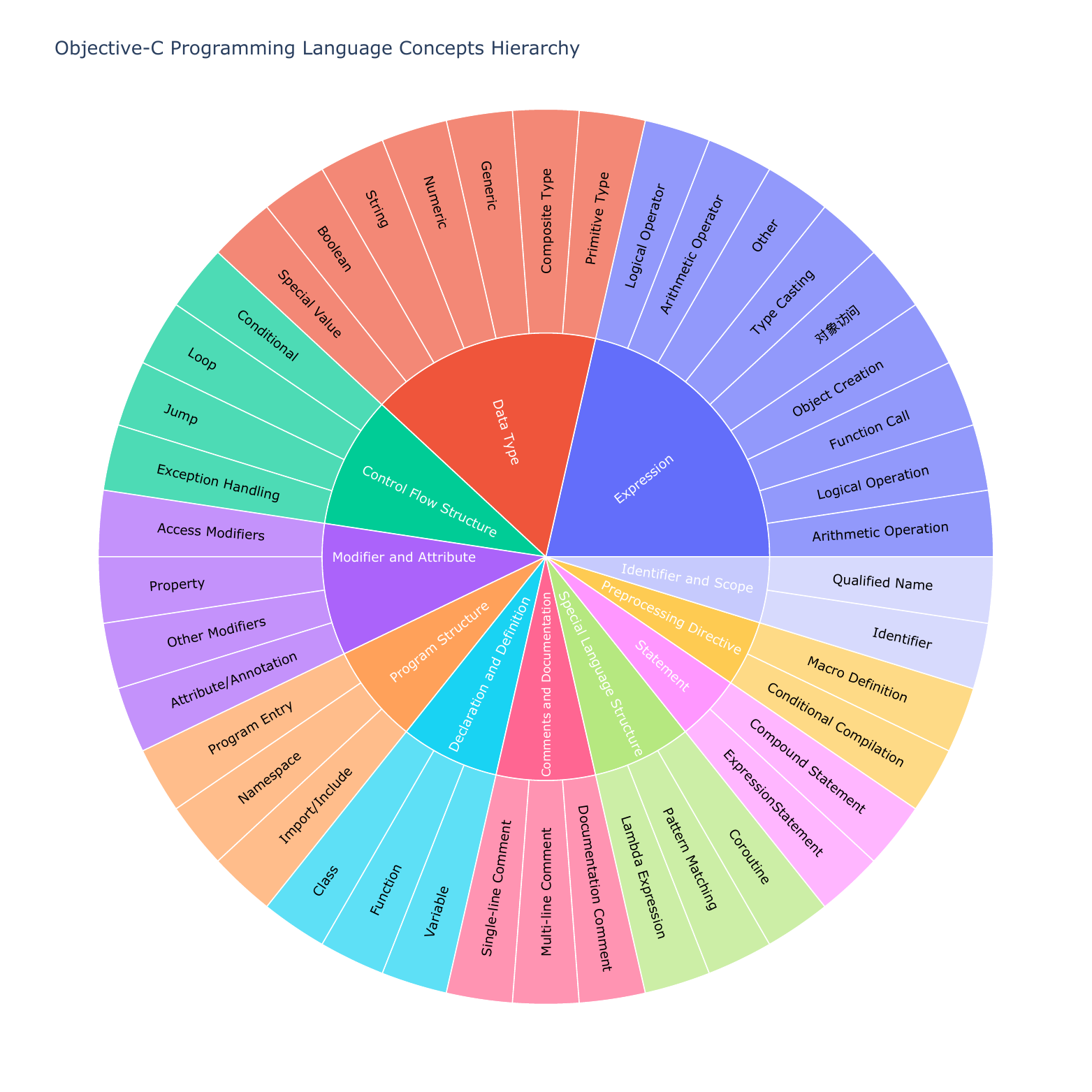}}
 	\subfigure[Java]{\includegraphics[width=0.25\textwidth]{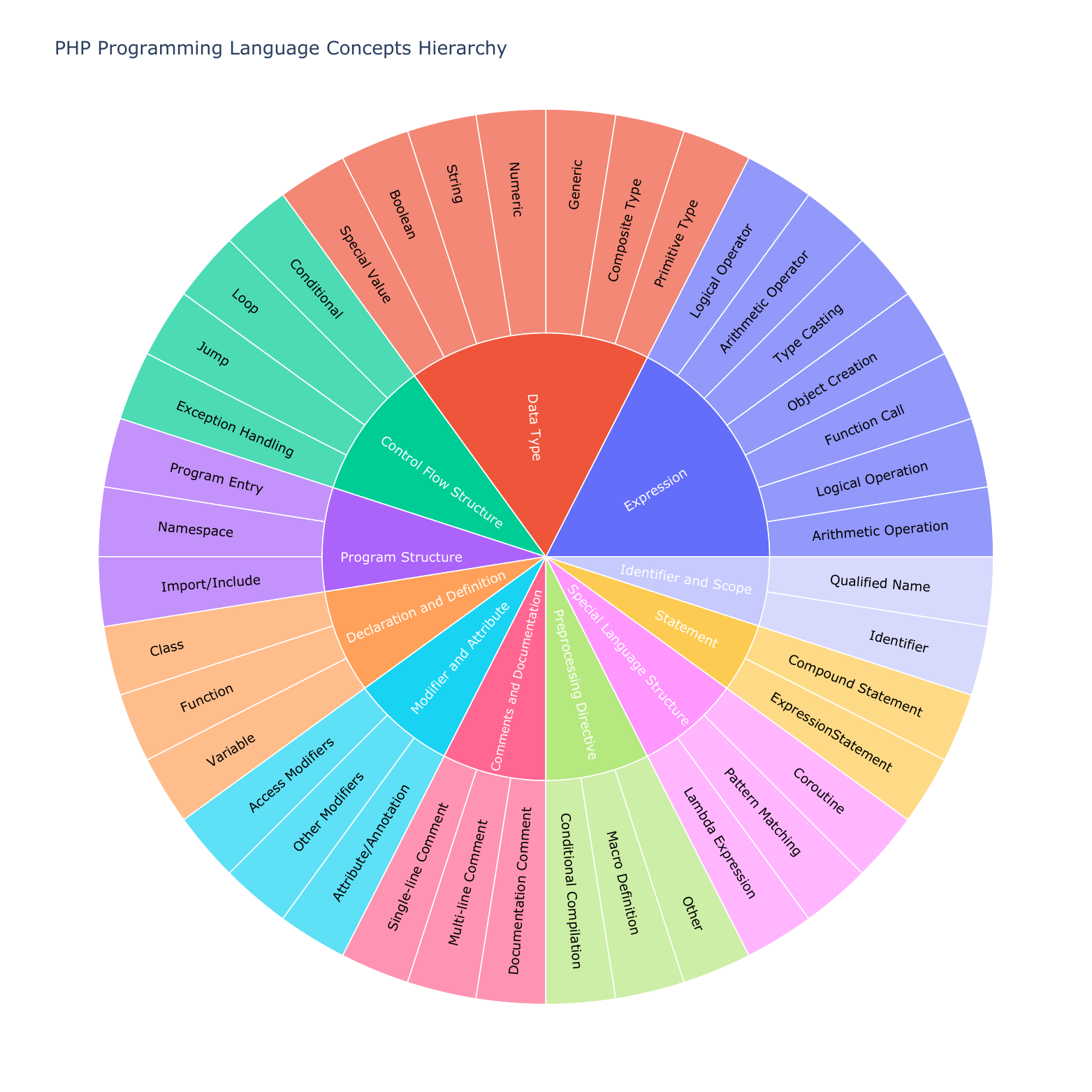}}
	\subfigure[Go]{\includegraphics[width=0.25\textwidth]{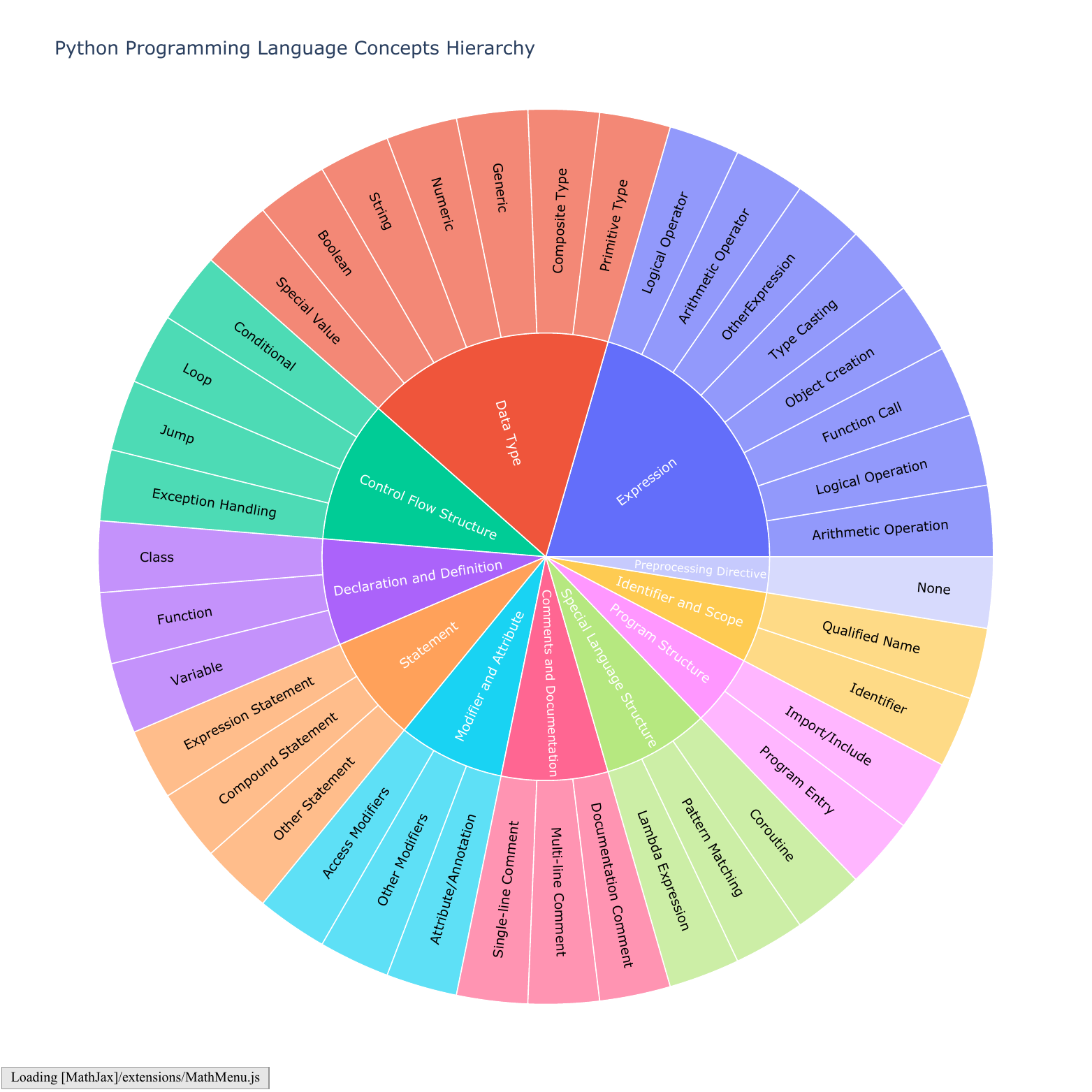}}
	\subfigure[Scala]{\includegraphics[width=0.25\textwidth]{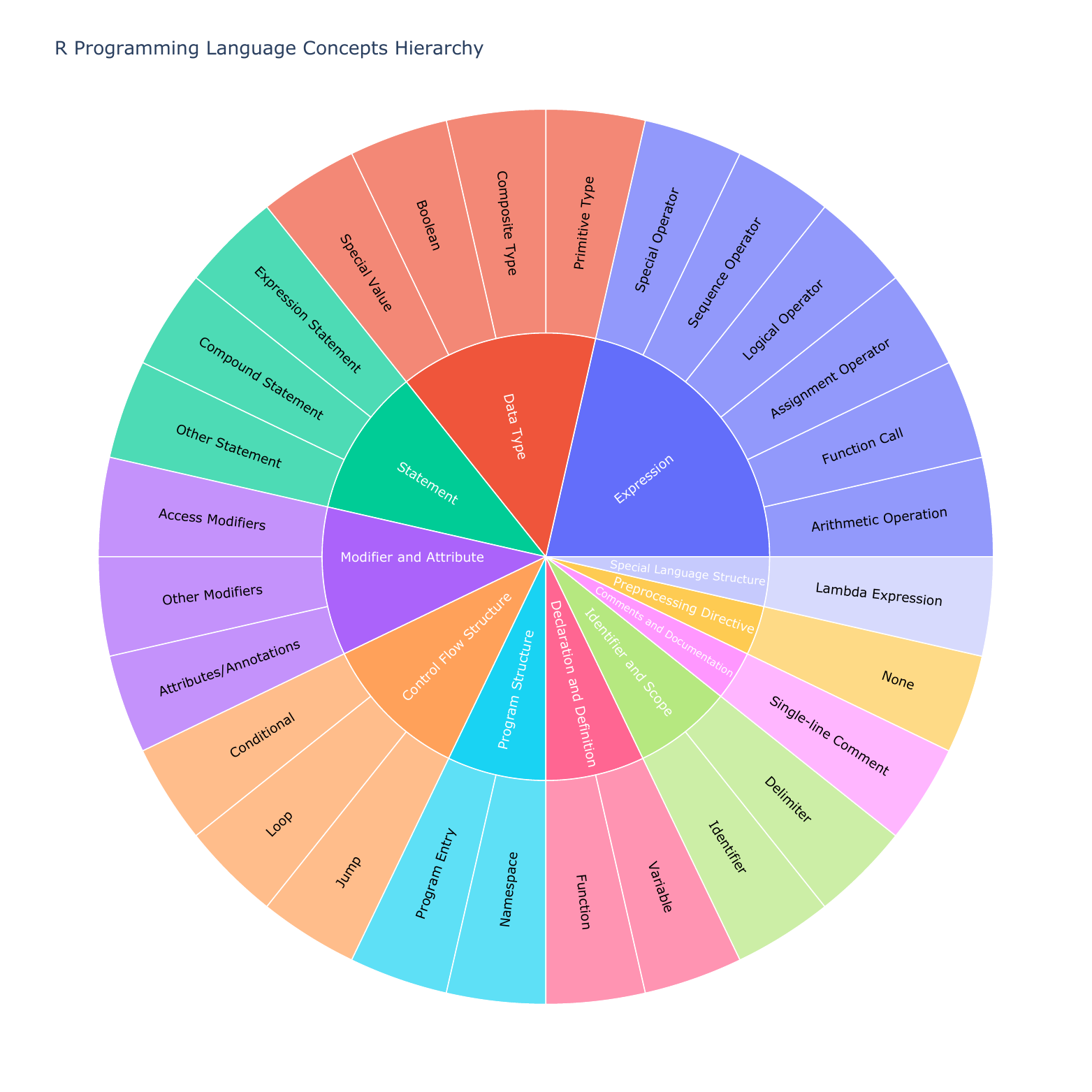}}
 	\subfigure[Java]{\includegraphics[width=0.25\textwidth]{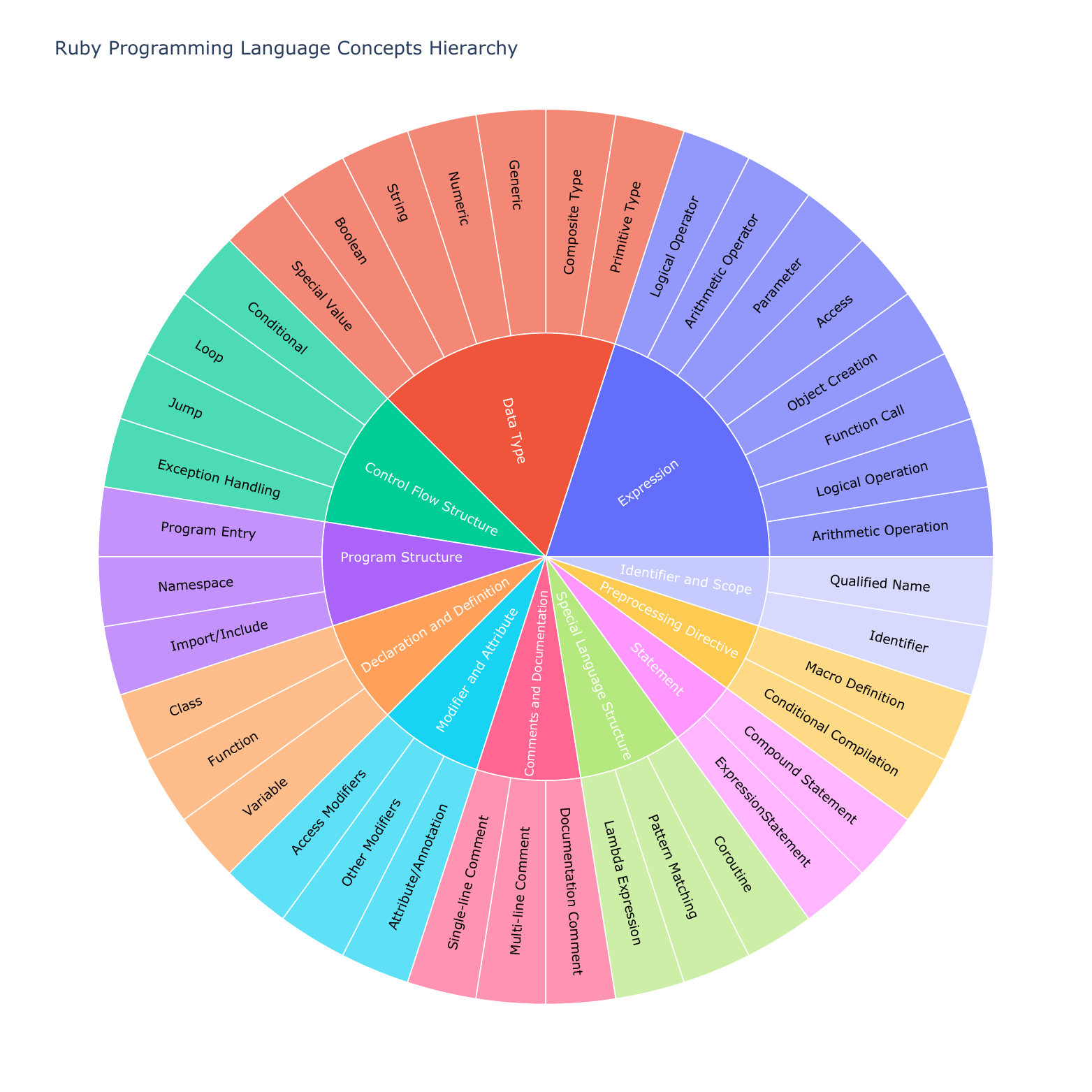}}
	\subfigure[Go]{\includegraphics[width=0.25\textwidth]{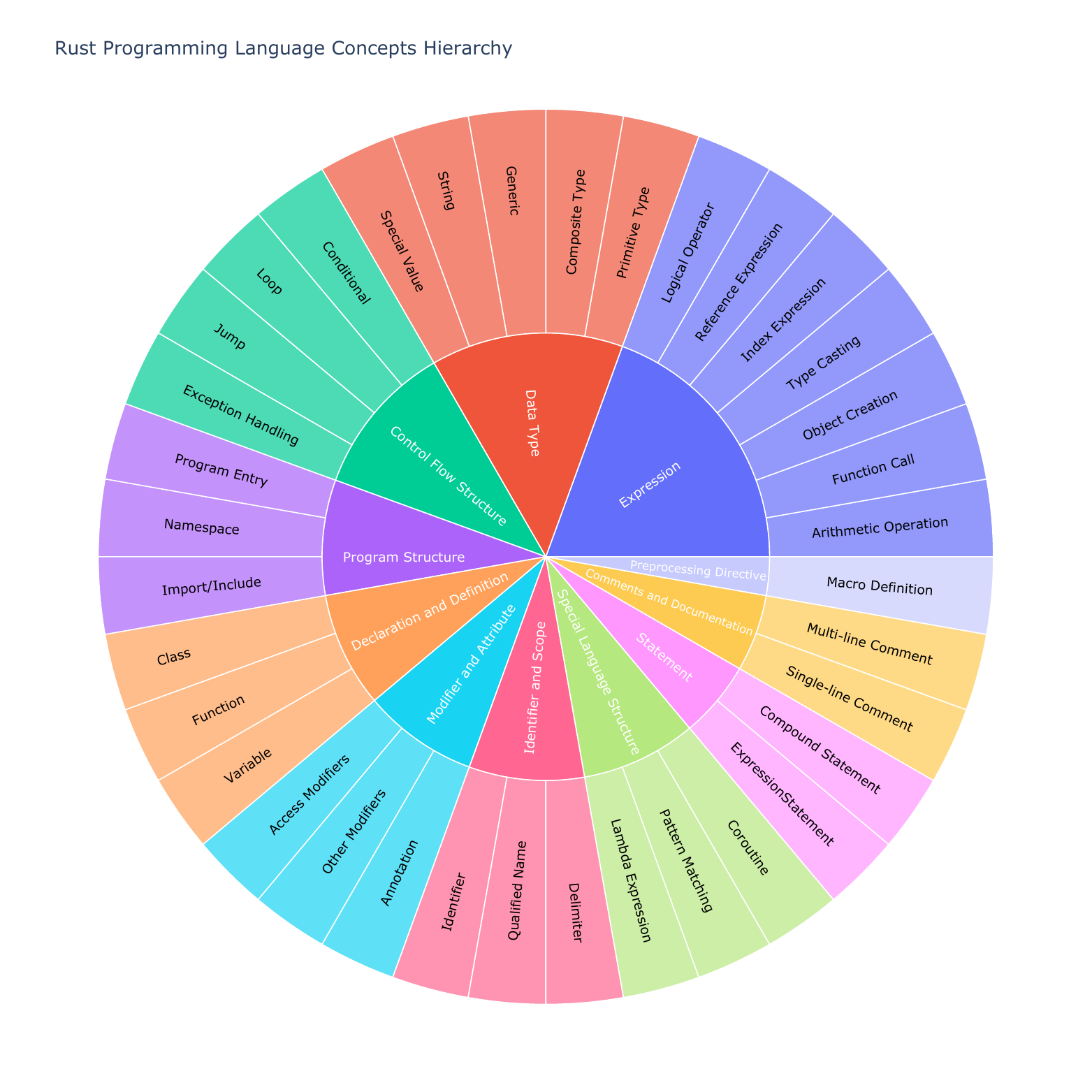}}
	\subfigure[Scala]{\includegraphics[width=0.25\textwidth]{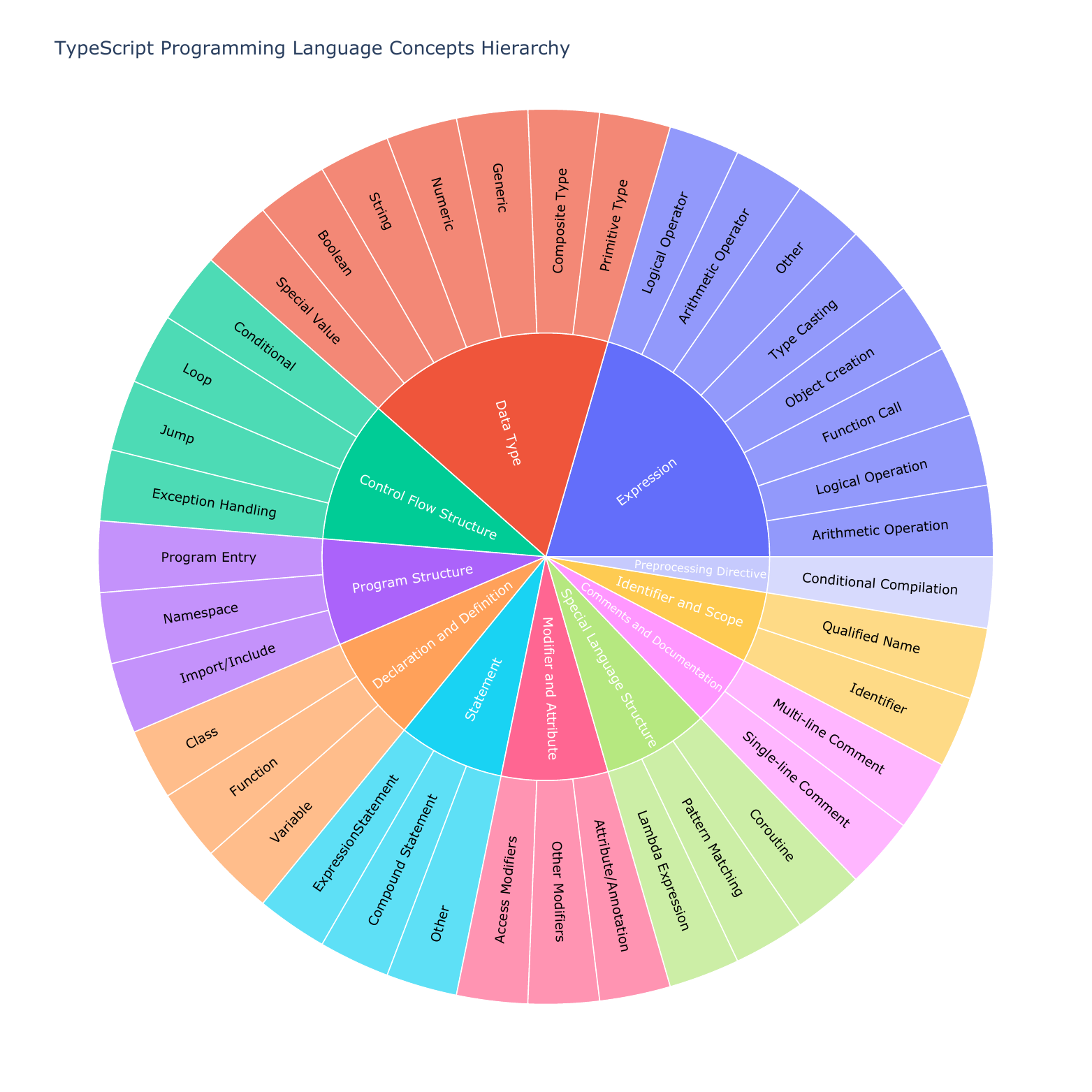}}
	\caption{Semantic-level annotations on different types of programming languages. ``none'' is used if this language does not have corresponding subcategories.}
	\label{fig:semantic-label-app}
\end{figure*} 

\subsection{Broader Impacts} {In this paper, we propose a repository-level code completion benchmark with 18 programming languages. Therefore, we hope our work can enhance the improvements on the multilingual repository-level code completion task.}

\subsection{Limitations} First, 
there are several hyperparameters (e.g., training sizes, input length) to tune,
which is laborious and expensive.
Second,
the current work only focuses on the repository-level code completion task,
where other repository-level code intelligence tasks are not considered.
Third,
only textual similarity scores (EM and ES) are used and execution-based evaluation based on test cases is not applied,
which may not reflect the performance of different code LLMs well.
\label{app:limitation}



\subsection{Details of the Baseline Models}
\label{app:models}
\noindent\textbf{StarCoder} \citep{li2023starcoder} is a series of generative language models (e.g., 7B, 15.5B). These decoder-only models are trained on the Stack dataset~\citep{kocetkov2022stack} and can support 8K tokens in context.

\noindent\textbf{DeepSeekCoder} \citep{guo2024deepseek} is  a collection of code-oriented models with capacities from 1.3B to 33B parameters. Trained on a manually curated 2-trillion-token corpus, these models leverage Fill-in-the-Middle (FIM)~\citep{bavarian2022efficient} and Rotary Position Embedding (RoPE) ~\citep{su2024roformer} techniques,
which enables efficient code generation and infilling within a 16K token window.

\noindent\textbf{Code Llama}~\citep{codellama} is a family of code large language models based on Llama 2 \citep{touvron2023llama} with 7B, 13B, 34B, and 70B parameters. 
While trained on 16K token sequences, these models can handle inputs up to 100K tokens during inference. 
\subsection{Discussion on no execution-based evaluation}
Current datasets for repository-level code completion evaluation, such as CrossCodeEval~\citep{ding2023cceval} and RepoBench~\citep{liu2023repobench}, only assess textual similarity between predictions and ground-truth. We hypothesize this limitation stems from several challenges:
Firstly, generating comprehensive unit tests for each completion position in a repository is problematic. Single-line completions often fail to construct executable functions, and ensuring adequate test coverage is difficult.
Secondly, execution-based evaluation necessitates creating diverse environments for each repository, accommodating various software packages and hardware requirements. This process is intricate and challenging to implement.
Thirdly, existing benchmarks with unit tests typically focus on simpler scenarios, like single-file completions or function body generation. Examples include commonly used datasets such as Humaneval~\cite{humaneval_x} and MBPP~\cite{mbpp}.
Despite these obstacles, we recognize the importance of execution-based evaluation for accurately assessing code completion effectiveness, and we will continue to investigate how to evaluate repository-level code completion well.

\subsection{More experiments}
\begin{figure}[t]
    \centering
    \includegraphics[width=1.0\linewidth]{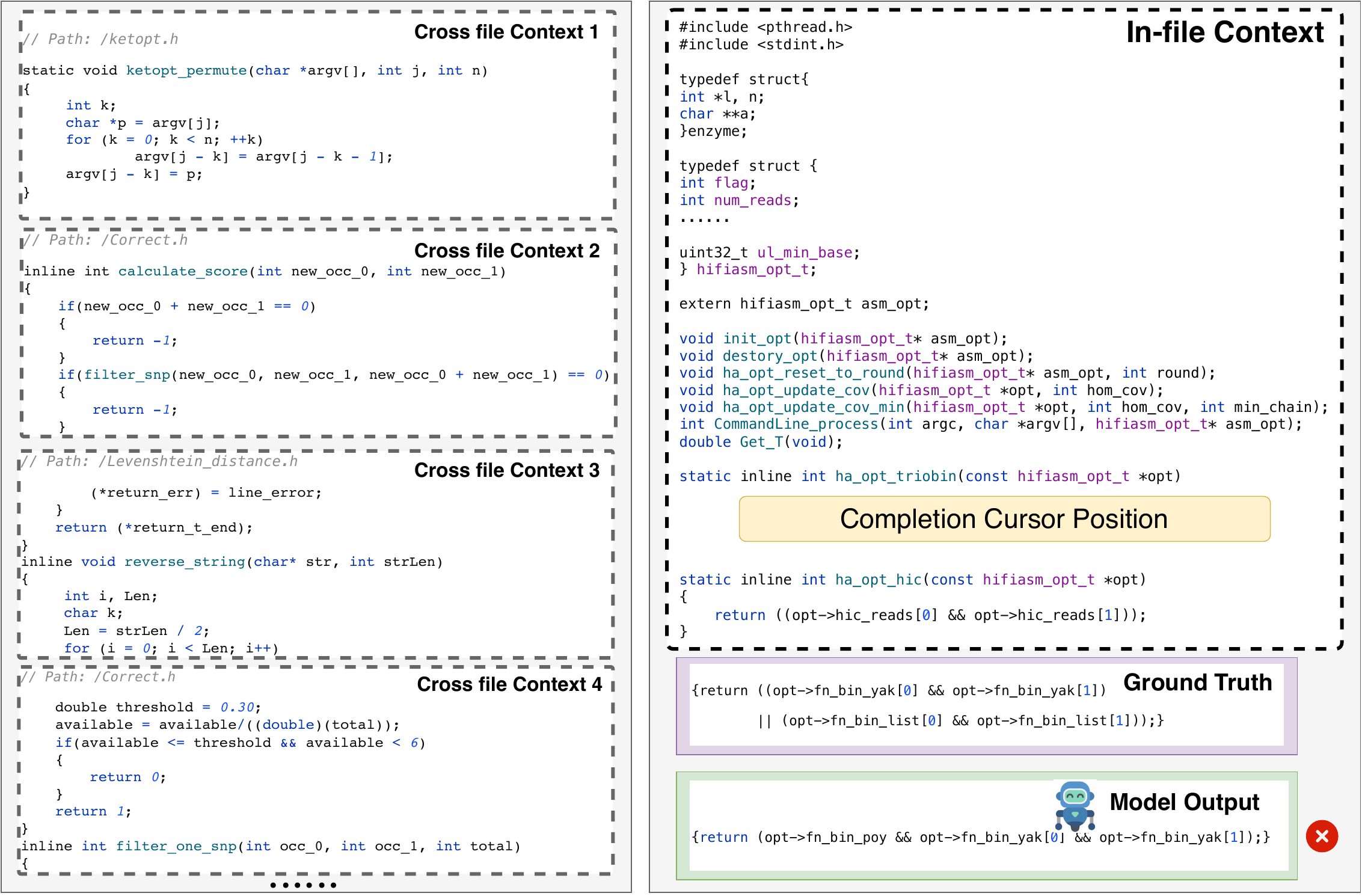}
    \caption{Visualization on failure case for the C language.}
    \label{fig:failure}
\end{figure}

\begin{figure}[t]
    \centering
    \includegraphics[width=1.0\linewidth]{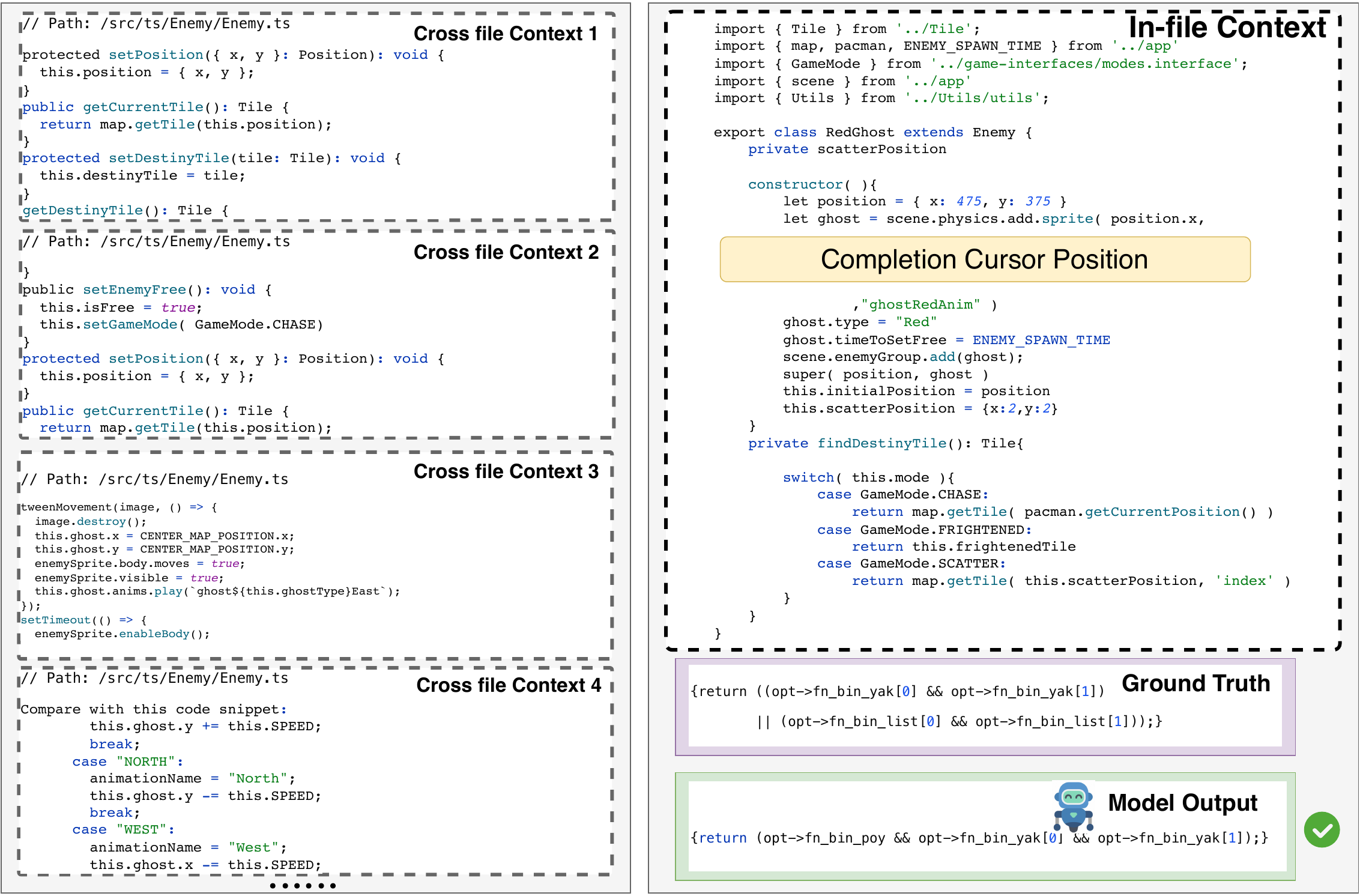}
    \caption{Visualization on success case for the TypeScript language.}
    \label{fig:success}
\end{figure}

\begin{itemize}
    \item We provide the analysis on the bucket levels in Fig.~\ref{fig:level-python-12} and Fig.~\ref{fig:level-python-6},
    respectively.
    \item We analyze the effect of different semantic levels on Rust, Objective-C and Haskell in Fig.~\ref{fig:semantic-levels}.
        respectively.
        \item We provide the semantic-level annotations on another 15 languages in Fig.~\ref{fig:semantic-label-app}.
            \item We provide the visualization on the completion cases in Fig.~\ref{fig:failure} and Fig.~\ref{fig:success}.
    
    \end{itemize}

\end{document}